\documentclass{article}

\usepackage{microtype}
\usepackage{graphicx}
\usepackage{subcaption}
\usepackage{booktabs} 
\usepackage{algpseudocode} 
\usepackage{amsmath,amssymb,amsfonts}
\usepackage{hyperref}
\usepackage[nameinlink]{cleveref}
\usepackage{multirow}
\usepackage{tcolorbox}
\tcbuselibrary{breakable} 
\usepackage{enumitem}
\usepackage{hyperref} 
\tcbuselibrary{breakable,listings}
\usepackage{listings}
\usepackage{colortbl}
\usepackage{xcolor}
\usepackage{bbm}

\usepackage{tcolorbox}
\tcbuselibrary{breakable,listings}
\usepackage{listings}

\lstdefinestyle{promptstyle}{
  basicstyle=\ttfamily\small,
  breaklines=true,
  columns=fullflexible,
  keepspaces=true,
  showstringspaces=false
}

\usepackage{hyperref}

\usepackage[preprint]{icml2026}

\usepackage{amsmath}
\usepackage{amssymb}
\usepackage{mathtools}
\usepackage{amsthm}
\usepackage{amsfonts}

\theoremstyle{plain}
\newtheorem{theorem}{Theorem}[section]
\newtheorem{proposition}[theorem]{Proposition}
\newtheorem{lemma}[theorem]{Lemma}
\newtheorem{corollary}[theorem]{Corollary}
\theoremstyle{definition}
\newtheorem{definition}[theorem]{Definition}
\newtheorem{assumption}[theorem]{Assumption}
\theoremstyle{remark}
\newtheorem{remark}[theorem]{Remark}

\usepackage[textsize=tiny]{todonotes}

\icmltitlerunning{MAPLE: Modality-Aware Post-training and Learning Ecosystem}

\begin{document}

\twocolumn[
  \icmltitle{MAPLE: Modality-Aware Post-training and Learning Ecosystem}



  \icmlsetsymbol{equal}{*}
  \icmlsetsymbol{correspondance}{+}

  \begin{icmlauthorlist}
    \icmlauthor{Nikhil Verma}{equal,correspondance,tail}
    \icmlauthor{Minjung Kim}{equal,hqlab}
    \icmlauthor{JooYoung Yoo}{equal,hqlab}
    \icmlauthor{Kyung-Min Jin}{equal,hqlab}
    \icmlauthor{Manasa Bharadwaj}{tail}
    \icmlauthor{Kevin Ferreira}{tail}
    \icmlauthor{Ko Keun Kim}{hqlab}
    \icmlauthor{Youngjoon Kim}{hqlab}
    
  \end{icmlauthorlist}

  \icmlaffiliation{tail}{LG Electronics Toronto AI Lab, Toronto, Canada}
  \icmlaffiliation{hqlab}{LG Electronics CTO AI Lab, Seoul, Republic of Korea}

  \icmlcorrespondingauthor{Nikhil Verma}{nikhil.verma@lge.com}

  \icmlkeywords{Modality-stratified RLHF, Multimodal reinforcement learning, Gradient variance reduction, Curriculum learning, Multimodal benchmarks}

  \vskip 0.3in
]



\printAffiliationsAndNotice{\icmlEqualContribution}

\begin{abstract}

Multimodal language models now integrate text, audio, and video for unified reasoning. 
Yet existing RL post-training pipelines treat all input signals as equally relevant, ignoring which modalities each task actually requires.
This modality-blind training inflates policy-gradient variance, slows convergence, and degrades robustness to real-world distribution shifts where signals may be missing, added, or reweighted. 
We introduce \textbf{MAPLE}, a complete modality-aware post-training and learning ecosystem comprising: 
(1) {MAPLE-bench}, the first benchmark explicitly annotating minimal signal combinations required per task; 
%
(2) {MAPO}, a modality-aware policy optimization framework that stratifies batches by modality requirement to reduce gradient variance from heterogeneous group advantages;
(3) Adaptive weighting and curriculum scheduling that balances and prioritizes harder signal combinations.
Systematic analysis across loss aggregation, clipping, sampling, and curriculum design establishes MAPO{'}s optimal training strategy. 
Adaptive weighting and curriculum focused learning further boost performance across signal combinations.
MAPLE narrows uni/multi-modal accuracy gaps by 30.24\%, converges 3.18$\times$ faster, and maintains stability across all modality combinations under realistic reduced signal access. 
MAPLE constitutes a complete recipe for deployment-ready multimodal RL post-training.
\end{abstract}

\section{Introduction}

Recent advances in Multimodal Language Models (MLMs) enable unified reasoning across text, audio, and video sensory streams~\cite{team2024gemini, hurst2024gpt, xu2025qwen2}.
RL post-training has become central to adapting these models for complex decision-making and alignment objectives~\cite{schulman2017proximal, yin2024mllm-survey, rafailov2023direct}.
Effective post-training must simultaneously sustain high task accuracy, robustness to distributional shifts, and efficiency under realistic compute limits~\cite{khatri2025art, mckinzie2023robustness}.
Yet most current approaches, inherited from unimodal language settings, treat all multimodal trajectories as if drawn from a single joint distribution.
This assumption neglects modality-specific cues that determine which input channels truly matter for solving a given query~\cite{wei2025unsupervised}.

\begin{figure}[t]
    \centering
    \includegraphics[width=0.78\linewidth]{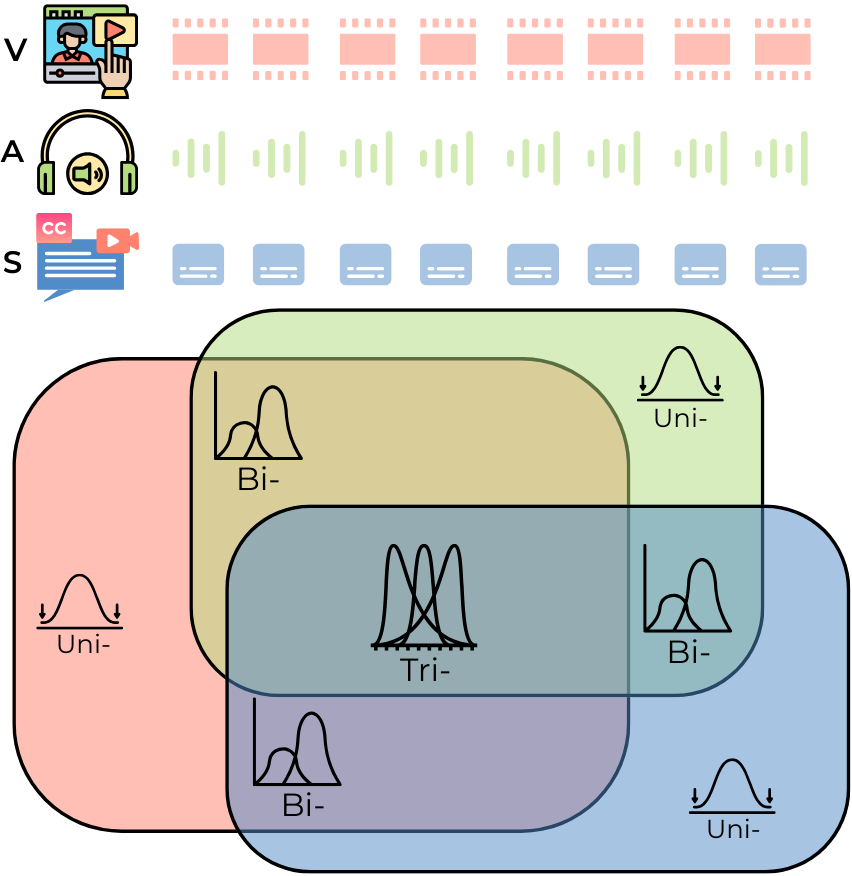}
    \caption{\textbf{Heterogeneous multimodal latent geometry needs stratified RL.} 
    Unimodal clusters (V: video, A: audio, S: subtitle) connect through bi-modal (VA, VS, AS) and tri-modal (VAS) manifolds. 
    Modality-blind RL mixes trajectories from these 7 distinct regions indiscriminately, inflating gradient variance across heterogeneous reward scales. 
    MAPLE's stratified training assigns each trajectory to its native signal combination for stable optimization.
    }
    \label{fig:latent_space}
\end{figure}

In real-world deployments, queries may typically depend on subset of input streams (text, audio, or video) rather than all modalities simultaneously. 
Fig.~\ref{fig:latent_space} illustrates this: distinct unimodal clusters gradually fuse into shared bi- and tri-modal manifolds in latent space.
However, existing RL post-training pipelines process every rollout as if all modalities contribute equally, blending fundamentally heterogeneous interactions.
Under value-model-free methods like Group Relative Policy Optimization~(GRPO)~\cite{shao2024deepseekmath}, trajectories that leverage rich multimodal redundancy are jointly optimized with single-channel reasoning tasks.
This uniform treatment biases policy updates toward dominant modality combinations while under-training regimes with sparse or noisy signals, precisely those most common in real test conditions.

This representational heterogeneity across modality regimes translates directly into measurable optimization challenges.
When reward scales and noise characteristics are conflated across modality subsets, modality-blind batching amplifies policy-gradient variance, mirroring known batch-composition effects in policy optimization~\cite{che2023correcting,vasan2024deep} and cross-modality gradient conflicts~\cite{anonymous2025goal}.
Empirically, our analysis~(Sec~\ref{sec:opt_trn_strat}) shows that mixed-modality batches exhibit higher gradient-norm volatility than modality-stratified batches, leading to slower convergence and instability patterns~\cite{rha2025learning}.
%
More fundamentally, modality-blind optimization collapses modality-specific structure. 

The consequences of this mismatch intensify under deployment-time distribution shifts.
In real-world settings, models frequently encounter ``missing-at-test'' or partial-observation regimes that violate the full-signal assumptions made during training~\cite{mckinzie2023robustness}.
Robust policies must therefore operate reliably under absent, spurious, or reweighted input signal, conditions where failures tend to be systematic rather than random.
Modality-unaware optimization exacerbates this gap by training on noisy, mixed-modality batches while evaluating under shifted signal configurations.
Across benchmark evaluations, omni-modal models experience substantially greater degradation than modality-specialized counterparts~\cite{jiang2025specific, li2025mbq}.

We address these challenges through three concrete contributions that together constitute \textbf{MAPLE}, a \textit{modality-aware post-training and learning ecosystem} for multimodal reinforcement learning.
First, we present \textbf{MAPLE-bench}, a modality-aware benchmark where each instance is explicitly annotated with the minimal subset of signals, drawn from video, audio, and subtitles, necessary to solve the target task.
This design enables principled conditioning of both data sampling and policy optimization, bridging the gap between synthetic multimodal mixtures and realistic input dependencies.
{MAPLE-bench} spans multiple-choice question answering and open-ended omni-modal captioning across seven signal subsets, including uni-modal, bi-modal, and tri-modal configurations, thereby capturing the diversity of plausible deployment regimes.

Second, we revisit value-model-free RL in multimodal settings.
While recent optimizers such as GRPO~\cite{shao2024deepseekmath}, Dr.~GRPO~\cite{liu2025drgrpo}, and DAPO~\cite{yu2025dapo} demonstrate strong performance in unimodal domains, their design choices do not generalize well to heterogeneous signal regimes.
We conduct a systematic ablation of loss aggregation, clipping behaviors, dynamic sampling, and data filtering strategies, explicitly conditioning each component on the active signal subset.
This analysis yields \textbf{Modality-Aware Policy Optimization (MAPO)}, a unified optimization framework that stabilizes multimodal post-training without added critics or auxiliary regularization.
We benchmark MAPO against a modality-unaware policy that always accesses all signals.

Finally, we propose \textbf{Adaptive-training strategies} that progressively trains from uni-modal reasoning tasks toward multi-signal integration.
This adaptation explicitly exposes how varying signal diversity influences convergence dynamics and robustness.
Across all settings, {MAPLE} attains stable and competitive results, while operating under reduced modality access.
%
It also narrows accuracy gaps between uni- and multi-modal regimes by $30.24\%$ and converges $3.18\times$ faster than full-signal training dependent on redundant modalities.
Taken together, \textbf{MAPLE} constitutes a complete and practical recipe that integrates benchmark design, policy optimization, and curriculum learning for efficient, robust, and deployment-ready multimodal RL post-training.




\noindent
In summary, this work:
\begin{enumerate}
    \item Introduces \textbf{MAPLE-bench}, a modality-aware benchmark suite for QA and captioning with \emph{Required Modality Tags} for explicit control of sampling, optimization, and evaluation.
    \item Develops \textbf{MAPO} and adaptive training strategies that exploit per-sample modality structure to stabilize training and enhance performance across diverse signal configurations.
    \item Demonstrates, through extensive experiments, that MAPLE yields efficient and robust multimodal RL post-training under real-world variability.
\end{enumerate}
 
\section{Related Works}

\paragraph{Multimodal Large Language Models.}
Recent multimodal LLMs unify text, vision, and audio through shared token interfaces, with systems like Qwen2.5-Omni, Gemini~3, and Phi-4-Multimodal demonstrating strong omni-modal reasoning~\cite{yin2024mllm-survey,qwen25-omni-arxiv,phi4-multimodal}. 
These models typically assume full modality availability during both training and evaluation, treating heterogeneous signals as uniformly relevant inputs. 
Extended architectural comparisons appear in Appendix~\ref{app:related}.

\paragraph{Multimodal Benchmarks.}
Benchmarks such as MME, MMBench, OmniBench, and MAVERIX~\cite{liu2024mmbenchmultimodalmodelallaround,li2025omnibenchfutureuniversalomnilanguage,xie2025maverixmultimodalaudiovisualevaluation} assess cross-modal reasoning under complete signal access but rarely examine performance across varying modality combinations. 
Even ablation studies maintain fixed ground-truth answers despite changing inputs, conflating true information deficits with fusion failures. 
Table~\ref{tab:benchmark_comparison} (Appendix~\ref{app:related}) contrasts these limitations with MAPLE-bench's modality-conditioned evaluation.

\paragraph{Post-training Optimization.}
Value-model-free RL methods like GRPO, Dr.~GRPO, and DAPO~\cite{shao2024deepseekmath,liu2025drgrpo,yu2025dapo} excel in text domains through group-normalized advantages and simplified reward modeling. 
Modality-balancing~\cite{li2025mbq} and curriculum strategies~\cite{penas2023curriculum} address signal dominance but do not condition optimization on per-query required signal subsets. 
MAPLE extends these paradigms with modality-aware data stratification, loss aggregation, and curriculum design. 
Additional post-training variants are discussed in Appendix~\ref{app:related}.

\section{MAPLE-bench: A modality annotated benchmark }
\label{sec:datasets}

\begin{figure}[t]
    \centering
    \includegraphics[width=1.0\columnwidth]{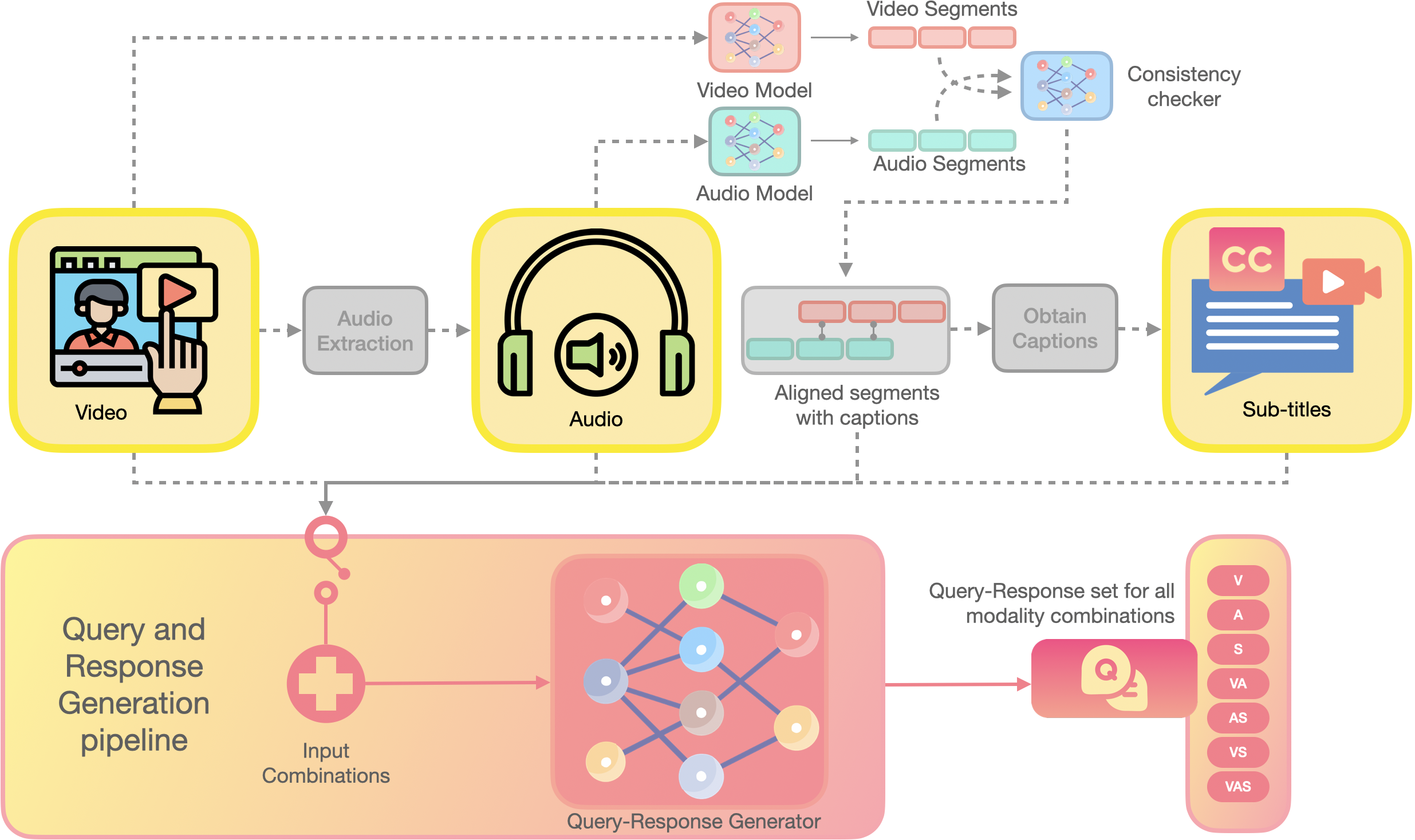}
    \caption{ \textbf{Overview of the MAPLE-bench curation pipeline.} Seed videos are sampled from large-scale corpora and decomposed into audio, visual, and textual streams. 
    Each modality undergoes independent annotation and cross-modal consistency alignment before generating captions and query–response pairs. 
    A modality-isolated prompting stage then synthesizes tasks across uni-, bi-, and tri-modal configurations, producing a balanced dataset for modality-aware post-training.}
    \label{fig:dataset_pipeline}
\end{figure}

We introduce {MAPLE-bench}, a modality-aware dataset designed for evaluating post-training strategies under heterogeneous signal access. 
MAPLE-bench extends subsets of the Daily-Omni~\cite{zhou2025dailyomni} and VAST-Omni~\cite{chen2023vast} corpora, curated to ensure strong multimodal coverage across video, audio, and text streams.

\subsection{Dataset Construction}

Our pipeline begins by sampling seed videos from these corpora and extracting aligned audio tracks. 
Each modality is independently segmented and annotated using domain-specific models, followed by a {cross-modal consistency and alignment phase} driven by Gemini-2.5-Flash~\cite{team2024gemini}. 
This stage ensures temporal coherence and semantic agreement between visual and acoustic narratives. 
From the aligned segments, captions and structured event descriptions are generated.

Next, we apply a \emph{modality-isolated protocol} based query–response generation procedure, prompting a reasoning model with controlled modality combinations to synthesize datasets for two complementary tasks: 
(i) open-ended captioning (generative) and 
(ii) multiple-choice question answering (discriminative). 
Each example is labeled with a set of Required Modality Tags~(RMTs) denoting the minimal modality subset necessary to solve the task, enabling fine-grained control over modality conditioning during training and evaluation.
An overview of the full curation and query–response generation pipeline is illustrated in Fig.~\ref{fig:dataset_pipeline}.

The curated corpus yields query–response pairs across uni-, bi-, and tri-modal regimes. 
To validate data integrity, we apply multiple LLM-based consistency checks and manual audits on the test split. 
This process establishes MAPLE-bench as a robust foundation for studying modality-aware reinforcement learning. 
Further synthetic dataset curation details are provided in Appendix~\ref{app:dataset_generation}, and dataset statistics are summarized in the following section.

\subsection{Dataset Statistics}

MAPLE-bench spans all seven Required Modality Tags: $\{\mathrm{V}, \mathrm{A}, \mathrm{S}, \mathrm{VA}, \mathrm{VS}, \mathrm{AS}, \mathrm{VAS}\}$, where V, A, and S denote video, audio, and subtitles respectively.

\noindent\textbf{MAPLE-QA.}
MAPLE-QA-train provides 47,893 QA pairs across 546 videos, each tagged with minimal required RMTs and structured as discriminative 4-option multiple choice questions for verifiable reward signals~\cite{guo2025deepseek}. 
The human-evaluated MAPLE-QA-eval contains 5,001 samples from a different set of 68 videos, following identical generation protocols. 
Automatic filtering via GPT-4o, GPT-5, and Gemini-2.5-Flash~\cite{hurst2024gpt, team2024gemini} in data generation pipeline removes trivially solvable pairs.

\noindent\textbf{MAPLE-Caption.}
We construct MAPLE-Caption-train with 5,120 captioning samples for post-training (no human curation). 
The human-curated MAPLE-Caption-eval benchmark comprises 5,348 samples, perfectly balanced at 764 samples per RMT (764 videos total). 
All evaluation captions are verified for factual consistency and modality coverage relative to provided signals.

A detailed comparison of MAPLE-bench against existing multimodal benchmarks appears in Appendix Table~\ref{tab:benchmark_comparison}.

\subsection{Evaluation Protocol}
\label{sec:evaluation_protocol}

We evaluate both base model and trained policy checkpoints using consistent inference settings. 
For QA tasks, pass@1 accuracy is estimated from 5 samples per query, leveraging the discriminative 4-option structure for exact binary rewards. 
Captioning uses LLM-as-Judge scoring~\cite{gu2024survey} with modality-conditioned reference captions to detect hallucination errors and modality-fusion across RMTs.
Scores lie in $[0,1]$.
To assess modality-aware post-training beyond standard accuracy, we introduce complementary metrics aligned with our core research problem.
All metrics use two-stage normalization: per-sample superset averaging, then RMT-balanced macro-average.

\begin{itemize}
    \item \textbf{Modality Gap}: Relative performance scaling across Uni/Bi/Tri-modal RMT groups.
    \item \textbf{Training Efficiency}: Wall-clock time per optimization step, normalized by batch size.
    %
    %
    \item \textbf{Fusion Gain}: Fraction of samples where multi-modal signal information outperform best uni-modal response.
\end{itemize}

These metrics directly quantify the stability, robustness, and efficiency claims from our introduction. 
Detailed prompts, judge model specifications, and additional analysis appear in Appendix~\ref{app:evaluation}.

\section{Modality-Aware Post-Training}
\label{sec:modality_aware_post_training}

\subsection{RL Training Pipeline}
\label{subsec:rl_training_pipeline}

Post-training optimizes policy $\pi_\theta$ via GRPO~\cite{shao2024deepseekmath} without KL regularization. 
The old policy used for generation $\pi_{\theta_\text{old}}^{\text{gen}}$ produces $G$ rollouts $\{y_i\}_{i=1}^G \sim \pi_{\theta_\text{old}}^{\text{gen}}(\cdot|x)$ for query $x \sim \mathcal{D}$ distribution, each scored $r_i = r(x,y_i)$.
Group advantages normalize within the generator batch:
$$
\hat{A}_i^G = \frac{r_i - \mu_G}{\sigma_G}, \quad \mu_G = \frac{1}{G}\sum r_i, \quad \sigma_G = \mathrm{std}(\{r_i\}).
$$

Trainer $\pi_\theta^{\text{train}}$ optimizes the token-level objective with symmetric clipping. Importance sampling ratios are:
$$
\rho_{i,t}(\theta) = \frac{\pi_\theta^{\text{train}}(y_{i,t} \mid x, y_{i,<t})}{\pi_{\theta_\text{old}}^{\text{gen}}(y_{i,t} \mid x, y_{i,<t})}, \quad \mathrm{clip}(\rho,1-\epsilon,1+\epsilon)
$$
The surrogate loss aggregates across all tokens:
\begin{equation}
\begin{aligned}
\mathcal{L}_{\theta}^{\text{GRPO}, \mathcal{D}}
&= \mathbb{E}_{x \sim \mathcal{D},\, \{y_i\}_{i=1}^G \sim \pi^{\text{gen}}_{\theta_{\text{old}}}(\cdot \mid x)}
\Bigg[
\frac{1}{\sum_{i=1}^G |y_i|}
\sum_{i=1}^G \sum_{t=1}^{|y_i|} \\
&\qquad
\min\!\left(
\rho_{i,t}(\theta)\,\hat{A}_i^{G},\;
\mathrm{clip}(\rho_{i,t}(\theta), \epsilon)\,\hat{A}_i^{G}
\right)
\Bigg]
\end{aligned}
\end{equation}

Each query $x$ requires modality subset $M_x \subseteq \mathcal{M} = \{\mathrm{V,A,S}\}$ (video/audio/subtitles), the minimal signals needed (e.g., $M_x=\{\mathrm{V,A}\}$ for audio-visual, $M_x=\{\mathrm{S}\}$ for subtitles-only). 
Let $R_M$ be rewards under requirement $M$.
Different $M$ yield heterogeneous $R_M$.

\paragraph{Modality-Unaware baseline.}
Ignores $M_x$, feeds full $(\mathrm{V,A,S})$ signals, and mixes all $R_M$ in batches $\mathcal{B} = \{(x_j,y_j,r_j)\}_{j=1}^B$.
Therefore the gradient estimate takes the standard form:
$$
\hat{g}_{\mathrm{MU}} = \frac{1}{B} \sum_{j=1}^B \nabla_\theta \log \pi_\theta(y_j \mid x_j) \hat{A}_j
$$
so rewards from heterogeneous $R_M$ are normalized together as if drawn from a single distribution.
This implicitly treats all modality regimes as homogeneous and inflates variance through between-subset differences in means and scales.

\paragraph{Modality-Aware scheme.}
A modality-aware scheme respects the requirement tags by forming per-subset (or compatible union) mini-batches
\[
\mathcal{B}_M = \{(x_j,y_j,r_j) : M_{x_j} = M\}
\]
and computing per-group advantages using only rewards from the same requirement set:
\[
\hat{g}_{\mathrm{MA}}
= \sum_{M} \frac{1}{|\mathcal{B}_M|} 
\sum_{(x_j,y_j,r_j)\in \mathcal{B}_M}
\nabla_\theta \log \pi_\theta(y_j \mid x_j)\,\hat{A}_j^{(M)}
\]
where each $\hat{A}_j^{(M)}$ is normalized within $R_M$. 
Intuitively, this stratified normalization removes between-subset reward variance and aligns gradient updates with the difficulty and noise characteristics of each modality regime.

Under standard assumptions (independent samples and bounded score function $\nabla_\theta \log \pi_\theta$), the variance of the modality-unaware estimator $\hat{g}_{\mathrm{MU}}$ decomposes into within-subset and between-subset components, whereas the modality-aware estimator $\hat{g}_{\mathrm{MA}}$ retains only the within-subset term.
As long as the reward distributions $R_M$ differ across modality requirements (non-zero between-subset variance), this yields
\[
\mathrm{Var}(\hat{g}_{\mathrm{MA}}) \leq \mathrm{Var}(\hat{g}_{\mathrm{MU}})
\]
suggesting faster and more stable convergence when optimization is conditioned on required modalities instead of mixing all regimes in a modality-unaware manner~\cite{mckinzie2023robustness,jiang2025rethinking}.

\subsection{MAPLE's Modality-Aware Policy Optimization}
\label{subsec:mapo_and_adaptive_variants}

Building on the modality-aware training foundation above, we establish the optimal strategy for policy optimization. 
With stratified batches $\mathcal{B}_M$, we systematically study four design axes---loss aggregation, clipping, sampling, and modality curriculum---each conditioned on RMTs $M_x$ and compared against the modality-unaware baseline.
We measure per-RMT accuracy, robustness gaps across complexity levels, and wall-clock convergence.

Baseline MAPO uses stratified GRPO:
\begin{equation}
\mathcal{L}^{\mathrm{MAPO}}_\theta = \sum_M \frac{1}{|\mathcal{B}_M|} \mathcal{L}^{\mathrm{GRPO}, \mathcal{B}_M}_\theta
\end{equation}
The modality-aware gradient scales with per-group advantage: $|\hat{g}_{\mathrm{MA}}| \propto |\hat{A}_j^{(M)}|$.
Harder queries produce systematically lower advantages (smaller gradients), while easier queries yield stronger signals.
To detect and compensate for this skew, we measure batch difficulty via KL divergence between empirical reward distribution $p_{\mathrm{emp}}$ and target $p_{\mathrm{tgt}} \sim \mathrm{Beta}(100,1)$ (mode 0.99):
$$
D_{\mathrm{KL}}(p_{\mathrm{emp}} \| p_{\mathrm{tgt}})
$$
Hard queries show $p_{\mathrm{emp}} \ll p_{\mathrm{tgt}}$ (large $D_{\mathrm{KL}}$); easy queries converge to zero.
We introduce per-RMT difficulty-adaptive weights $w_{M} \in (0,1)$:
\begin{equation}
\label{eq:total_loss}
\mathcal{L}^{\mathrm{MAPO}_{adp_w}}_\theta = \sum_{M \in \mathcal{M}} \frac{w_{M}}{|\mathcal{B}_M|} \mathcal{L}^{\mathrm{GRPO},\mathcal{B}_M}_\theta
\end{equation}

Each $w_{M}$ adapts via KL history $\mathcal{H}_{M}$ over window of size $\approx$ steps-per-epoch:
\begin{equation}
\label{eq:adaptive_weights}
z_{M} = \frac{D_{\mathrm{KL},M} - \mu_{\mathcal{H}_{M}}}{\sigma_{\mathcal{H}_{M}} + \varepsilon}, \quad w_{M} = \mathrm{sigmoid}(z_{M})
\end{equation}
This aggregates across tagged batches per training cycle for robust difficulty estimation.

Further, during each epoch we use KL statistics to arrange stratified batches by increasing complexity, counteracting gradient skew from reward-scale differences.
Harder batches produce smaller gradient updates, so when they are naively interleaved with easier batches that have larger gradient norms, their learning signals are effectively overshadowed.
We use KL statistics as dynamic curriculum signal, front-loading underperforming RMTs so the model learns their weaker signals first.

Formally, for each RMT $M$ we maintain KL history $\mathcal{H}_M$ and compute priority:
$$
s_M = \frac{1}{|W_M|} \sum_{k \in W_M} D_{\mathrm{KL},M}^{(k)}
$$
where $|W_M| = \min(L_W, |\mathcal{H}_M|)$ and $L_W=5$. Higher $s_M$ schedules harder tags earlier. For the first epoch---lacking KL history---we initialize via zero-shot performance, assigning higher priority to lower-accuracy tasks.
See Appendix~\ref{app:formal_multi_modal_rl} for complete MAPO pseudocode.

\section{Experiments \& Results}
\label{sec:experiments_results}

\subsection{Experimental Setup}
\label{subsec:setup}

We conduct all experiments using the Qwen2.5-Omni-3B~\cite{Qwen2.5-Omni} multimodal language model, which natively processes text, audio, and video inputs. 
Training uses a sequence length of 10,240 tokens (8,192 input + 2,048 response) and the veRL~\cite{sheng2025hybridflow} library for RL optimization.

\textbf{Training hyperparameters:} In our setup, we trained for multiple epochs over the
training prompts with AdamW optimizer~\cite{loshchilov2017decoupled} and a constant learning rate $2\times10^{-6}$. 
We used global batch size 256 (mini-batch 32, yielding 8 updates per rollout). 
Rollouts generate $G=8$ responses per prompt with temperature 1.0, top-$p=1.0$, within a 8,096-token context limit. 
All runs employ mixed precision with the FP32 language model head fix~\cite{chen2025minimax} to eliminate generator-trainer probability mismatches that destabilize importance sampling ratios.

Training runs were conducted on 4$\times$ NVIDIA H100-80GB nodes.
We first establish a strong modality-unaware baseline~(MUPO) that always receives full (V, A, S) signals, leveraging multimodal redundancy for robust task solving.
This follows standard post-training practice: token-level importance sampling and loss aggregation with symmetric clipping ($\pm0.2$), full-dataset on-policy updates, and per-group advantage normalization.
All experiments report pass@1 accuracy estimated from 5 samples per query, with other hyperparameter sweeps detailed in Appendix ~\ref{app_subsec:impl_details}.
On MAPLE-QA, this baseline achieves an average task performance of 58.58\%, while on MAPLE-Caption achieves 67.67\%.

\begin{table*}[h!]
\centering
\caption{Results of MAPLE's four design axes with MAPLE-QA on Qwen2.5-Omni-3B model, studying loss aggregation, clipping, dynamic sampling, and curriculum learning. 
Each row optimizes one axis atop modality-stratified MAPO while others follow standard settings. 
MAPLE's optimal recipe achieves average accuracy of 58.72\%, lowest modality gap (1.74\%), and fastest training (164.72~s/step).}
\small
\setlength{\tabcolsep}{4pt}
\renewcommand{\arraystretch}{1.0}
\resizebox{0.9\textwidth}{!}{
\begin{tabular}{llcccccccccccccl}
\toprule
\multirow{2}{*}{Method} & \multirow{2}{*}{Variant} 
& \multicolumn{8}{c}{\textbf{Modality Accuracy (Pass@1 \%)}} 
& \multicolumn{4}{c}{\textbf{Modality Gap (\%)}} 
& \multicolumn{2}{c}{\textbf{Efficiency}} \\
\cmidrule(lr){3-10} \cmidrule(lr){11-14} \cmidrule(lr){15-16}
&
& V & A & S & VA & VS & AS & VAS & \textbf{Avg} 
& U-B & U-T & B-T & \textbf{Avg} 
& \textbf{Time (s/step)} \\
\midrule
\multicolumn{2}{c}{\textit{\# Samples}} 
& 552 & 75 & 76 & 1145 & 1784 & 171 & 1198 & 5001 &   \\
\midrule

\multicolumn{2}{l}{Zero-shot} 
& 34.54 & 36.89 & 43.42 & 37.79 & 40.83 & 42.30 & 40.46 & \textbf{39.38} 
& 11.30 & 13.17 & 1.68 & \textbf{8.72} 
& {--} \\

\multicolumn{2}{l}{MUPO}  
& 55.08 & 65.34 & 63.82 & 57.47 & 60.15 & 58.77 & 58.14 & \textbf{58.58} 
& 3.43 & 1.79 & 1.59 & \textbf{2.27} 
& \textbf{523.28} \\
\midrule

\multicolumn{15}{c}{MAPO} \\
\midrule

\multirow{2}{*}{Loss aggregation}
& Token-level 
& 55.79 & 65.78 & 61.40 & 57.85 & 60.26 & 59.06 & 57.76 & \textbf{58.68} 
& 3.20 & 0.52 & 2.60 & \textbf{2.11} 
& \textbf{389.11} \\

& Sample-level 
& 55.74 & 65.78 & 62.28 & 58.69 & 60.42 & 58.87 & 57.48 & \textbf{58.86} 
& 3.79 & 0.07 & 3.72 & \textbf{2.53} 
& \textbf{395.03} \\
\cmidrule(lr){3-10} \cmidrule(lr){11-14} \cmidrule(lr){15-16}
\multirow{1}{*}{Clipping}

& Asymmetric
& 55.44 & 68.89 & 57.89 & 58.02 & 60.65 & 57.70 & 57.18 & \textbf{58.62} 
& 4.17 & 0.07 & 3.93 & \textbf{2.72} 
& \textbf{379.40} \\
\cmidrule(lr){3-10} \cmidrule(lr){11-14} \cmidrule(lr){15-16}

\multirow{2}{*}{Dynamic Sampling}
& Early filtering 
& 55.50 & 64.89 & 63.16 & 64.63 & 59.70 & 58.48 & 57.12 & \textbf{58.00} 
& 2.04 & 0.37 & 2.36 & \textbf{1.59} 
& \textbf{197.02} \\

& Mid-training 
& 55.32 & 69.33 & 61.84 & 57.29 & 60.18 & 59.65 & 57.10 & \textbf{58.39} 
& 2.71 & 0.73 & 3.35 & \textbf{2.26} 
& \textbf{265.13} \\
\cmidrule(lr){3-10} \cmidrule(lr){11-14} \cmidrule(lr){15-16}

\multirow{1}{*}{Curriculum learning}
& Modality-based 
& 56.40 & 65.78 & 60.53 & 58.05 & 60.83 & 59.65 & 57.96 & \textbf{59.05} 
& 3.27 & 0.19 & 2.98 & \textbf{2.15} 
& \textbf{424.03} \\
\cmidrule(lr){2-16}

& Full-recipe 
& 56.22 & 67.55 & 63.16 & 57.50 & 60.58 & 56.92 & 57.71 & \textbf{58.72} 
& 1.82 & 0.81 & 2.58 & \textbf{1.74} 
& \textbf{164.72} \\

\bottomrule
\end{tabular}
}

\label{tab:maple_ablations}
\end{table*}

\subsection{Optimal Training Strategy}
\label{sec:opt_trn_strat}

\begin{figure*}[t]
    \centering
    \begin{subfigure}{0.24\linewidth}
        \centering
        \includegraphics[width=\linewidth]{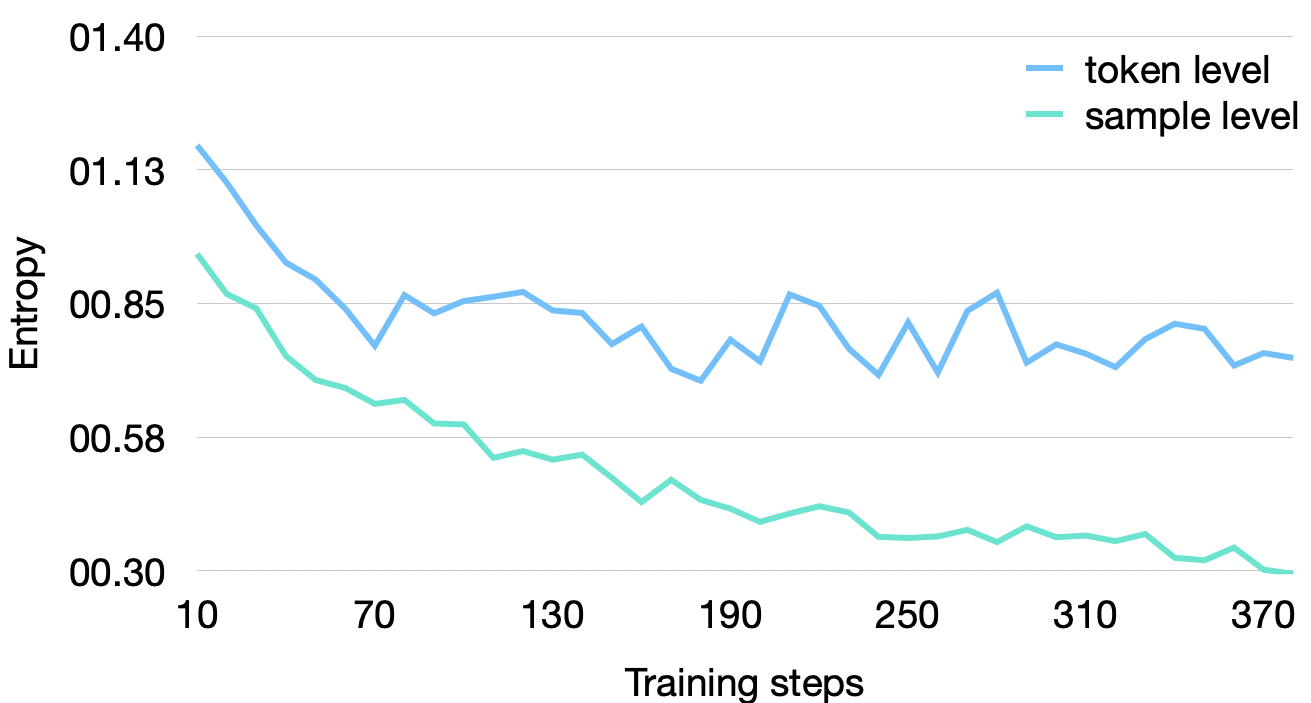}
        \caption{Loss aggregation}
        \label{fig:loss_agg}
    \end{subfigure}
    \hfill
    \begin{subfigure}{0.24\linewidth}
        \centering
        \includegraphics[width=\linewidth]{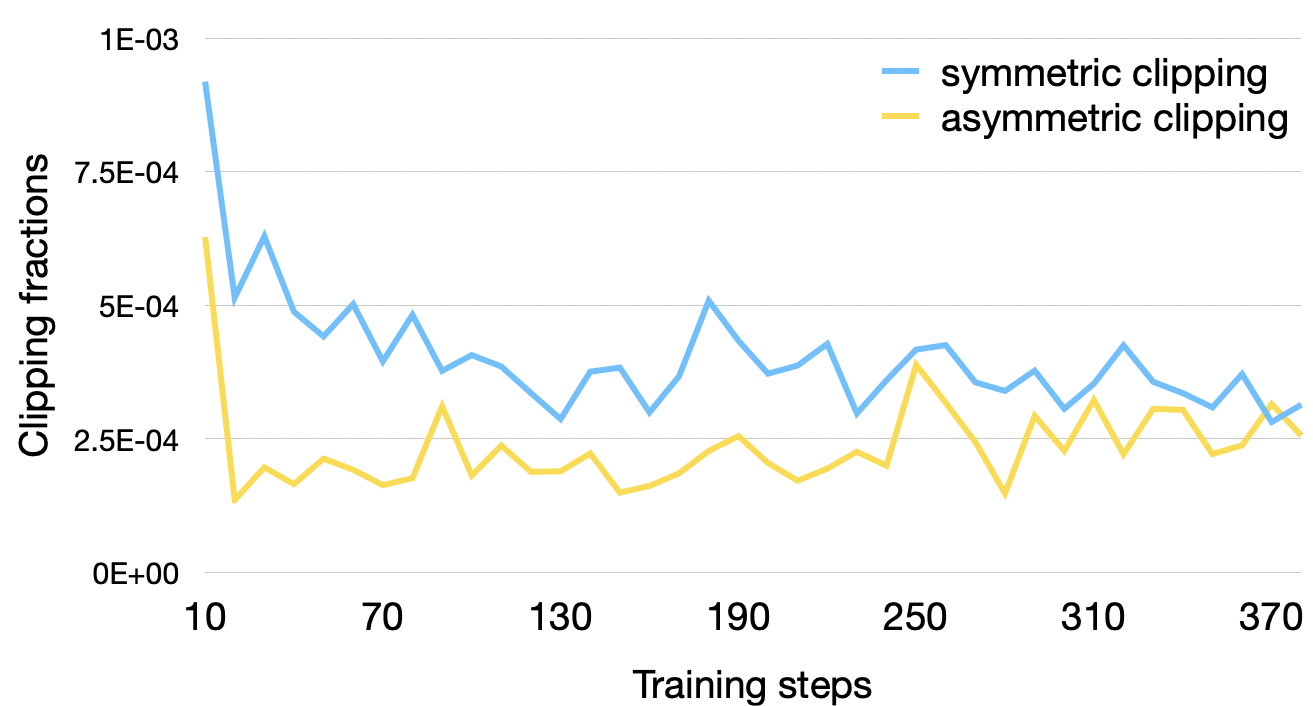}
        \caption{Clipping}
        \label{fig:clipping}
    \end{subfigure}
    \hfill
    \begin{subfigure}{0.24\linewidth}
        \centering
        \includegraphics[width=\linewidth]{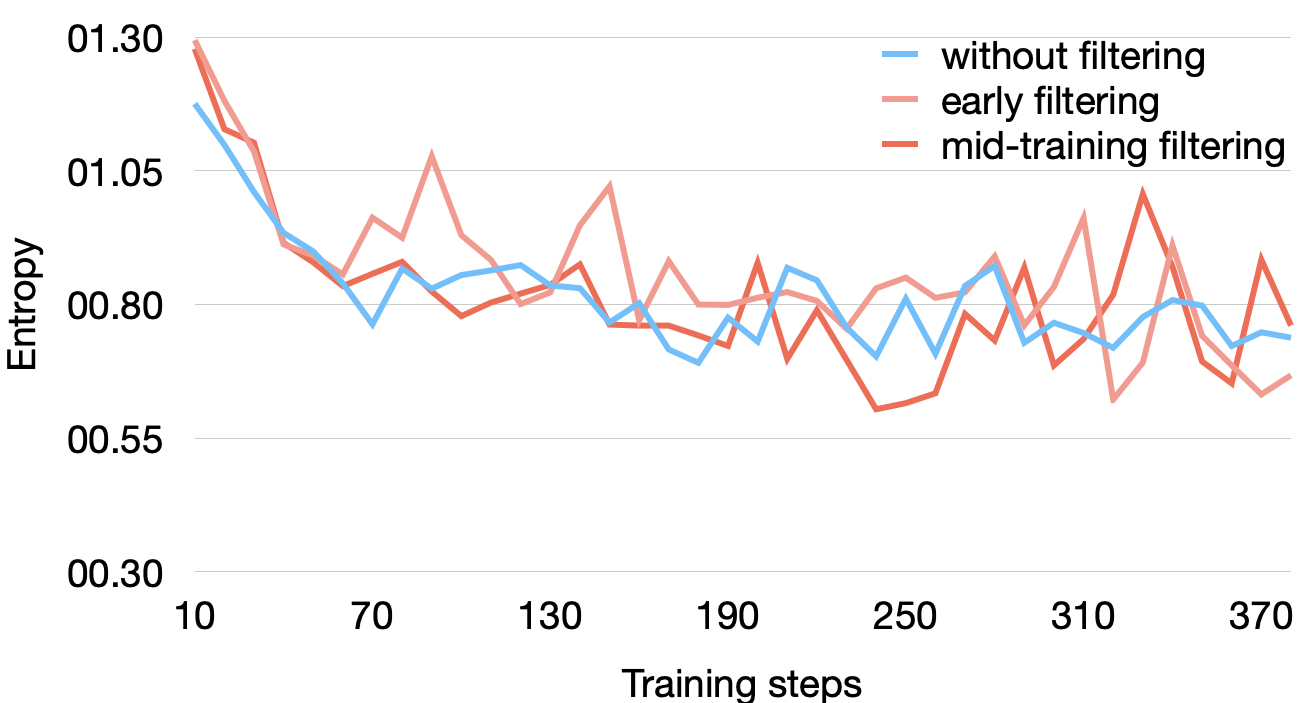}
        \caption{Sampling}
        \label{fig:sampling}
    \end{subfigure}
    \hfill
    \begin{subfigure}{0.24\linewidth}
        \centering
        \includegraphics[width=\linewidth]{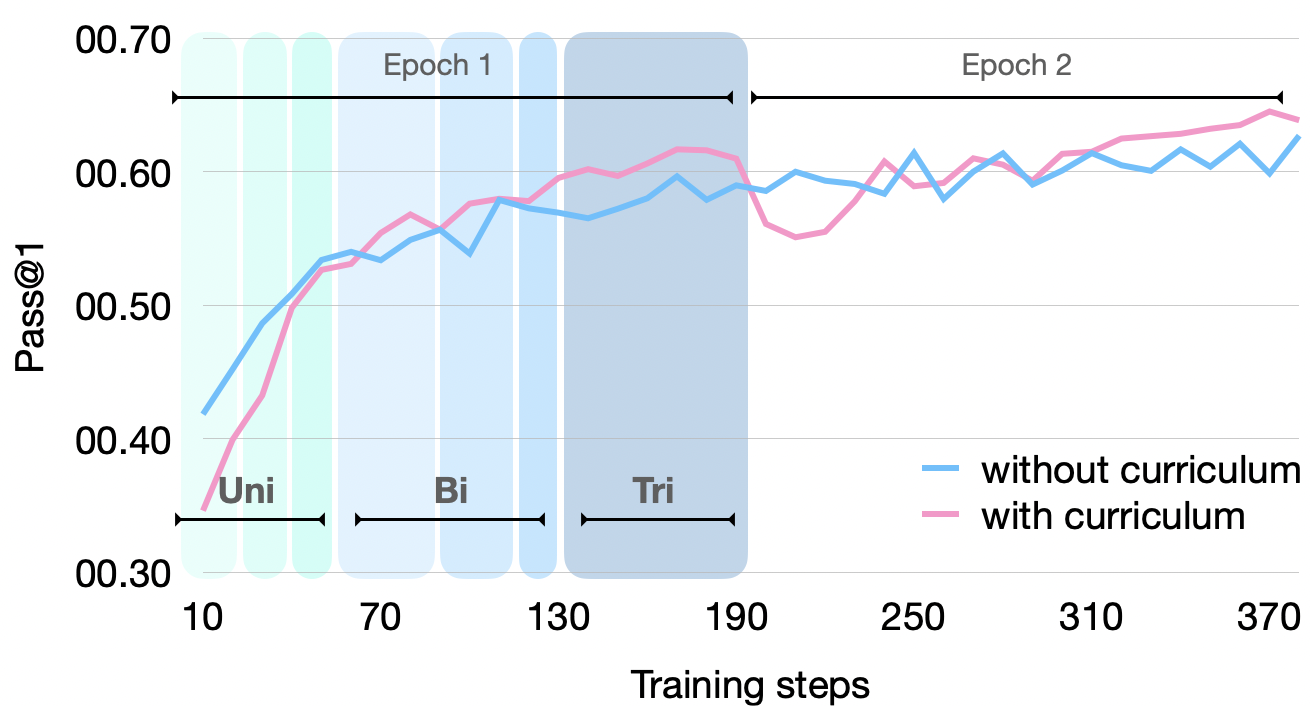}
        \caption{Curriculum}
        \label{fig:curriculum}
    \end{subfigure}
    
    \caption{MAPLE algorithmic variations across four design axes: (a) Sample-level loss aggregation outperforms others, (b) Asymmetric clipping maintains low clip fractions, (c) Dynamic filtering accelerates reward gains and (d) Curriculum improves accuracy by modality complexity.}
    \label{fig:maple_ablations}
\end{figure*}

In this section, we conduct experiments on MAPLE-QA due to its verifiable 4-option multiple-choice rewards, enabling precise algorithmic analysis without reward hacking concerns.

We first establish MAPO{'}s effectiveness using the same configuration as our baseline MUPO, differing only in modality-stratified batching $\mathcal{B}_M$. 
Despite receiving strictly limited signals matching each query's RMT, MAPO achieves 58.68\% pass@1, better than baseline performance while reducing policy gradient variance by 12.89\%.
Building on this foundation, we systematically optimize four algorithmic axes critical to multimodal RL stability.
To validate that these choices remain optimal when combined, we conduct
leave-one-out experiments.
Each variant improves accuracy, robustness (uni/multi-modal gaps), or training efficiency, with results shown in Table~\ref{tab:maple_ablations} and all training characteristic curves are mentioned in Appendix~\ref{app_subsec:impl_details}.

\textbf{Loss aggregation.} Loss aggregation controls credit assignment across heterogeneous signal regimes. We evaluate three strategies on MAPLE-QA: 
(1) \emph{token-level}~(average across all tokens in batch);
(2) \emph{sample-level}~(per-sequence mean, then batch-average); 
(3) \emph{prompt-level}~(fixed sum per sequence). 
Sample-level aggregation achieves 58.86\% pass@1, outperforming token-level (58.68\%). Token granularity dilutes per-query signal structure, while prompt-level causes divergence (Appendix~\ref{app_subsec:impl_details}). 
Sample-level preserves query-specific credit assignment while remaining computationally tractable.

\textbf{Asymmetric clipping.} Symmetric clipping ($\epsilon=0.2$) is a common choice for post-training with RL pipeline but it destabilizes multimodal training due to varying information content across signal combinations. 
We adopt asymmetric thresholds $\epsilon^+=0.3$, $\epsilon^-=0.2$, expanding the upper trust region for positive advantages while maintaining conservative negative updates. 
This reduces clip fractions by 71.65\% as shown in Fig.~\ref{fig:clipping}, yielding healthier entropy and smoother policy losses compared to symmetric clipping.

\textbf{Dynamic sampling.} Some prompts produce identical rewards across all $G=8$ generations (zero-variance samples) that contribute nothing to policy gradients ($\nabla_\theta \log\pi_\theta \hat{A}_i^G=0$). 
Baseline training includes them, artificially inflating effective batch size while diluting signal. 
Early filtering at batch construction accelerates wall-clock training by 1.98$\times$ (58.00\% accuracy) due to reduced computation per step. 
Mid-training filtering~(training on full dataset for 1 epoch, then filtering) recovers accuracy to 58.39\% but slows training by 0.35$\times$ relative to early filtering.
Applying any filtering has nearly no change in entropy reduction as shown in Fig.~\ref{fig:sampling}.

\textbf{Curriculum learning.} We sequence training dataset by signal complexity: uni-modal $\to$ bi-modal $\to$ tri-modal queries. 
Curriculum training produces cleaner gradient magnitudes and monotonically increasing rollout rewards compared to mixed stratified training (Fig.~\ref{fig:curriculum}). 
Observing first epoch, uni-modal batches show gradual reward gains under curriculum ordering, while bi- and tri-modal batches accelerate sharply, outpacing non-curriculum settings. 
This pattern persists into epoch 2, confirming effective policy adaptation to increasing signal complexity. 
Final accuracy reaches 59.05\%, validating that structured progression prevents overfitting to simple signal patterns.

To validate our optimizations, leave-one-out experiments revert each axis individually to MUPO baseline while keeping others active, showing every component contributes positively with no destructive interactions.
From these analyses, we consolidate the optimal configuration, our complete \textbf{Full-recipe}, using sample-level loss aggregation, asymmetric clipping with a wider upper trust region, early zero-variance filtering at batch construction to prioritize speed, and curriculum learning that sequences training by increasing signal complexity.
Compared to the modality-unaware baseline, this recipe achieves 58.72\% accuracy, the lowest modality gap of 1.74\%, and most efficient gains~(164.72 secs per training step) that is 3.18$\times$ faster than MUPO.
Training characteristic curves for each setting appear in Appendix~\ref{app_subsec:impl_details}.

\subsection{Adaptive Training Strategy}

\begin{table}[t]
\centering
\caption{MAPLE-QA results on the Qwen2.5-Omni-3B model. 
The table reports modality-wise accuracy for MAPO adaptive strategies. 
The configuration with reweighting and adaptive curriculum ($adp_w + adp_{\text{cur}}$) achieves the highest average accuracy (59.82\%).
}
\small
\setlength{\tabcolsep}{5pt}
\renewcommand{\arraystretch}{1.0}
\resizebox{1.0\columnwidth}{!}{
\begin{tabular}{lcccccccc}
\toprule
\multirow{2}{*}{Method} 
& \multicolumn{8}{c}{\textbf{Modality Accuracy (Pass@1 \%)}} \\
\cmidrule(lr){2-9} 
& V & A & S & VA & VS & AS & VAS & \textbf{Avg} \\
\midrule
Zero-shot
& 34.03 & 33.60 & 42.16 & 38.58 & 41.40 & 43.45 & 40.86 & \textbf{39.78} \\
MUPO 
& 55.08 & 65.34 & 63.82 & 57.47 & 60.15 & 58.77 & 58.14 & \textbf{58.58} \\
\cmidrule(lr){2-9}
MAPO
& 55.79 & 65.78 & 61.40 & 57.85 & 60.26 & 59.06 & 57.76 & \textbf{58.68} \\

\hspace{2mm}+ ${adp_w}$
& 55.34 & 67.47 & 62.16 & 58.28 & 60.77 & 58.57 & 57.96 & \textbf{58.98} \\

\hspace{2mm}+ ${cur}$
& 56.40 & 65.78 & 60.53 & 58.05 & 60.83 & 59.65 & 57.96 & \textbf{59.05} \\

\hspace{2mm}+ ${adp_{\text{cur}}}$
& 56.87 & 65.33 & 61.89 & 58.62 & 61.00 & 59.52 & 58.27 & \textbf{59.38} \\

\hspace{2mm}+ ${adp_w + cur}$
& 57.24 & 65.87 & 59.46 & 59.37 & 61.63 & 58.81 & 58.13 & \textbf{59.73} \\

\hspace{2mm}+ ${adp_w + adp_{\text{cur}}}$
& 57.13 & 66.40 & 61.35 & 59.16 & 61.64 & 58.33 & 58.67 & \textbf{59.82} \\
\bottomrule
\end{tabular}
}

\label{tab:adaptive_results_qa}
\end{table}

\begin{table}[t]
\centering
\caption{MAPLE-Caption results on the Qwen2.5-Omni-3B model. 
The table reports modality-wise accuracy for MAPO adaptive strategies. 
The configuration with reweighting and adaptive curriculum ($adp_w + adp_{\text{cur}}$) achieves the highest average accuracy (74.00\%).
}
\small
\setlength{\tabcolsep}{5pt}
\renewcommand{\arraystretch}{1.0}
\resizebox{1.0\columnwidth}{!}{
\begin{tabular}{lcccccccc}
\toprule
\multirow{2}{*}{Method} 
& \multicolumn{8}{c}{\textbf{Modality Accuracy (LLM-as-Judge score)}} \\
\cmidrule(lr){2-9} 
& V & A  & S  & VA  & VS  & AS  & VAS& \textbf{Avg} \\
\midrule
Zero-shot
& 56.41 & 63.25 & 61.98 & 59.51 & 59.10 & 63.63 & 58.38 & \textbf{60.32} \\

MUPO
& 63.78 & 73.69 & 68.20 & 64.46 & 65.69 & 71.71 & 65.18 & \textbf{67.67} \\
\cmidrule(lr){2-9}

MAPO 
& 66.78 & 83.46 & 87.50 & 66.44 & 67.99 & 78.59 & 66.40 & \textbf{73.88} \\

\hspace{2mm}+ ${adp_w}$
& 65.65 & 82.39 & 85.82 & 65.89 & 67.72 & 77.57 & 65.18 & \textbf{72.89} \\

\hspace{2mm}+ ${cur}$
& 65.03 & 81.65 & 86.39 & 64.71 & 66.21 & 76.83 & 63.67 & \textbf{72.07} \\

\hspace{2mm}+ ${adp_{\text{cur}}}$
& 65.11 & 79.89 & 86.14 & 65.12 & 66.94 & 76.26 & 64.17 & \textbf{71.95} \\

\hspace{2mm}+ ${adp_w + cur}$
& 66.43 & 82.91 & 85.57 & 66.30 & 68.48 & 78.23 & 66.38 & \textbf{73.47} \\

\hspace{2mm}+ ${adp_w + adp_{\text{cur}}}$
& 66.20 & 81.42 & 84.97 & 66.98 & 69.10 & 81.71 & 67.58 & \textbf{74.00} \\
\bottomrule
\end{tabular}
}
\label{tab:adaptive_results_caption}
\end{table}

\begin{figure}[t]
    \centering
    \begin{subfigure}{0.45\columnwidth}
        \centering
        \includegraphics[width=\columnwidth]{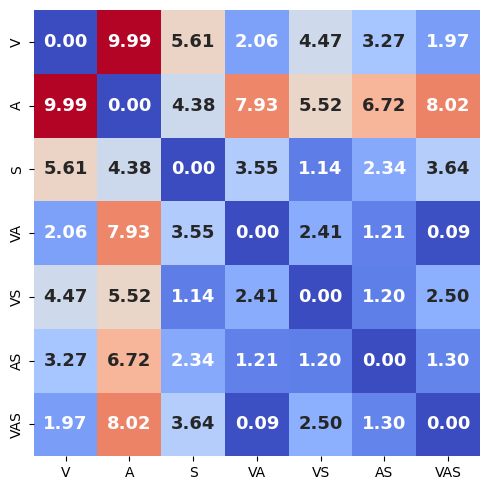}
        \caption{MAPO}
        \label{fig:qa_mapo}
    \end{subfigure}
    \hfill
    \begin{subfigure}{0.45\columnwidth}
        \centering
        \includegraphics[width=\columnwidth]{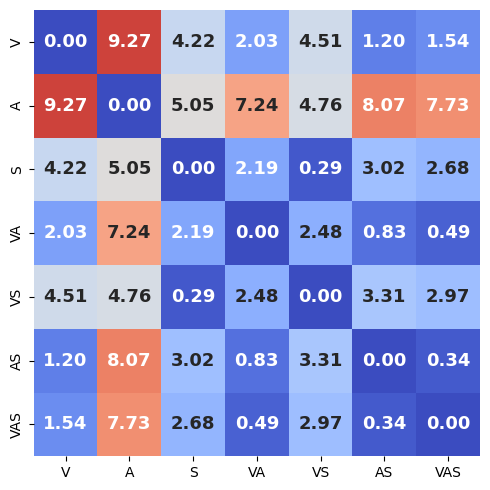}
        \caption{${adp_w + adp_{\text{cur}}}$}
        \label{fig:qa_mapo_adp}
    \end{subfigure}
    
    \caption{
    \textbf{Tag-wise disparity heatmaps across MAPLE-QA-eval.} 
    {Left:} Basic MAPO reduces cross-tag variance. 
    {Right:} Full adaptive MAPO ($adp_w + adp_{\text{cur}}$) achieves minimal disparity (smaller values better). }
    \label{fig:heatmap_qa}
\end{figure}

\begin{figure}[t]
    \centering
    \begin{subfigure}{0.45\columnwidth}
        \centering
        \includegraphics[width=\columnwidth]{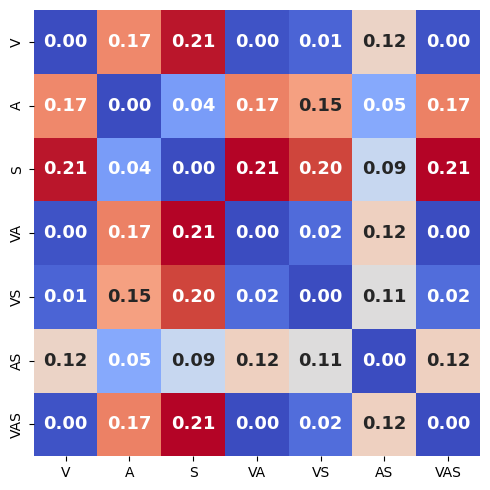}
        \caption{MAPO}
        \label{fig:caption_mapo}
    \end{subfigure}
    \hfill
    \begin{subfigure}{0.45\columnwidth}
        \centering
        \includegraphics[width=\columnwidth]{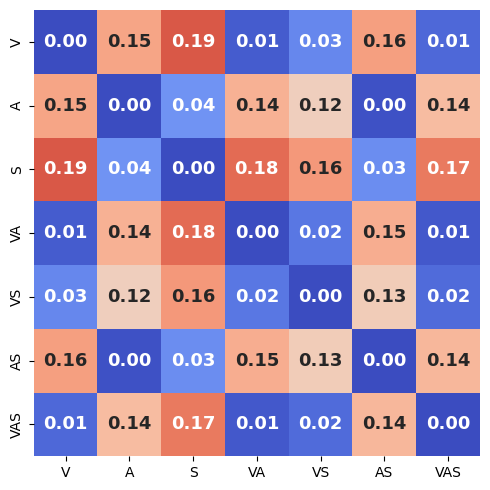}
        \caption{{\text{${adp_w + adp_{\text{cur}}}$}}}
        \label{fig:caption_mapo_adp}
    \end{subfigure}
    
    \caption{
    \textbf{Tag-wise disparity heatmaps across MAPLE-Caption-eval.} 
    {Left:} Basic MAPO reduces cross-tag variance. 
    {Right:} Full adaptive MAPO ($adp_w + adp_{\text{cur}}$) achieves minimal disparity (smaller values better). }
    \label{fig:heatmap_caption}
\end{figure}

Having established the optimal static training configuration, we next explore adaptive strategies to further boost accuracy by addressing gradient skewness observed in multimodal RL.
As detailed in Sec.~\ref{subsec:mapo_and_adaptive_variants}, we evaluate two adaptation techniques: 
(1) {${adp_w}$}: batch weighting by historical KL divergence between empirical and target reward distributions;
(2) {${adp_{\text{cur}}}$}: KL-based dynamic curriculum adaptation.
Since our static curriculum already delivers the strongest gains on MAPLE-QA [Table~\ref{tab:maple_ablations}], we include it~(${cur}$) for fair comparison.
We also test combinations: {${adp_w + cur}$} (weighting + static curriculum) and {${adp_w + adp_{\text{cur}}}$} (weighting + dynamic curriculum).
Results across both benchmarks from Sec.~\ref{sec:datasets} appear in Tables~\ref{tab:adaptive_results_qa} and \ref{tab:adaptive_results_caption}.

For QA with discrete rewards, MAPO baseline achieves 58.68\% average accuracy, improving to 59.82\% with adaptive weighting and dynamic curriculum (${adp_w + adp_{cur}}$). 
Hard modalities gain most (V: 55.79\%~$\to$~57.13\%; VA: 57.85\%~$\to$~59.16\%) while easy text tags remain comparable, showing reweighting corrects scale imbalances and adaptive curriculum controls update timing. 
Captioning also shows better performance with continuous rewards: from MAPO{'}s 73.88\% baseline, it reaches 74.00\% while lifting heavy tags (VS: 67.99\%~$\to$~69.10\%; AS: 78.59\%~$\to$~81.71\%; VAS: 66.40\%~$\to$~67.58\%), confirming both scale (reweighting) and order (adaptive curriculum) corrections counteract easy-hard skew.

Our full adaptive MAPO strategy consistently outperforms all variants per-sample, achieving the highest average accuracy across both benchmarks.
This combination of reweighting and adaptive curriculum secures top performance across modality subsets and demonstrates superior stability over basic MAPO.
This dominance beyond aggregate averages, establishes the full adaptive recipe for heterogeneous multimodal training.
MAPLE's full adaptive strategy substantially enhances multimodal fusion ability, with 30.24\% of MAPLE-Caption samples showing multi-modal captions outperforming the best uni-modal baseline, the 18.19\% achieved by basic MAPO. 
Heatmap analysis further confirms reduced cross-modality disparities: 
adaptive strategies match or exceed base MAPO{'}s variance reduction on QA (Fig.~\ref{fig:heatmap_qa}), while captioning also exhibits the same pattern. 
Together, these metrics demonstrate that adaptive MAPO optimally balances overall accuracy, tag-wise robustness, per-sample consistency, and genuine multimodal integration by addressing both \emph{how much} (reweighting) and \emph{when} (adaptive curriculum) to update under heterogeneous signal distributions.

%
%

\subsection{Ablations}

\subsubsection{Data Augmentation}

\begin{table}[t]
\centering
\caption{MAPLE-QA+ ablation: MAPO trained across modality-exact, -superset, and -deficit configurations. 
Model generalizes beyond fixed RMT patterns, correctly abstaining via ``None'' option when evidence is insufficient.
}

\small
\setlength{\tabcolsep}{4pt}
\renewcommand{\arraystretch}{1.0}
\resizebox{1.0\columnwidth}{!}{
\begin{tabular}{lcccccccc}
\toprule
\multirow{2}{*}{Training dataset} 
& \multicolumn{8}{c}{\textbf{Modality Accuracy (Pass@1 \%)}} \\
\cmidrule(lr){2-9} 
& V & A & S & VA & VS & AS & VAS & \textbf{Avg} \\
\midrule
Zero-shot
&56.26&52.19&52.12&56.05&51.46&50.34&47.94&\textbf{52.19} \\
\cmidrule(lr){2-9}
MAPLE-QA
&62.58&61.52&57.72&54.93&53.99&44.90&40.91 & \textbf{51.90}  \\

MAPLE-QA+
&72.68&78.67&72.20&79.72&78.31&74.23&74.98 &		\textbf{76.99} \\
\bottomrule
\end{tabular}
}
\label{tab:maple_qa_extra_results}
\end{table}

To investigate whether MAPO simply overfits to the exact modality patterns seen during training or truly learns modality-aware reasoning, we construct an augmented QA dataset that perturbs the mapping between queries and their available modalities. 
Instead of always supplying only the modalities specified as the required modalities~(RMT), we generate three strata for each query: 
1) modality-exact (information exactly matches the RMT specification), 
2) modality-superset (additional modalities are provided beyond those indicated by the RMT), and
3) modality-deficit (one or more RMT-specified modalities are removed). 
We refer to this augmented dataset as MAPLE-QA+, which contains 137{,}313 QA pairs with 103,265 and 34,048 samples as a train–test split, partitioned across the three strata. 
To further discourage superficial pattern matching, we also introduce a ``None'' option. 
For 25\% of modality-exact and modality-superset examples, the original correct answer is replaced by this new option.
Also for all modality-deficit cases the ``None'' option is treated as correct.

Table~\ref{tab:maple_qa_extra_results} shows MAPO on MAPLE-QA+ achieving 77.0\% average accuracy across all modality combinations vs. modality-exact MAPLE-QA with largest gains on mixed/incomplete settings. 
QA's discrete rewards isolate true modality-awareness from scoring noise. 
This confirms MAPO learns robust reasoning under missing/redundant signals, correctly abstaining via ``None'' when evidence is insufficient.


\subsubsection{Contrastive Training}
\begin{figure}
    \centering
    \includegraphics[width=0.9\columnwidth]{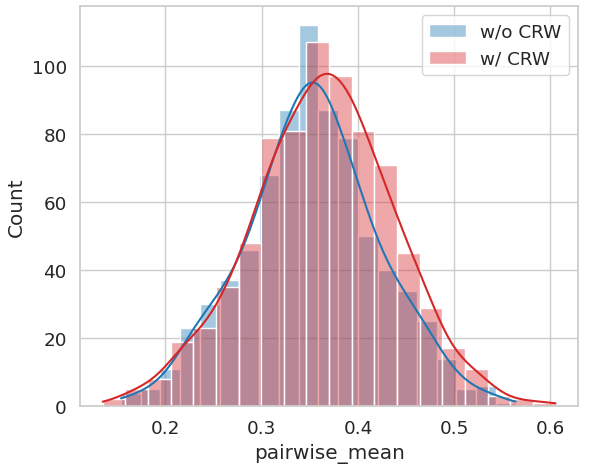}
    \caption{Intra-group dispersion histogram showing a rightward shift of the distribution under CRW.}
    \label{fig:crw_distance}
\end{figure}

Despite MAPO's gains, captioning policies still suffer from \emph{mode collapse}: high-reward captions remain nearly identical regardless of input modality (VAS vs. VA vs. V-only), failing to reflect available signals. 
We address this through {Contrastive Reward Weighting (CRW)}, which upweights high-reward ``positives'' from complete modalities only when they maintain embedding separation from deliberately-generated ``negatives'' under modality-deficit conditions for the same prompt (Appendix~\ref{app:crw}). 
CRW is applied exclusively to captioning, where continuous $[0,1]$ rewards enable rich contrastive gradients unlike QA's binary signals.
CRW improves {fusion ability} from 18.19\% to 18.46\% (multi-modal $>$ best uni-modal captions), intra-group dispersion from 0.35 to 0.36 (greater caption diversity within each RMT), and inter-group separation from 0.41 to 0.46 (clearer modality-conditioned manifolds). 
Pairwise distance histograms shift rightward and t-SNE projections show more distinct modality clusters (Figs.~\ref{fig:crw_distance},~\ref{fig:crw_tsne}), confirming CRW simultaneously boosts within-condition diversity and cross-condition separation for truly modality-aware generation.

\section*{Conclusion}

MAPLE addresses a fundamental oversight in multimodal RL post-training: treating heterogeneous signal combinations as homogeneous training data. 
Through MAPLE-bench, MAPO, and adaptive strategies, we achieve highest performance under realistic reduced-signal conditions, while converging faster and reducing uni- to multi-modal gaps.
This work reveals that modality-stratified optimization eliminates between-group variance, a pervasive yet unrecognized source of instability in omni-modal training. 
MAPLE constitutes a complete, production-ready recipe that will enable robust multimodal RL deployment across diverse real-world signal configurations, establishing modality-aware post-training as essential infrastructure for next-generation multimodal reasoning systems.

\section*{Impact Statement}

This paper advances the field of machine learning through MAPLE, a modality-aware RL post-training ecosystem that improves efficiency, stability, and robustness for multimodal language models. 
MAPLE enables more reliable deployment of multimodal reasoning systems under realistic partial-signal conditions commonly encountered in production environments.
While our technical contributions focus on benchmark dataset creation and algorithmic improvements, we note potential downstream applications in accessibility (e.g., robust audio-visual processing for assistive technologies) and content analysis across diverse signal availability. 
We do not foresee specific ethical concerns beyond standard responsible AI practices for improved model performance, and no negative societal consequences are anticipated from these optimization techniques.

\bibliography{refs}
\bibliographystyle{icml2026}

\newpage
\appendix
\onecolumn

\section{Extended Discussion of Related Literature}
\label{app:related}

\subsection{Multimodal LLM Architectures}

Recent MLLMs follow two dominant architectural paradigms: \emph{deep fusion} via cross-attention within LLM layers (Type-A/B) versus \emph{early fusion} at the input stage (Type-C/D). 

\textbf{Deep fusion} approaches (Type-A) employ standard cross-attention, as in early VideoLLaMA~2~\cite{videollama2}, 
while Type-B uses custom layers like mPLUG-Owl2's modality-adaptive attention~\cite{ye2024mplug}. 
\textbf{Early fusion} dominates recent work: Type-C uses non-tokenized modality encoders (e.g., CLIP-ViT + Whisper feeding directly into LLaMA~\cite{video-salmonn}), while Type-D applies discrete tokenization across modalities (e.g., Qwen2.5-Omni's TMRoPE for synchronized audio-visual streaming~\cite{qwen25-omni-arxiv}).

Industry systems emphasize end-to-end omni-modality: Gemini-3 offers native multimodal I/O with developer controls for reasoning depth~\cite{team2024gemini};
Mini-Omni2 implements staged vision-to-speech alignment~\cite{mini-omni2}; 
and Phi-4-Multimodal uses Mixture-of-LoRAs for parameter-efficient vision-speech augmentation~\cite{phi4-multimodal}. 
Despite architectural sophistication, all assume static, fully-available modality combinations during post-training~\cite{yin2024mllm-survey}.

\subsection{Detailed Benchmark Comparison}
\begin{table*}[h!]
\centering
\caption{Comparison of multimodal benchmarks.
I: Image, V: Video, A: Audio, T: Text.
\textbf{Cond.\ Output} indicates whether valid QA answers or captions vary
given identical VAT instances under different modality combinations.}
\label{tab:benchmark_comparison}
\setlength{\tabcolsep}{5pt}  

\begin{tabular}{l l l l c}
\hline
\textbf{Benchmark} &
\textbf{Modality} &
\textbf{Task} &
\textbf{Size (\#Q)} &
\textbf{Cond.\ Output} \\
\hline
MME & I, T & QA (Y/N) & -- & No \\
MMBench & I, T & QA & 2.9k & No \\
MM-Vet (v2) & I, T & Free-form & 0.5k & No \\
SEED-Bench-2 & I / V, T & QA & 24k & No \\
OmniBench & I, A, T & QA & 1.14k & No \\
MuAViC & A, V, T & Free-form & 1.1M & No \\
AVHBench & A, V, T & QA, Free-form & 5.3k , 1.1k & No \\
MAVERIX & A, V, T & QA & 2.5k & No \\
OmniVideoBench & A, V, T & QA & 1k & No \\
JointAVBench & A, V, T & QA & 2.8k & No \\
\textbf{MAPLE-bench (ours)} & A, V, T & QA, Free-form & $\approx 5$k, $\approx 5$k & Yes \\
\hline
\end{tabular}
\vspace{-2mm}
\end{table*}
Existing benchmarks reveal two critical limitations: 
(i) \emph{fixed ground truth} despite varying inputs, and 
(ii) lack of explicit \emph{modality-deficit awareness}. 
For example, OmniBench~\cite{li2025omnibenchfutureuniversalomnilanguage} varies I/A/T inputs but maintains identical QA options/answers; 
OmniVideoBench~\cite{li2025omnivideobenchaudiovisualunderstandingevaluation} offers A/V-only ablations under static problem definitions. 
Neither distinguishes tasks unsolvable due to true information gaps from fusion failures, nor adapts caption references to available signals.

MAPLE-bench addresses this through RMTs and combination-conditioned outputs, enabling fine-grained diagnosis of modality dependence.

\subsection{Post-training Method Taxonomy}

Post-training has evolved from PPO-based RLHF~\cite{schulman2017proximal} toward value-model-free methods:

\begin{itemize}
    \item \textbf{GRPO family}: Group Relative Policy Optimization~\cite{shao2024deepseekmath} eliminates critics via multi-sample group normalization; Dr.~GRPO~\cite{liu2025drgrpo} and DAPO~\cite{yu2025dapo} refine clipping and aggregation.
    \item \textbf{Scale-focused}: ScaleRL~\cite{khatri2025art} drops KL regularization for 100k+ GPU scaling; GDPO~\cite{liu2026gdpo} decouples the normalization of individual rewards, more faithfully preserving their relative differences 
    \item \textbf{Unimodal ablations}: Recent analysis~\cite{park2025clip, liu2025part} dissects loss aggregation and asymmetric clipping on reasoning benchmarks but excludes multimodal data.
\end{itemize}

Modality-specific strategies exist but remain disconnected from core RL optimization: 
MBQ~\cite{li2025mbq} balances signal contributions pre-RLHF; 
IntraCurriculum~\cite{penas2023curriculum} sequences task complexity without per-query signal tagging. 
No prior work conditions batch composition, reward normalization, or clipping on minimal required signals---the core innovation of MAPLE's MAPO.

\subsection{Robustness and Modality Gap Evidence}

Recent analyses reveal three fundamental limitations of modality-agnostic training that MAPLE directly addresses:

\paragraph{1. Train-Test Modality Mismatch}
Omni-modal training often underperforms uni-modal specialists under deployment shifts. 
\citet{mckinzie2023robustness} documents large gaps between average and worst-case performance across modality configurations, with follow-ups~\cite{robustness-multimodal-learning} confirming additional modalities can be under-utilized or even harmful without targeted interventions.
\citet{jiang2025specific} similarly shows omni-MLLMs trail specialists on per-modality benchmarks due to mixed-alignment interference.

\paragraph{2. Modality Imbalance and Dominance} 
Strong modalities systematically dominate weaker ones during training. 
\citet{jiang2025rethinking} demonstrates this causes lagging performance even when all signals are present, while explicit metrics reveal average vs.\ worst-case gaps over train-test setups.
MBQ~\cite{li2025mbq} confirms symmetric treatment (quantization, optimization) degrades performance---modality-balanced allocation is essential.
CLIP-Refine~\cite{yamaguchi2025post} shows persistent vision-language ``modality gaps'' post-training under naive omni approaches.

\paragraph{3. Optimization Instability Across Configurations}
Even advanced post-training exhibits uneven gains. Safe multimodal surveys~\cite{zhao2024survey} highlight vulnerability to imbalance; unsupervised GRPO-style methods~\cite{wei2025unsupervised} yield inconsistent improvements despite significant compute.
OmniEval~\cite{zhang2025omnieval} confirms wide performance variance across modality combinations despite strong aggregate scores.

MAPLE resolves these through per-query Required Modality Tags: MAPLE-bench provides tagged supervision, MAPO conditions batching/normalization/clipping on minimal signal subsets, and curriculum learning sequences by signal sparsity.

\section{Dataset generation details}
\label{app:dataset_generation}

\subsection{MAPLE-QA task}
We performed several refinement and validation steps to ensure high-quality, modality-dependent supervision.
Within the generation pipeline outlined in Fig~\ref{fig:dataset_pipeline}, all captions and annotations undergo cross-modal consistency checks. 
Respective captions are then temporally aligned across audio and visual streams to maintain coherent event ordering and sub-title generation.
Our dataset generation pipeline follows the automated annotation and optimization framework of Daily-Omni~\cite{zhou2025dailyomni}. 
Specifically, we revised the Advanced QA system prompt as described below~\ref{app:qa_prompts}, while following the remaining configuration from Daily-Omni~\cite{zhou2025dailyomni}.

Using the aligned subtitles, we generate QA pairs with multiple tag label variants, subsequently refined with state-of-the-art models (e.g., GPT-4o, GPT-5~\cite{hurst2024gpt}) to improve linguistic clarity and logical consistency. 
This process ensures that each query–response instance accurately reflects its corresponding modality cues.
To guarantee faithful modality attribution, we apply an automatic filtering stage in which only QA items solvable exclusively under their assigned Required Modality Tags~((RMTs)) are retained.
Multi-model voting is used for this filtering—questions that can be answered without the required modalities are discarded.
Despite leveraging multiple high-performance models, automatic TAG labeling remains imperfect.

For both the MAPLE-bench subsets, human reviewers inspect and revise the test portion, with approximately {74.2\%} (for MAPLE-QA) of TAG assignments updated after manual validation. 
This highlights the current limitations of state-of-the-art models in accurately understanding and attributing modality-specific reasoning.
These additional refinement steps close several observed “generation gaps,” addressing inconsistencies in modality correlation, annotation drift, and task solvability. 
They collectively ensure that MAPLE-bench provides rigorous, modality-grounded supervision for evaluating multimodal post-training.

\subsection{MAPLE-QA+ task}
To further assess whether MAPO genuinely acquires modality-aware reasoning rather than memorizing modality–query correspondences, we extend the MAPLE-QA dataset into an augmented version termed \textbf{MAPLE-QA+}. This extension deliberately perturbs the alignment between each question and its accessible modalities, thereby testing the model’s robustness under incomplete or redundant sensory conditions.

\vspace{0.5em}
\noindent
For every QA instance, we construct three distinct strata:
\begin{enumerate}
    \item \textbf{Modality-Exact:} The available modalities precisely match the Required Modality Tags (RMTs).
    \item \textbf{Modality-Superset:} Additional modalities beyond those specified by the RMTs are included.
    \item \textbf{Modality-Deficit:} One or more RMT-specified modalities are intentionally omitted.
\end{enumerate}

\vspace{0.5em}
\noindent
The resulting MAPLE-QA+ dataset comprises a total of \textbf{137{,}313} QA pairs, partitioned into \textbf{103{,}265} for training and \textbf{34{,}048} for testing, evenly distributed across the three modality strata. To discourage superficial pattern recognition and encourage genuine modality reasoning, we introduce a set of ``\textit{None-answer}'' options that represent cases where the provided modalities are insufficient to yield a valid response.

\vspace{0.5em}
\noindent
Specifically, for all \textbf{modality-deficit} samples, the correct answer is replaced with one of the predefined \textit{none-answer} statements, compelling the model to recognize when information is incomplete. In addition, for \textbf{25\%} of the \textbf{modality-exact} and \textbf{modality-superset} examples, the original incorrect answer is substituted with a randomly selected \textit{none-answer} to further regularize the model’s confidence calibration. The training set includes \textbf{100} distinct \textit{none-answer} variants, each expressing the notion of insufficient multimodal evidence in slightly different linguistic forms.

\vspace{0.5em}
\noindent
To construct the MAPLE-QA+ dataset, we first expanded the human-verified MAPLE-QA test set of \textbf{4,864} samples across all modality combinations, resulting in a total of \textbf{34,048} test samples. For the training set, to balance training efficiency and the ratio between \textit{None-answer} and valid answer choices, we did not expand all seven modality combinations. Instead, we included the all original MAPLE-QA training set of \textbf{47,893} samples as the \textbf{Modality-Exact} subset, and added \textbf{31,635} \textbf{Modality-Superset} and \textbf{23,737} \textbf{Modality-Deficit} samples. The 25\% substitution ratio for \textit{None-answer} replacement in the Modality-Exact and Modality-Superset strata is summarized in Table~\ref{tab:none_ratio}.

\begin{table}[h!]
\centering
\caption{Number of \textit{None-answer} substitutions in \textbf{MAPLE-QA+} Train set by strata.}
\label{tab:none_ratio}
\begin{tabular}{l r}
\toprule
\textbf{Strata} & \textbf{None-answer Count} \\
\midrule
Modality-Exact & 10,969 \\
Modality-Superset & 7,316 \\
Modality-Deficit & 23,614 \\
\bottomrule
\end{tabular}
\end{table}

\vspace{0.5em}
\noindent
Below is the complete list of \textit{none-answer} candidates used in MAPLE-QA+:
\begin{tcolorbox}[
  breakable,
  colback=gray!5,
  colframe=gray!40,
  title=Tag-specific QA generation Prompt,
  listing only,
  listing options={
    basicstyle=\ttfamily\small,
    breaklines=true,
    columns=fullflexible
  }
]
{\small
\begin{itemize}[noitemsep, topsep=0pt, leftmargin=1.5em]
    \item Cannot determine the answer from the given information.
    \item The provided data is insufficient to infer a valid response.
    \item Unable to answer with the current set of modalities.
    \item The available inputs do not contain enough evidence.
    \item Not enough multimodal context is provided.
    \item The input lacks sufficient cross-modal information.
    \item Missing modality prevents accurate reasoning.
    \item Unable to infer due to incomplete sensory input.
    \item The visual, audio, or speech data is inadequate.
    \item Cannot make a reliable decision with missing modalities.
    \item The input lacks necessary multimodal cues.
    \item Insufficient perceptual evidence to determine the answer.
    \item The question cannot be resolved with the current data.
    \item Incomplete sensory input prevents a valid conclusion.
    \item The available modalities do not support a confident response.
    \item Unable to conclude due to missing sensory information.
    \item The input data is too limited to determine an answer.
    \item Cannot infer without complete multimodal evidence.
    \item The response cannot be generated with partial modalities.
    \item Missing visual or speech information restricts reasoning.
    \item The given input lacks essential modality signals.
    \item Unable to decide due to insufficient multimodal evidence.
    \item The data provided is incomplete for this task.
    \item Cannot produce an answer without all required modalities.
    \item The current input does not provide enough context.
    \item Missing modality information leads to uncertainty.
    \item The answer cannot be derived from partial inputs.
    \item Insufficient multimodal alignment to determine output.
    \item The available data does not support a clear conclusion.
    \item Unable to respond accurately with missing modality data.
    \item The input lacks critical sensory components.
    \item Cannot determine the answer without full modality coverage.
    \item Incomplete multimodal input prevents reasoning.
    \item The provided information is too sparse to answer.
    \item Missing vision or audio data limits understanding.
    \item The question requires more complete sensory input.
    \item Unable to infer due to lack of multimodal coherence.
    \item The current modalities do not provide enough evidence.
    \item Cannot answer confidently with incomplete data.
    \item The input is missing key perceptual information.
    \item Insufficient modality coverage for reliable inference.
    \item The available inputs are inadequate for this question.
    \item Missing sensory data prevents accurate interpretation.
    \item Cannot determine outcome with partial modality input.
    \item The multimodal data is incomplete for reasoning.
    \item Unable to conclude due to missing perceptual signals.
    \item The input lacks sufficient cross-modal information.
    \item Not enough multimodal evidence to support an answer.
    \item The provided modalities are insufficient for inference.
    \item Cannot generate a valid response with missing data.
    \item Incomplete modality input leads to uncertainty.
    \item The question cannot be answered with current inputs.
    \item Missing modality context prevents understanding.
    \item The available sensory data is too limited.
    \item Unable to reason effectively with incomplete modalities.
    \item The input does not contain enough multimodal cues.
    \item Cannot infer due to lack of modality completeness.
    \item The data lacks necessary cross-modal features.
    \item Insufficient perceptual input to determine the answer.
    \item The current modalities do not provide enough clarity.
    \item Missing modality signals hinder reasoning.
    \item The input is incomplete for multimodal inference.
    \item Cannot produce a confident answer with missing inputs.
    \item The available data lacks multimodal consistency.
    \item Incomplete sensory evidence prevents decision-making.
    \item The question requires additional modality input.
    \item Unable to determine due to missing visual or audio data.
    \item The input lacks sufficient multimodal representation.
    \item Cannot answer without complete sensory information.
    \item The provided modalities are incomplete for reasoning.
    \item Insufficient multimodal context to generate an answer.
    \item The data is too limited to support a conclusion.
    \item Missing modality features prevent accurate inference.
    \item The input does not provide enough perceptual evidence.
    \item Unable to respond due to incomplete modality coverage.
    \item The available inputs are insufficient for reasoning.
    \item Cannot determine the answer with partial sensory data.
    \item The multimodal input is incomplete for this task.
    \item Missing modality information limits interpretability.
    \item The question cannot be resolved with current modalities.
    \item Insufficient sensory evidence to produce an answer.
    \item The input lacks necessary multimodal alignment.
    \item Unable to infer due to missing perceptual context.
    \item The data provided is not enough for reliable reasoning.
    \item Cannot generate an answer with incomplete modalities.
    \item The available sensory inputs are too limited.
    \item Missing modality data prevents accurate response.
    \item The input lacks full multimodal representation.
    \item Insufficient modality information to conclude.
    \item The question cannot be answered with partial data.
    \item Unable to reason due to incomplete sensory input.
    \item The provided modalities do not cover all required aspects.
    \item Cannot determine outcome with missing multimodal evidence.
    \item The input data is incomplete for this inference.
    \item Missing modality signals reduce confidence in response.
    \item The available information is insufficient for reasoning.
    \item Unable to produce an answer due to missing modalities.
    \item The multimodal input lacks completeness for inference.
    \item Cannot answer accurately with partial modality data.
    \item The input does not contain enough multimodal evidence.
\end{itemize}}
\end{tcolorbox}

\noindent
By exposing the model to all modality configurations and explicitly teaching it to abstain when evidence is lacking, MAPLE-QA+ acts as a strong regularizer. It encourages the model to calibrate its confidence based on information sufficiency, thereby reducing hallucinations and improving factual grounding under realistic multimodal conditions.

\subsection{Tag-specific QA generation Prompt}
\label{app:qa_prompts}

\begin{tcolorbox}[
  breakable,
  colback=gray!5,
  colframe=gray!40,
  title=Tag-specific QA generation Prompt,
  listing only,
  listing options={
    basicstyle=\ttfamily\small,
    breaklines=true,
    columns=fullflexible
  }
]
\#\#\# Task:

Given a detailed description that summarizes the visual and audio(sound event sequence and speech) content of a video and a series of audio and visual events that occurs at the same time, generate question-answer pairs that based on the description to help human better understand the video.

\#\#\# Guidelines For Question-Answer Pairs Generation:

- The QAs you generate should be answered with and only with BOTH audio(human speech and object sound) and visual information.

- When generating choices, make sure the choices are equally long so that there won't be data bias.

- To increase the difficulty, generate answer choices with some ambiguity or confusion, ensuring that it requires careful attention to both visual and audio elements to answer correctly.

- Make the answer choices more deceptive so that multiple options appear plausible, rather than having only one or two clearly reasonable choices. This will increase the question's difficulty.

- For all types of questions, don't explictly mention too much information in the question otherwise the question can be answered correctly without using information from the video.

- For every type of question, you need to write a QA for each Tag Types (Modality-specific QA generation). For example, you need to create a QA for each type of V/A/S/VS/VA/AS/VAS for the reasoning type, and for the rest of the question types, make a question for all combinations as well.

- Please tag each modality combination carefully. For example, if you can answer with just vision information without needing any other modality information, you should tag it with "V", if you can answer with sound information only, "A", and "S" if you answer with speech information only. In the same way, the rest of the combinations "VA", "VS", "AS", and "VAS" can be answered with just that modality. Please verify and tag whether you need that modality.\\

\#\#\# Question Types:

You should generate the following types of questions:

- AV Event Alignment: Formulate questions to determine which audio and visual events occurred simultaneously with each other.

- Event Sequence: Formulate questions to determine the temporal sequence of visual and audio events in the video.

- Reasoning: Formulate questions to explain the cause or reason behind the occurrence of a visual or audio event in the video.

- Inference: Formulate questions to speculate on information not explicitly presented in the video.

- Comparative: Formulate questions to compare the similarity or difference between the audio and visual information of two or more events in the video.

- Context understanding: Formulate questions to determine the contextual information surrounding a specific event in the video.\\

\#\#\# Tag Types (Modality-specific QA generation):

When generating each Question-Answer pair, assign one of the following tags to indicate which modalities are considered:

- 'V': Use only the visual information from the video. Do not use speech or other audio information.

- 'A': Use only the non-speech audio information from the video. Do not use speech or visual information.

- 'S': Use only the speech information from the video. Do not use other audio or visual information.

- 'VA': Use both visual and non-speech audio information. Do not use speech information.

- 'VS': Use both visual and speech information. Do not use non-speech audio information.

- 'AS': Use both non-speech audio and speech information. Do not use visual information.

- 'VAS': Use all available information from the video (visual, non-speech audio, and speech).\\

\#\#\# Output Format:

The questions should be in the form of multiple choice. The choices should look like:["A. Choice 1", "B. Choice 2", "C. Choice 3", "D. Choice 4"]

The answer should be a single capital letter A, B, C, or D.

Your output should be formed in a JSON file.

You response should look like:

\textasciigrave{}\textasciigrave{}\textasciigrave{}json

[\{``Type": \textless type-1\textgreater, ``Question": \textless question-1\textgreater, ``Choice": \textless choice-1\textgreater, ``Answer": \textless answer-1\textgreater, ``Explaination": \textless explanation-1\textgreater, ``tag": \textless tag-1\textgreater\},
\{``Type": \textless type-2\textgreater, ``Question": \textless question-2\textgreater, ``Choice": \textless choice-2\textgreater, ``Answer": \textless answer-2\textgreater, ``Explaination": \textless explanation-2\textgreater, ``tag": \textless tag-2\textgreater\},
...]
\end{tcolorbox}

\subsection{Benchmark Example:MAPLE-QA+}
\label{app:caption_example}

\begin{tcolorbox}[
    colback=gray!4,
    colframe=gray!40,
    title=\textbf{MAPLE-QA+ VA Question Example Instance},
    fonttitle=\small,
    boxrule=0.6pt,
    arc=2mm,
    left=1.2mm,
    right=1.2mm,
    top=1.2mm,
    bottom=1.2mm
]
\noindent\textbf{Seed image (from video \texttt{-spqqACI6UQ.mp4}).}\par
\vspace{0.4em}

{\centering
\includegraphics[width=0.55\linewidth]{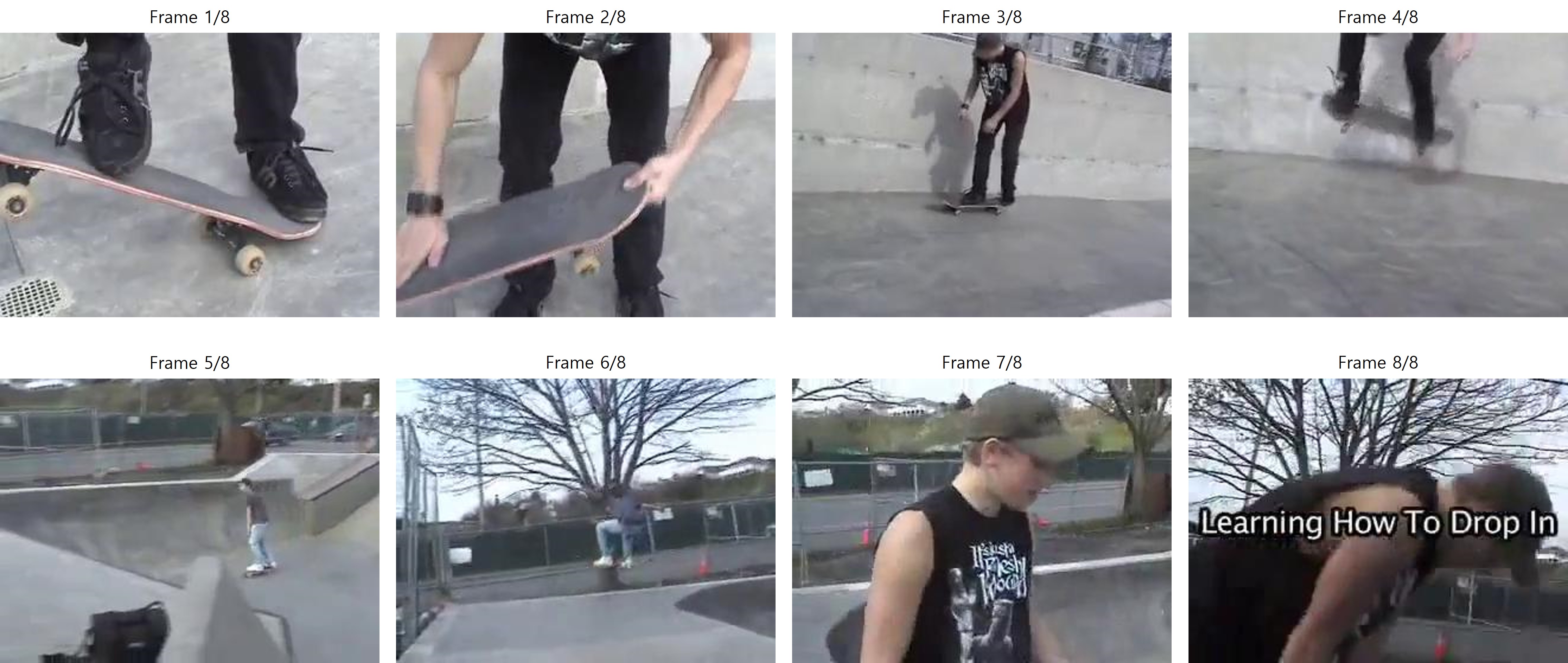}\par
}

\vspace{0.6em}

\noindent\textbf{Subtitle: 0-10s: In this section, a man is speaking with an instructional tone. He says: ``And as it's coming up around here, then you push down with your feet to come back down again.'' 10-20s: In this section, there is the sound of a skateboard hitting the ground. Then, there are generic sound effects. Finally, there are more sound effects. 20-30s: In this section, a male speaker is instructing someone in a casual tone. Then, the male speaker says, ``Now we're going to drop in. Okay. So, you leave your foot on the tail, leave it up.''}

\{Question\}

\{Choices\}

The final option MUST BE put in [].\\

\noindent\textbf{Question:}

What sequence of non-speech audio events closely corresponds to the second skateboarder's aerial trick and subsequent landing on his board?\\

\noindent\textbf{Modality-Exact \& Expansion Choices VA / VAS:} 

A. A bang followed by a metallic clank, then rolling sounds continuing for several seconds.

\textbf{B. A thud, a rapid whoosh, a sharp thwack, a second whoosh, and a final heavy thud.}

C. A prolonged scraping sound followed by rising impact noises and a clicking sound.

D. Continuous grinding sounds with intermittent clunking and faint wind-like noises.\\

\par
\vspace{0.25em}

\noindent\textbf{Modality-Deficit Choices V / A / S / VS / AS:} 

A. A bang followed by a metallic clank, then rolling sounds continuing for several seconds.

\textbf{B. Cannot answer based on the provided information.}

C. A prolonged scraping sound followed by rising impact noises and a clicking sound.

D. Continuous grinding sounds with intermittent clunking and faint wind-like noises.\par

\vspace{0.25em}

\end{tcolorbox}

\subsection{MAPLE-Caption task}
\label{app:caption_prompts}
We generate modality-tagged caption instances from seed videos sampled from the VAST-Omni corpora~\cite{chen2023vast},
using Gemini-2.5-Flash~\cite{team2024gemini} under a modality-isolated protocol that controls which modalities are exposed to the generator.
We provide the exact prompts used to generate caption data for each modality tag.
Our prompt consists of (i) a shared base instruction, (ii) a tag-specific question template,
and (iii) a common ending instruction that enforces a unified, modality-agnostic sentence.

\vspace{8pt}
\noindent\textbf{Shared Base Prompt}
\begin{tcolorbox}[
  breakable,
  colback=gray!5,
  colframe=gray!40,
  title=Base Prompt,
  listing only,
  before skip=2pt,  
  after skip=2pt,   
  listing options={
    basicstyle=\ttfamily\small,
    breaklines=true,
    columns=fullflexible
  }
]
Given the video, there are visual, audio, and subtitle information.
The audio can be divided into sound and speech information, and
the audio information should cover both sound and speech aspect.
\end{tcolorbox}

\vspace{8pt}
\noindent\textbf{Tag-specific Question Templates}
The following templates are used depending on the available modality tag:
$\mathrm{V}$ (visual only),
$\mathrm{A}$ (audio only),
$\mathrm{S}$ (subtitle only),
$\mathrm{VA}$,
$\mathrm{VS}$,
$\mathrm{AS}$,
and $\mathrm{VAS}$.

\begin{tcolorbox}[
  breakable,
  colback=gray!5,
  colframe=gray!40,
  title=Question Templates by Modality Tag,
  listing only,
  before skip=2pt,  
  after skip=2pt,   
  listing options={
    basicstyle=\ttfamily\small,
    breaklines=true,
    columns=fullflexible
  }
]
[V]
Following a video without an audio, make a context understanding
video description in one sentence.
The answer format should be follow '\#\#\# Video Description' format. \\

[A]
Following an audio, make a context understanding video description
in one sentence. The audio is a part of video.
The description should include both speech and background sound.
The answer format should be follow '\#\#\# Video Description' format. \\

[S]
For given subtitles, make a context understanding video description
in one sentence. Only refer to subtitle.
The answer format should be follow '\#\#\# Video Description' format.
subtitle: \{subtitle\} \\

[VA]
Following a video, make an unified context understanding
video description in one sentence.
The description should refer to both visual and audio evidences
(speech and background sound).
The answer format should be follow '\#\#\# Video Description' format. \\

[VS]
Following a video, make an unified context understanding
video description in one sentence, referring to both
visual evidences and subtitle.
Do not guess anything about sound.
The answer format should be follow '\#\#\# Video Description' format.
subtitle: \{subtitle\} \\

[AS]
Following an audio, make an unified context understanding
video description in one sentence.
The audio is a part of video.
The description should refer to both audio evidences and subtitle.
The answer format should be follow '\#\#\# Video Description' format.
subtitle: \{subtitle\} \\

[VAS]
Following a video, make an unified context understanding
video description in one sentence, referring to all of the
visual, audio evidences (speech and background sound),
and subtitle.
The answer format should be follow '\#\#\# Video Description' format.
subtitle: \{subtitle\}

\end{tcolorbox}

\vspace{8pt}
\noindent\textbf{Common Ending Instruction}

\begin{tcolorbox}[
  breakable,
  colback=gray!5,
  colframe=gray!40,
  title=End Prompt,
  listing only,
  before skip=2pt,  
  after skip=2pt,   
  listing options={
    basicstyle=\ttfamily\small,
    breaklines=true,
    columns=fullflexible
  }
]
Please combine the given information to create a natural sentence
without mentioning modality, ensuring that the sentence should be
well-aligned and unified given modalities information.
\end{tcolorbox}

\vspace{8pt}
\noindent\textbf{Final Prompt Composition.}
For each instance, we concatenate \texttt{Base Prompt} + (tag-specific question template) + \texttt{End Prompt}.
For tags that include subtitles ($\mathrm{S}$, $\mathrm{VS}$, $\mathrm{AS}$, $\mathrm{VAS}$),
we replace \texttt{\{subtitle\}} with the corresponding subtitle text.

\subsection{Benchmark Example:MAPLE-Caption}
\label{app:caption_example}

\begin{tcolorbox}[
    colback=gray!4,
    colframe=gray!40,
    title=\textbf{MAPLE-Caption Example Instance},
    fonttitle=\small,
    boxrule=0.6pt,
    arc=2mm,
    left=1.2mm,
    right=1.2mm,
    top=1.2mm,
    bottom=1.2mm
]
\noindent\textbf{Seed image (from video \texttt{A\_4TRaRiLRw.10.mp4}).}\par
\vspace{0.4em}

{\centering
\includegraphics[width=0.55\linewidth]{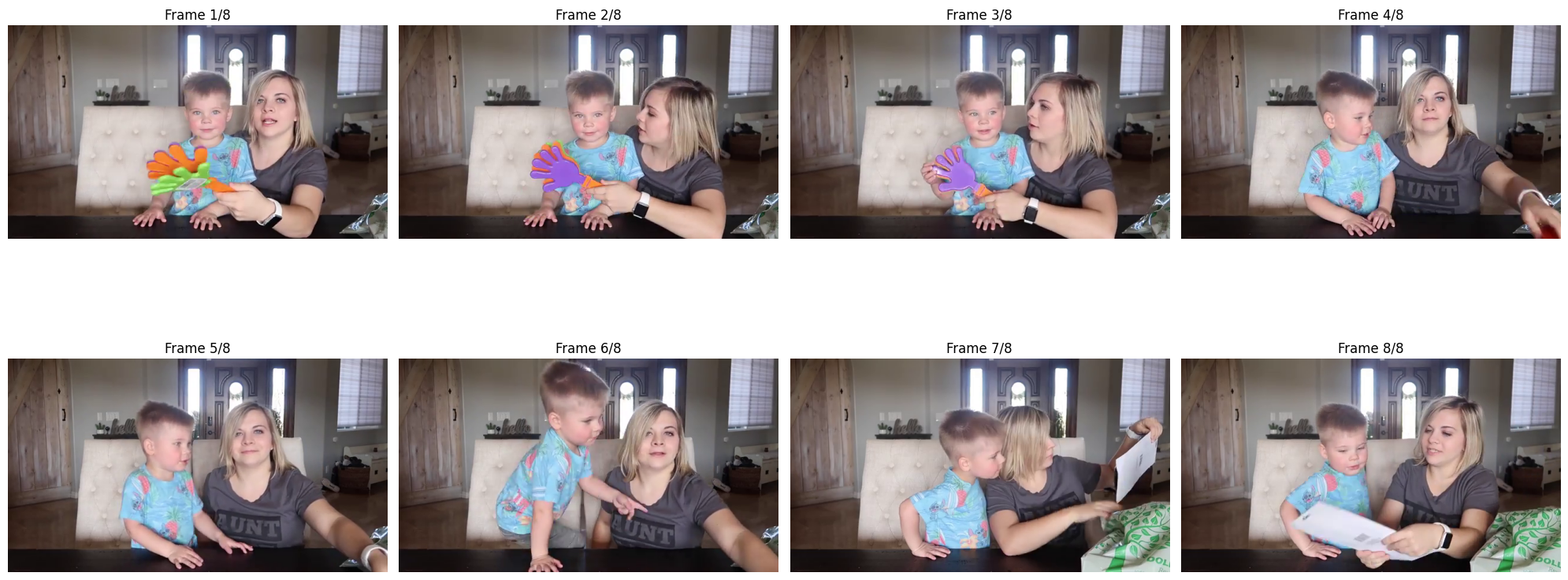}\par
}

\vspace{0.6em}

\noindent\textbf{Subtitle: So yes, you like this yeah. Alright, you like it.}

\vspace{0.4em}
\noindent\textbf{Modality-tagged ground-truth captions}\par
\vspace{0.3em}

\noindent\textbf{V:} A woman and young boy with a colorful clapper toy, before she opens a green bag to reveal an item inside.\par
\vspace{0.25em}

\noindent\textbf{A:} A female adult asking a child if he like something, as the child enthusiastically confirms he likes it amidst the rustling sounds of objects being handled, with low-volume background music.\par
\vspace{0.25em}

\noindent\textbf{S:} Someone confirms with another person that they like something.\par
\vspace{0.25em}

\noindent\textbf{VA:} A woman and a young boy are seen at a table, discussing a colorful clapper toy as the boy enthusiastically confirms he likes it, before the woman pulls an item from a green bag, accompanied by low-volume background music.\par
\vspace{0.25em}

\noindent\textbf{VS:} A woman shows a colorful hand clapper toy to a young boy, confirming his liking for it as she puts the toy away and then retrieves a paper from a green bag.\par
\vspace{0.25em}

\noindent\textbf{AS:} An adult asks a child if he like something and the child gives an enthusiastic ``Yep, yep'' in response, amidst the rustling sounds of objects being handled, with low-volume background music.\par
\vspace{0.25em}

\noindent\textbf{VAS:} A woman holding a clapper toy that makes clapping sounds asks a young boy sitting next to her if he likes it, and after he enthusiastically agrees, she reaches into a green bag to pull out another item, accompanied by low-volume background music.

\end{tcolorbox}

\section{Detailed Evaluation Protocol }
\label{app:evaluation}

\subsection{Modality Gap Computation}

For a model $M$ evaluated across all RMTs, we partition queries by complexity:

\begin{align*}
\mathcal{U} &= \{\text{V, A, S}\} & \bar{A}_\text{Uni} &= \frac{1}{|\mathcal{U}|} \sum_{r \in \mathcal{U}} \text{Acc}(M,r) \\
\mathcal{B} &= \{\text{VA, VS, AS}\} & \bar{A}_\text{Bi} &= \frac{1}{|\mathcal{B}|} \sum_{r \in \mathcal{B}} \text{Acc}(M,r) \\
\mathcal{T} &= \{\text{VAS}\} & \bar{A}_\text{Tri} &= \text{Acc}(M,\text{VAS})
\end{align*}

The Modality Gap~\cite{xiang2025understanding} captures average relative improvement:
$$
\Delta_\text{gap} = \frac{1}{3} \left( \left| \frac{\bar{A}_\text{Bi}-\bar{A}_\text{Uni}}{\bar{A}_\text{Uni}} \right| + \left| \frac{\bar{A}_\text{Tri}-\bar{A}_\text{Uni}}{\bar{A}_\text{Uni}} \right| + \left| \frac{\bar{A}_\text{Tri}-\bar{A}_\text{Bi}}{\bar{A}_\text{Bi}} \right| \right)
$$

\subsection{LLM-as-Judge Scoring (J)}
\label{sec:llm_judge}

We use an LLM-as-Judge with modality-conditioned \emph{textual references}
to detect hallucination errors and modality-fusion in the captioning task:
$$
J \in [0,1] \quad \text{(per RMT configuration)}.
$$

The Judge assigns three axis scores per sample:
\emph{modality missing} ($s_{\text{miss}}$),
\emph{hallucination} ($s_{\text{hall}}$),
and \emph{fusion quality} ($s_{\text{fus}}$),
each taking a value in $[0,1]$.
The final captioning reward is computed as their average:
$$
R_{\text{cap}} = \frac{1}{3}
\left(s_{\text{miss}} + s_{\text{hall}} + s_{\text{fus}}\right)
\in [0,1].
$$

For precise multimodal assessment, we adopt a \emph{ref-combine} protocol.
Specifically, the Judge is provided with a reference bundle
$\mathcal{R}_{\text{ref}} =
\{\text{GT}_{\text{tag}}, \text{Ref}_{\mathrm{V}},
\text{Ref}_{\mathrm{A}}, \text{Ref}_{\mathrm{S}}\}$,
where all elements are given as \emph{textual descriptions}:
the ground-truth caption corresponding to the current RMT,
together with uni-modal reference captions for visual, audio, and subtitle streams.
The Judge evaluates the generated caption $\hat{y}$ against these references
using an explicit scoring rubric.
We used GPT-4o~\cite{hurst2024gpt} as the Judge model.

\paragraph{Judge Prompt Template.}
We use the following text-only prompt template for GPT-4o as the captioning Judge.
All references (uni-modal captions) are provided in \emph{text} form,
and only those references whose modalities are contained in the current modality tag
are included in the reference bundle.

\begin{tcolorbox}[
  breakable,
  colback=gray!5,
  colframe=gray!40,
  title=Judge Prompt Template,
  listing only,
  listing options={
    basicstyle=\ttfamily\small,
    breaklines=true,
    columns=fullflexible
  }
]
You are given:\\

Ground Truth (GT) caption describing the video accurately.

Predicted caption generated by fusing information from one or more modalities (vision, audio, subtitle).

Separate captions for each modality (vision-only caption, audio-only caption, subtitle-only caption).

** If there is only one modality, the GT caption and the separate captions for each modality are the same, so only one is provided.\\

Your task:
Evaluate the predicted video caption by comparing it directly to the GT caption, while also using the modality captions to strengthen detection of hallucinations and assess fusion quality.\\

Scoring criteria:\\

[1. Missing Information] (Compare Predicted vs GT)\\

1.0 → No missing information at all.

0.8 → Minor omission, meaning unaffected.

0.6 → Moderate omission, some important details missing.

0.4 → Significant omission, core meaning partially lost.

0.2 → Severe omission, most important details missing.

0.0 → Almost all information missing.\\

[2. Errors / Hallucinations] (Check Predicted against GT and modality captions)

1.0 → No errors or hallucinations.

0.8 → Minor wording inaccuracy, meaning preserved.

0.6 → Moderate factual error, possible meaning distortion.

0.4 → Significant error, meaning largely affected.

0.2 → Severe error, multiple incorrect details.

0.0 → Completely incorrect description.\\

[3. Modality Fusion Accuracy] (Check if Predicted integrates all modality info into one coherent sentence)\\

1.0 → Perfect integration of all modalities into one coherent sentence.

0.8 → Mostly coherent fusion, very small detail underrepresented.

0.6 → Fusion partially successful; some modality info feels separate.

0.4 → Significant fusion issues; modalities mentioned in isolation.

0.2 → Severe fusion failure; modalities not meaningfully combined.

0.0 → No fusion at all.\\

Output format (MUST follow exactly, and DO NOT provide any explanation or reasoning):\\

Missing Info score: $\alpha$ score$\alpha $

Error score: $\beta $score$\beta$

Fusion score: $\gamma$ score$\gamma$ \\

Given Information as follows:

Generated caption: \{pred\}

GT caption: $\{gt_fused\}$

Vision caption:$ \{Ref_V\}$

Audio caption: $\{Ref_A\}$

Subtitle caption: $\{Ref_S\}$

\end{tcolorbox}

\subsection{Fusion Gain}

Out of total $N$ samples, count the fraction of samples where multi-modal captions outperform best uni-modal caption.
$$
\text{uni-modal} = \max\{\text{J(V)}, \text{J(A)}, \text{J(S)}\}
$$
$$
\text{multi-modal} = \max\{\text{J(VA)}, \text{J(VS)}, \text{J(AS)}, \text{J(VAS)}\}
$$
$$
\text{Fusion Gain} = \frac{|\{\text{samples: multi-modal $>$ uni-modal}\}|}{N}
$$

\section{RL formalization: Distributions and Training Schemes for MLM}
\label{app:formal_multi_modal_rl}

\subsection{General RL Baseline}

In this section, we explain the reinforcement learning from trajectories where prompts $x \sim \mathcal{D}$ distribution trigger completions $y \sim \pi_{\theta}(\cdot \mid x)$.
The distributed setup splits GPUs between \emph{generators} (high-throughput rollouts via optimized kernels) and \emph{trainers} (FSDP parameter updates), with $\pi_{\text{gen}}^\theta$ and $\pi_{\text{train}}^\theta$ denoting the policy on each backend.

Our basic training pipeline adopt GRPO~\citep{shao2024deepseekmath} without KL regularization, augmented with symmetric clipping~\citep{yu2025dapo} and token-level loss-aggregation because it provides a finer, more detailed optimization signal for reasoning, preventing issues like length bias and enabling better learning from complex signals~\cite{sheng2025hybridflow}.
For prompt $x$, the old policy $\pi_{\text{gen}}^{\theta_{\text{old}}}$ generates $G$ completions $\{y_i\}_{i=1}^G$, each rewarded $r_i = r(x, y_i)$. 
Group-normalized advantages follow:
$$
\hat{A}_i = r_i - \frac{1}{G}\sum_{j=1}^G r_j, \qquad \hat{A}_i^G = \frac{\hat{A}_i}{\mathrm{std}(\{r_j\}_{j=1}^G) }
$$

Token-level importance sampling ratios use symmetric clipping~($\epsilon^-$ = $\epsilon^+$ = $\epsilon$):
$$
\rho_{i,t}(\theta) = \frac{\pi_{\text{train}}^\theta(y_{i,t} \mid x, y_{i,<t})}{\pi_{\text{gen}}^{\theta_{\text{old}}}(y_{i,t} \mid x, y_{i,<t})}, \quad
\mathrm{clip}_{\mathrm{sym}}(\rho, \epsilon^-, \epsilon^+) = \mathrm{clip}(\rho, 1-\epsilon^-, 1+\epsilon^+)
$$


The token-level surrogate objective becomes:
\begin{equation}
\begin{aligned}
\mathcal{L}_{\theta}^{\text{GRPO}, \mathcal{D}}
&= \mathbb{E}_{x \sim \mathcal{D},\, \{y_i\}_{i=1}^G \sim \pi^{\text{gen}}_{\theta_{\text{old}}}(\cdot \mid x)}
\Bigg[
\frac{1}{\sum_{i=1}^G |y_i|}
\sum_{i=1}^G \sum_{t=1}^{|y_i|} \min\!\left(
\rho_{i,t}(\theta)\,\hat{A}_i^{G},\;
\mathrm{clip}(\rho_{i,t}(\theta), \epsilon)\,\hat{A}_i^{G}
\right)
\Bigg]
\end{aligned}
\end{equation}

\subsection{Modality based post-training}

\begin{figure}[h!]
    \centering
    \includegraphics[width=0.8\linewidth]{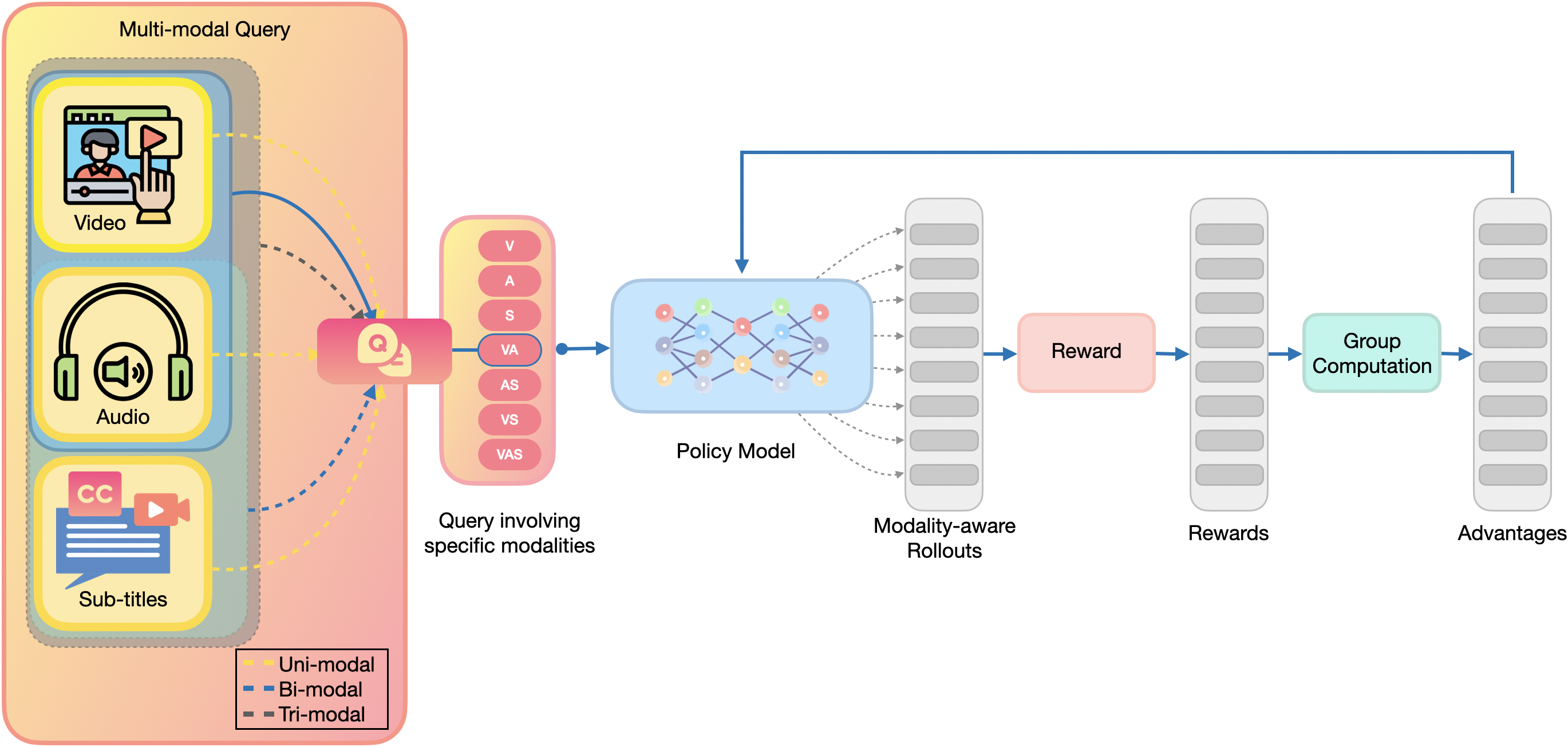}
    \caption{\textbf{Modality-Aware Policy Optimization using modality-stratified GRPO.} Algorithm flow showing RMT-driven stratification. (1) Sample $x$ and extract required modalities $M_x$; (2) Generate $G$ rollouts $\{y_i\}_{i=1}^G$ and compute rewards $r_i = r(x,y_i,M_x)$ and finally group advantages $\hat{A}_i^G$; (3) Form per-RMT batches $\mathcal{B}_M$; (4) Modality-unaware baseline $\hat{g}_{\mathrm{MU}}$ mixes all $R_M$ (high between-group variance); (5) MAPO base $\hat{g}_{\mathrm{MA}}$ normalizes within each $\mathcal{B}_M$ using $\hat{A}_j^{(M)}$ (variance $\mathrm{Var}(\hat{g}_{\mathrm{MA}}) \leq \mathrm{Var}(\hat{g}_{\mathrm{MU}})$). Stratification removes cross-modal reward scale differences, enabling stable multimodal post-training.}
    \label{fig:mapo_training}
\end{figure}

Consider a finite modality set $\mathcal{M} = \{V, A, S\}$ for video, audio, and text/subtitles. 
Each query $x_i \in \mathcal{D}$ is tagged with required-modality subset $M_i \subseteq \mathcal{M}$---the minimal signals needed to solve it reliably (e.g., $M_i = \{V,A\}$ for audio-visual reasoning, $M_i = \{S\}$ for subtitles-only). 
The full MAPO algorithm is mentioned in Alg.~\ref{alg:mapo_algo} and pictorially described for modality based required tags in Fig.~\ref{fig:mapo_training}.
The adaptive KL and dynamic curriculum based changes to MAPO are further described in Alg.~\ref{alg:mapo_w_adap_algo}. 

\begin{algorithm}[!]
\caption{Modality-Aware Policy Optimization (MAPO) $\approx$ Modality-Stratified GRPO}
\label{alg:mapo_algo}
\begin{algorithmic}[1]
\Require Dataset $\mathcal{D}$ with RMTs $M_x \subseteq \{\mathrm{V,A,S}\}$, group size $G$, clip $\epsilon$, learning rate $\alpha$
\Ensure Policy $\pi_\theta$

\While{not converged}
  \State Sample $x \sim \mathcal{D}$; extract modalities $M_x$
  \State Generate $\{y_i\}_{i=1}^G \sim \pi_{\theta_\text{old}}^{\text{gen}}(\cdot|x,M_x)$; score $r_i = r(x,y_i,M_x)$
  \State Compute group advantages: $\hat{A}_i^G = \frac{r_i - \mu_G}{\sigma_G }$
  
  \State \textbf{// Stratify by RMT}
  \State Form $\mathcal{B}_M = \{(x_j,y_j,r_j) : M_{x_j} = M\}$ for each $M$
  
  \State \textbf{Modality-Unaware (baseline):}
  \State $\hat{g}_{\mathrm{MU}} \gets \frac{1}{B} \sum_{j=1}^B \nabla_\theta \log \pi_\theta(y_j|x_j) \hat{A}_j$
  
  \State \textbf{Modality-Aware (MAPO base):}
  \State $\hat{g}_{\mathrm{MA}} \gets \sum_M \frac{1}{|\mathcal{B}_M|} \sum_{(x_j,y_j,r_j)\in\mathcal{B}_M} \nabla_\theta \log \pi_\theta(y_j|x_j,M_x) \hat{A}_j^{(M)}$
  
  \State $\theta \gets \theta - \alpha *  \hat{g}_{\mathrm{MA}}$ \Comment{Variance $\mathrm{Var}(\hat{g}_{\mathrm{MA}}) \leq \mathrm{Var}(\hat{g}_{\mathrm{MU}})$}
\EndWhile
\end{algorithmic}
\end{algorithm}

\begin{algorithm}[!]
\caption{MAPO with Adaptive KL Weighting and/or Curriculum}
\label{alg:mapo_w_adap_algo}
\begin{algorithmic}[1]
\Require RMTs $\mathcal{T}$, Beta$(100,1)$, $L_W=5$
\Ensure $\pi_\theta$, KL histories $\{\mathcal{H}_M\}_{M\in\mathcal{T}}$

\ForAll{$M\in\mathcal{T}$} \State $\mathcal{H}_M \gets []$ \EndFor
\State \textbf{// Epoch 1:} Sort $\mathcal{T}$ by zero-shot accuracy (ascending)
\For{$e=1$ to $E$}
  \If{using KL-based-curriculum}
    \State $s_M = \frac{1}{|W_M|} \sum_{k\in W_M} D_{\mathrm{KL},M}^{(k)}$ \Comment{$|W_M|=\min(L_W,|\mathcal{H}_M|)$}
    \State Reorder $\mathcal{T}$ by $s_M$ (descending)
  \EndIf
  \For{$M \in \mathcal{T}$}
    \State Sample $\mathcal{B}_M = \{(x_j,y_j,r_j):M_{x_j}=M\}$
    \State Compute $\hat{A}_j^{(M)}$ within $\mathcal{B}_M$
    \State $p_{\mathrm{emp}} \gets$ histogram($\{r_j\}_{j\in\mathcal{B}_M}$)
    \State $D_{\mathrm{KL},M} \gets D_{\mathrm{KL}}(p_{\mathrm{emp}} \| \mathrm{Beta}(100,1))$
    \State $\mathcal{H}_M$.append($D_{\mathrm{KL},M}$)
    \State $z_M = \frac{D_{\mathrm{KL},M} - \mu_{\mathcal{H}_M}}{\sigma_{\mathcal{H}_M} + \varepsilon}$
    \State $w_M = \mathrm{sigmoid}(z_M)$
    \State $\mathcal{L}^{\mathrm{MAPO}_{w}}_\theta = \frac{w_{M}}{|\mathcal{B}_M|} \mathcal{L}^{\mathrm{GRPO},\mathcal{B}_M}_\theta$
    \State $\theta \gets$ Optimizer step on $\mathcal{L}_\theta^{\mathrm{MAPO}_{w}}$
  \EndFor
\EndFor
\end{algorithmic}
\end{algorithm}

\section{Implementation details and observations}
\label{app_subsec:impl_details}

\begin{table}[!]
\centering
\caption{Key Hyperparameters for veRL Post-Training}
\label{tab:verl-hyperparams-caption}
\small
\resizebox{1.0\textwidth}{!}{
\begin{tabular}{@{}llll@{}}
\toprule
\textbf{High-Level Setting} & \textbf{Parameter} & \textbf{MAPLE-QA} & \textbf{MAPLE-Caption} \\
\midrule
\multirow{5}{*}{\textbf{Data}} 
& Dataset & tagged daily\_omni (train: 47893, test: 5001) & tagged vast\_omni (train: 5120, test: 5348) \\
& Train batch size & 256 & 64 \\
& Max prompt/response length & 8192 / 2048 & 8192 / 2048 \\
& Modalities & Audio, Video, Subtitles & Audio, Video, Subtitles \\
& Prompt key & tag\_based\_prompt & tag\_based\_prompt \\
\midrule
\multirow{5}{*}{\textbf{Models}} 
& Actor & Qwen2.5-Omni-3B (bfloat16, LoRA rank=0) & Qwen2.5-Omni-3B (bfloat16, LoRA rank=0) \\
& Reward Model & Ground Truth & GPT-4o \\
& Tokenizer & Qwen2.5-Omni-3B & Qwen2.5-Omni-3B \\
& FSDP & param+optimizer offload, gradient checkpointing & param+optimizer offload, gradient checkpointing \\
& Mixed precision & fp16 params/buffers & fp16 params/buffers \\
\midrule
\multirow{5}{*}{\textbf{RL Algorithm}} 
& Estimator & GRPO & GRPO \\
& $\gamma$ / $\lambda$ & 1.0 / 1.0 & 1.0 / 1.0 \\
& Actor clip ratio & 0.2 (low/high) & 0.2 (low/high) \\
& Advantage norm & std\_in\_grpo & std\_in\_grpo \\
& Loss aggregation & token-mean & token-mean \\
\midrule
\multirow{4}{*}{\textbf{Optimization}} 
& Actor LR & $2\times10^{-6}$ & $2\times10^{-6}$ \\
& Gradient clip & 1.0 & 1.0 \\
& Total epochs & 2~(optimal)/5~(adaptive) & 5~(adaptive) \\
& Warmup & constant (steps=-1) & constant (steps=-1) \\
\midrule
\multirow{9}{*}{\textbf{Rollout}} 
& Engine & vLLM (sync, bfloat16) & vLLM (sync, bfloat16) \\
& Rollout size & 8 & 8 \\
& Temperature & 0.6 (train and test) & 0.7 (train and test) \\
& Top-p & 0.95 (train and test) & 0.9 (train and test) \\
& GPU mem util & 0.6 & 0.6 \\
& Max batched tokens & 12288 & 12288 \\
& Chunked prefill & True & True \\
& Response length & 2048 & 2048 \\
& Ignore EOS & False & False \\
\midrule
\multirow{2}{*}{\textbf{Hardware}} 
& GPUs & 4× NVIDIA H100 80GB (Hopper) & 4× NVIDIA H100 80GB (Hopper) \\
& CUDA & 12.6 & 12.6 \\
\bottomrule
\end{tabular}
}
\end{table}

The veRL configurations used for training/testing are mentioned in Table~\ref{tab:verl-hyperparams-caption}.

\subsection{Optimal Training Strategy - Training characteristic curves}

\begin{figure*}
    \centering
    \begin{subfigure}{0.48\linewidth}
        \centering
        \includegraphics[width=\linewidth]{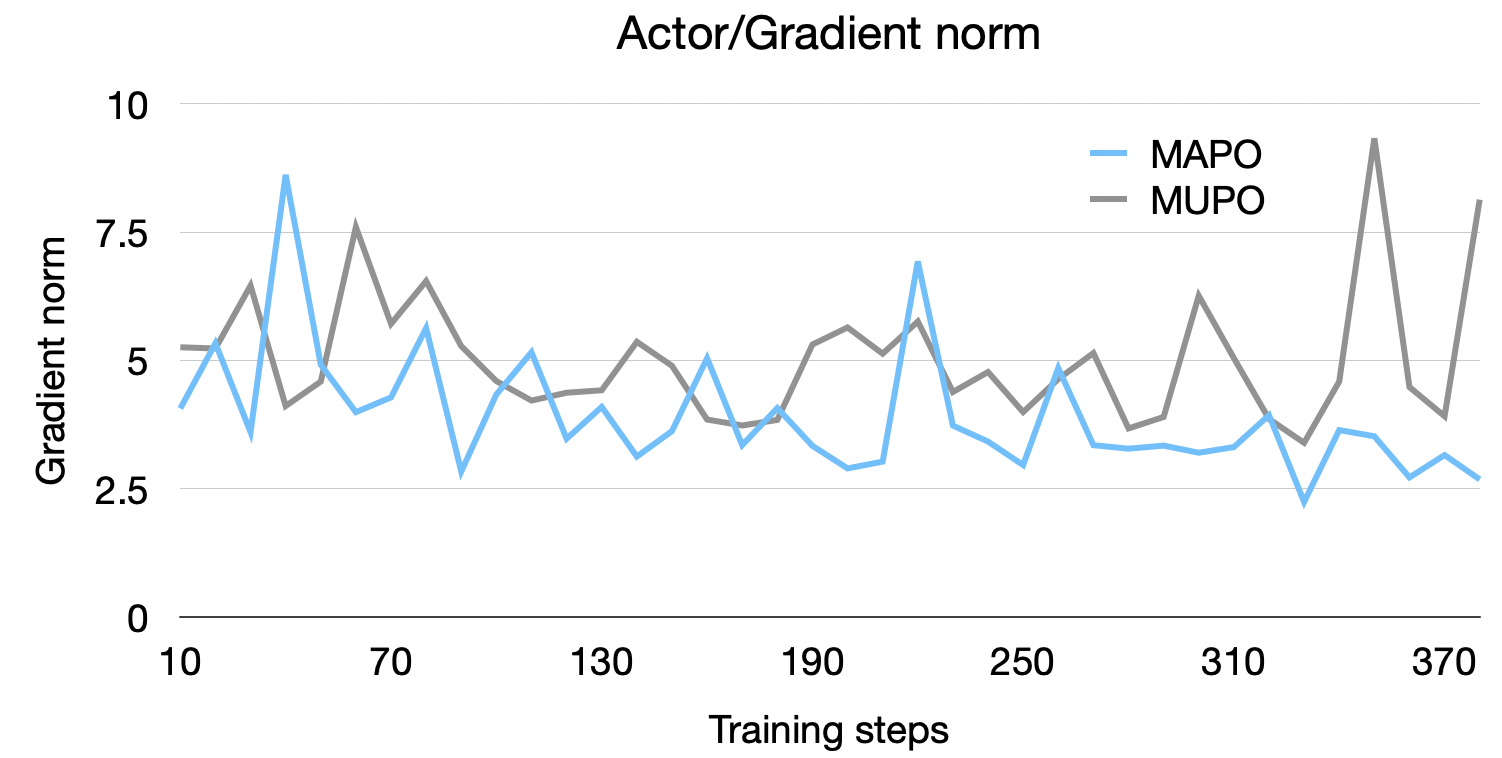}
        \label{fig:mu_vs_mapo_gradnorm}
    \end{subfigure}
    \caption{MAPO achieving smaller, lower-variance policy gradient norms vs. MUPO with similar training configurations.}
    \label{fig:gradient_norm_comparison}
\end{figure*}

\begin{figure*}
    \centering
    \begin{subfigure}{0.48\linewidth}
        \centering
        \includegraphics[width=\linewidth]{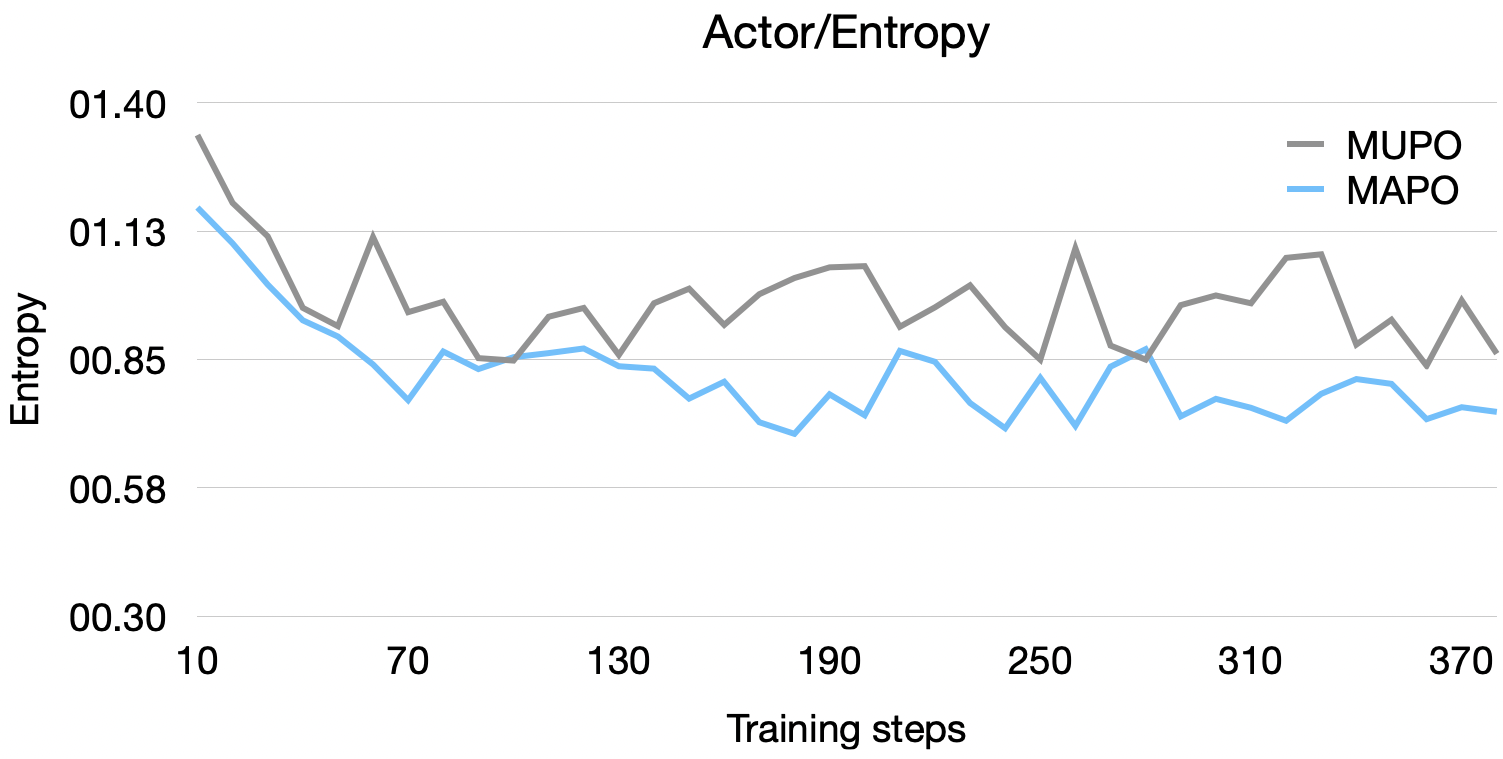}
        \caption{Entropy Reduction}
        \label{fig:mu_vs_mapo_entropy}
    \end{subfigure}
    \hfill
    \begin{subfigure}{0.48\linewidth}
        \centering
        \includegraphics[width=\linewidth]{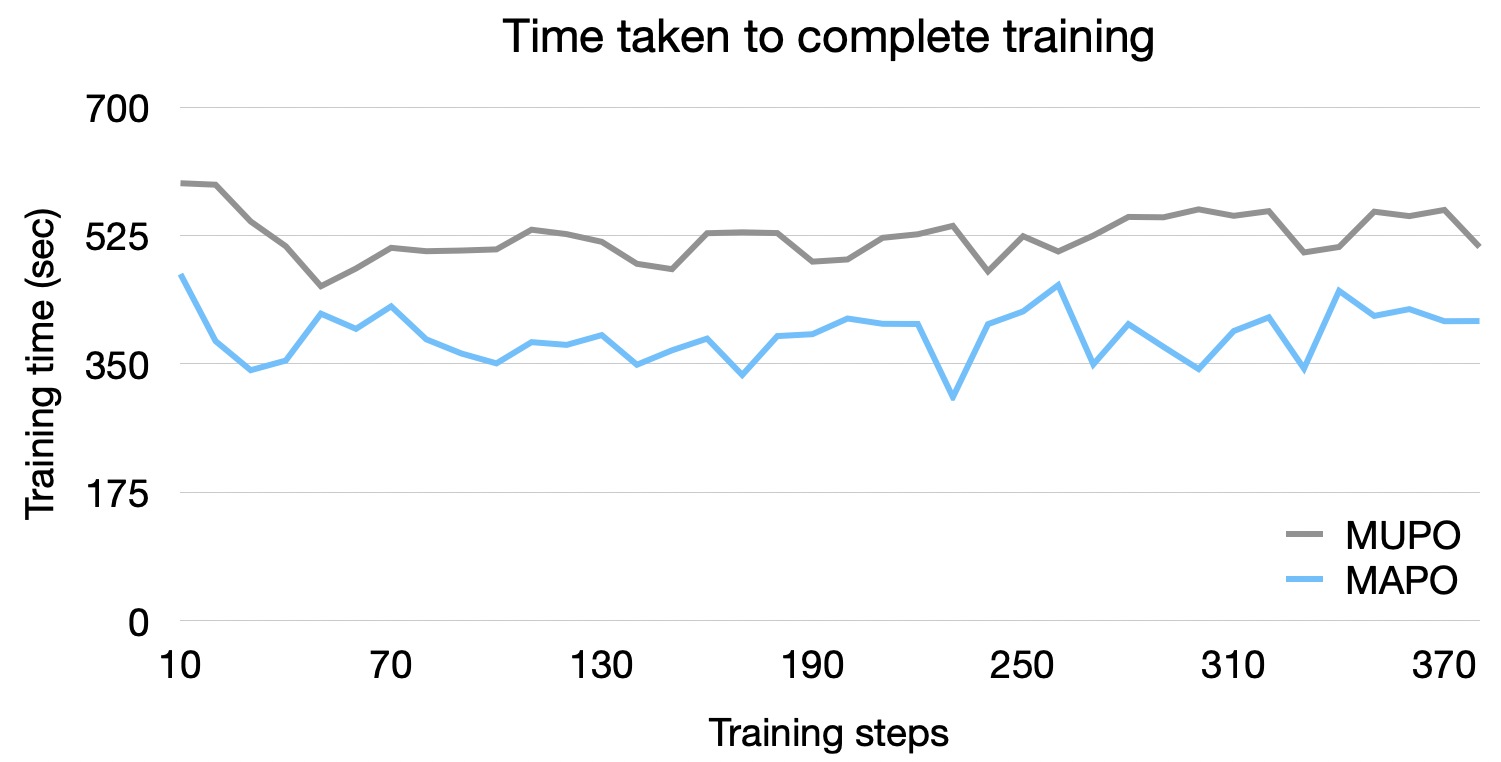}
        \caption{Training time per step}
        \label{fig:mu_vs_mapo_time}
    \end{subfigure}
    
    \caption{\textbf{Entropy and step-time comparison for MUPO vs. MAPO under identical training settings.} 
Left: MAPO exhibits lower and more stable policy entropy than the modality-unaware MUPO baseline, indicating better-regularized optimization. 
Right: MAPO achieves reduced wall-clock time per training step by processing only the required modality signals, improving efficiency without changing the overall configuration.}

    \label{fig:maple_ablations}
\end{figure*}

\begin{figure*}
    \centering
    \begin{subfigure}{0.48\linewidth}
        \centering
        \includegraphics[width=\linewidth]{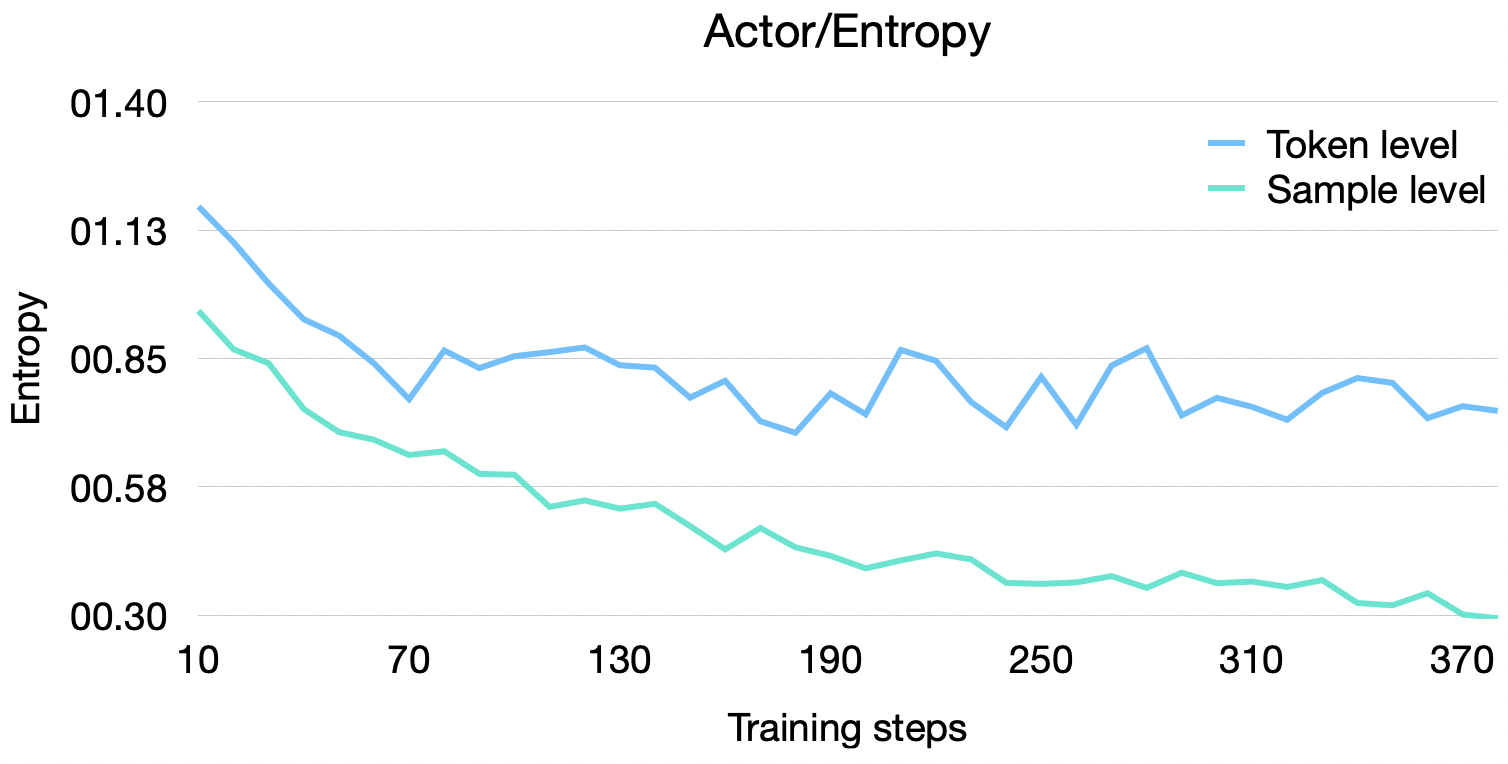}
        \caption{Entropy Reduction}
        \label{fig:loss_agg_entropy_compa}
    \end{subfigure}
    \hfill
    \begin{subfigure}{0.48\linewidth}
        \centering
        \includegraphics[width=\linewidth]{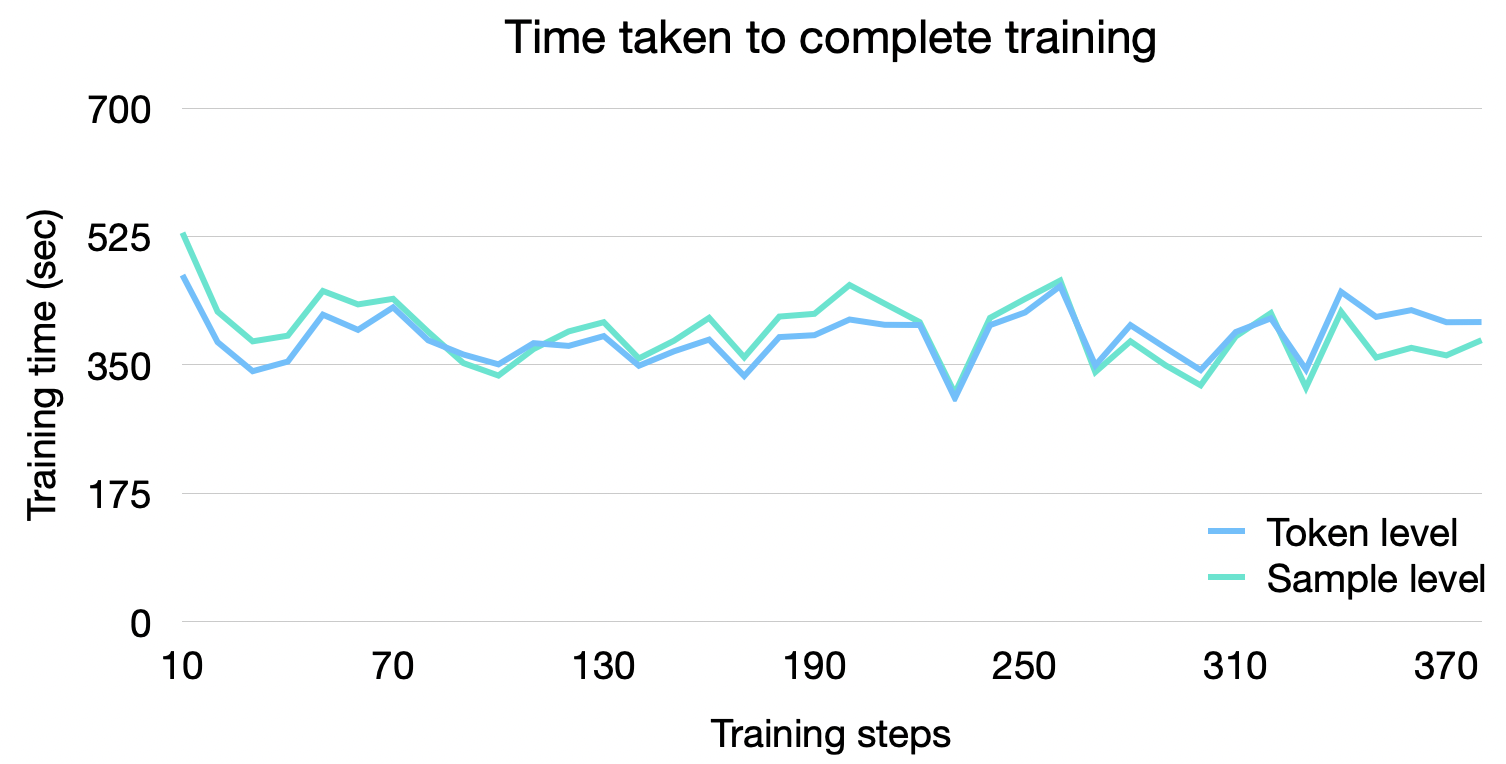}
        \caption{Training time per step}
        \label{fig:loss_agg_time_comp}
    \end{subfigure}
    
    \caption{\textbf{Entropy and step-time comparison across loss aggregation strategies for MAPO.} 
Left: Sample-level aggregation achieves substantially lower and more stable policy entropy than token-level aggregation. 
Right: Sample-level incurs slightly higher wall-clock time per step but yields superior optimization stability.}
    \label{fig:loss_agg_comparisons}
\end{figure*}

\begin{figure*}
    \centering
    \begin{subfigure}{0.48\linewidth}
        \centering
        \includegraphics[width=\linewidth]{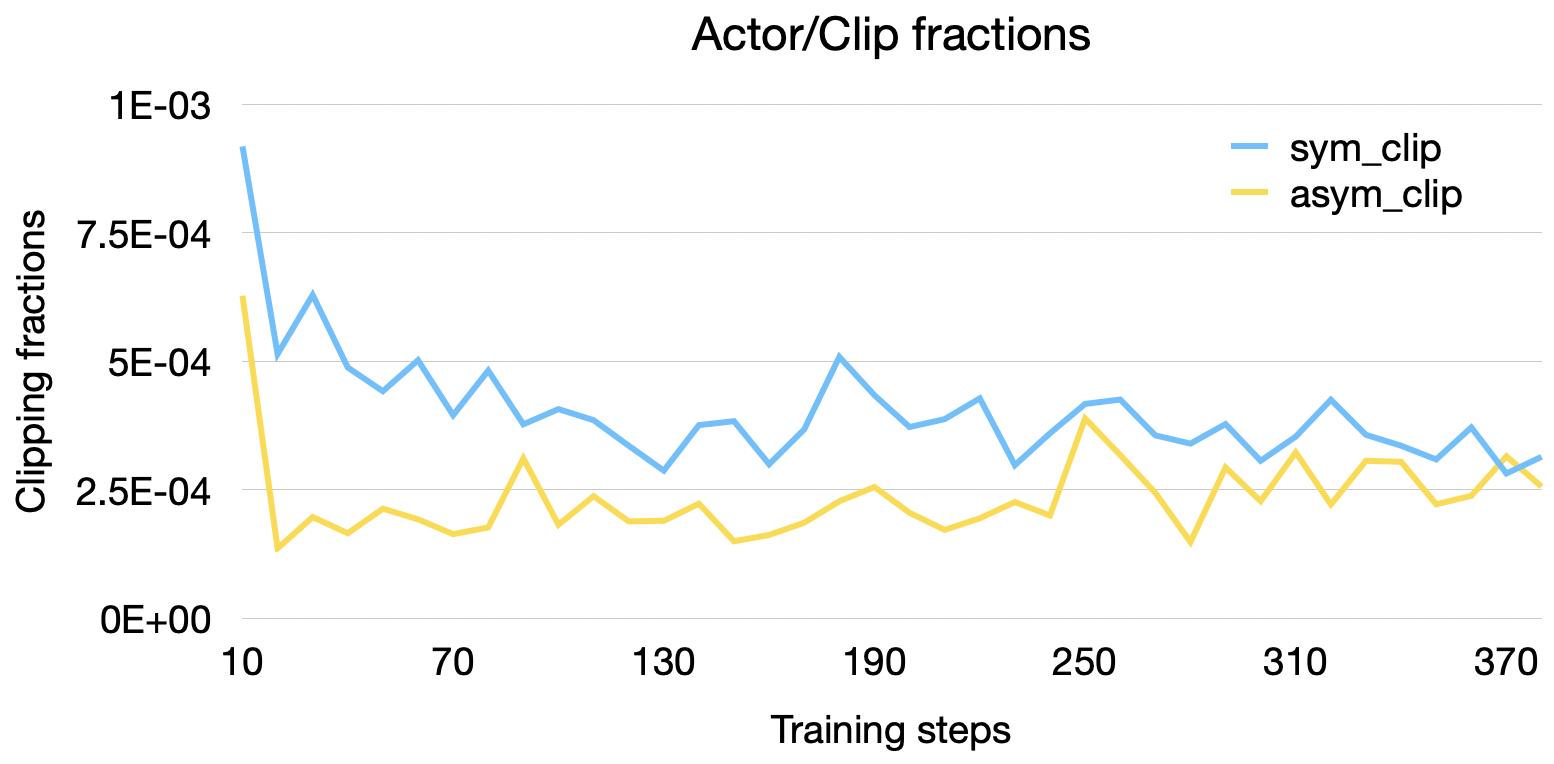}
        \caption{Clipping fractions ratio}
        \label{fig:clip_fracs}
    \end{subfigure}
    \hfill
    \begin{subfigure}{0.48\linewidth}
        \centering
        \includegraphics[width=\linewidth]{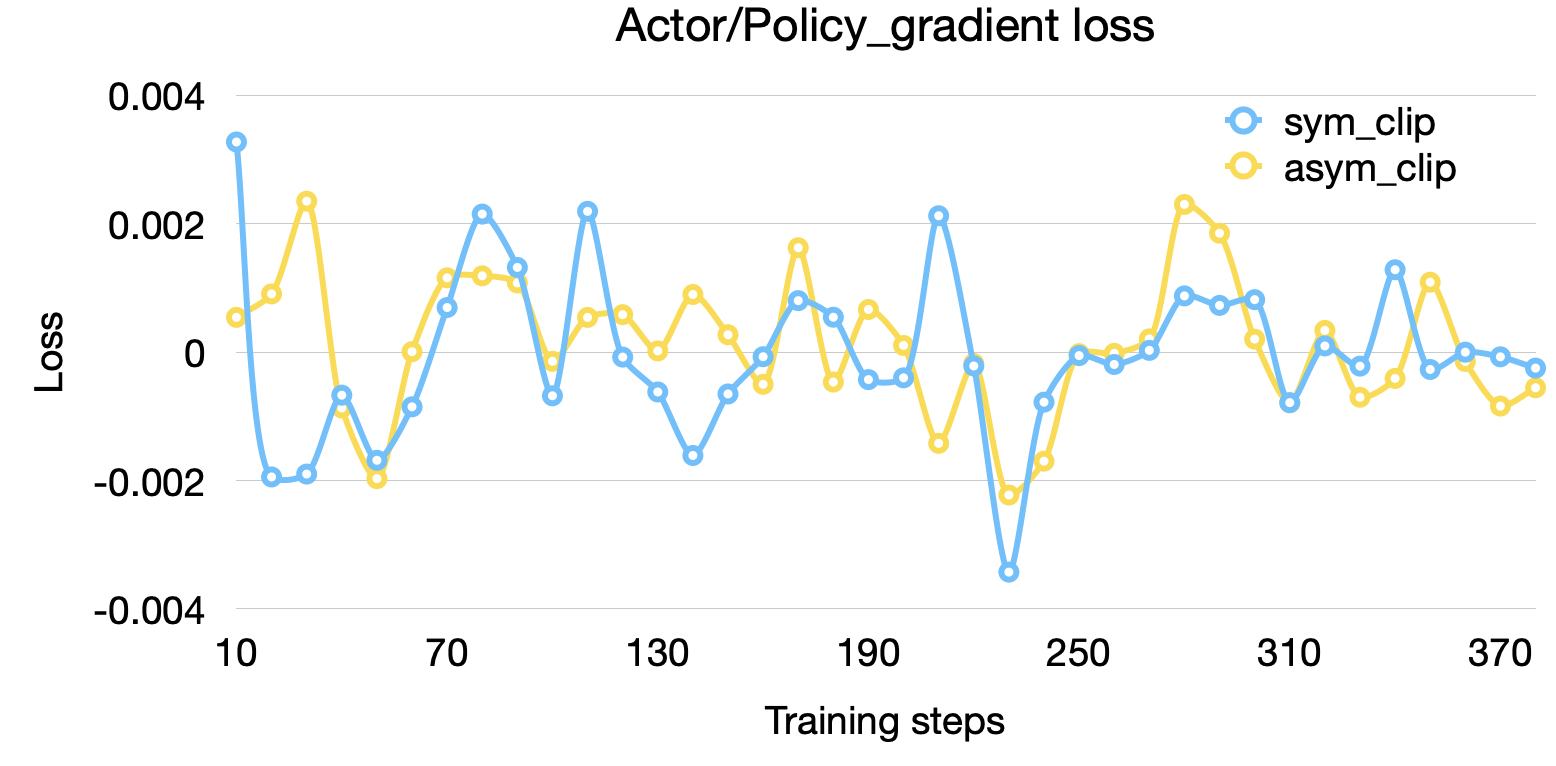}
        \caption{Policy gradient loss}
        \label{fig:clip_pg_loss}
    \end{subfigure}
    
    \caption{\textbf{Clipping behavior and policy loss stability: symmetric vs. asymmetric clipping in MAPO.} 
Left: Asymmetric clipping ($\epsilon^+=0.3$, $\epsilon^-=0.2$) reduces clip fractions by 71.65\% vs. symmetric clipping ($\epsilon=0.2$). 
Right: Lower clipping enables more exploration, yielding more stable final policy gradient loss.}
    \label{fig:clipping_comparison}
\end{figure*}

\begin{figure*}
    \centering
    \begin{subfigure}{0.48\linewidth}
        \centering
        \includegraphics[width=\linewidth]{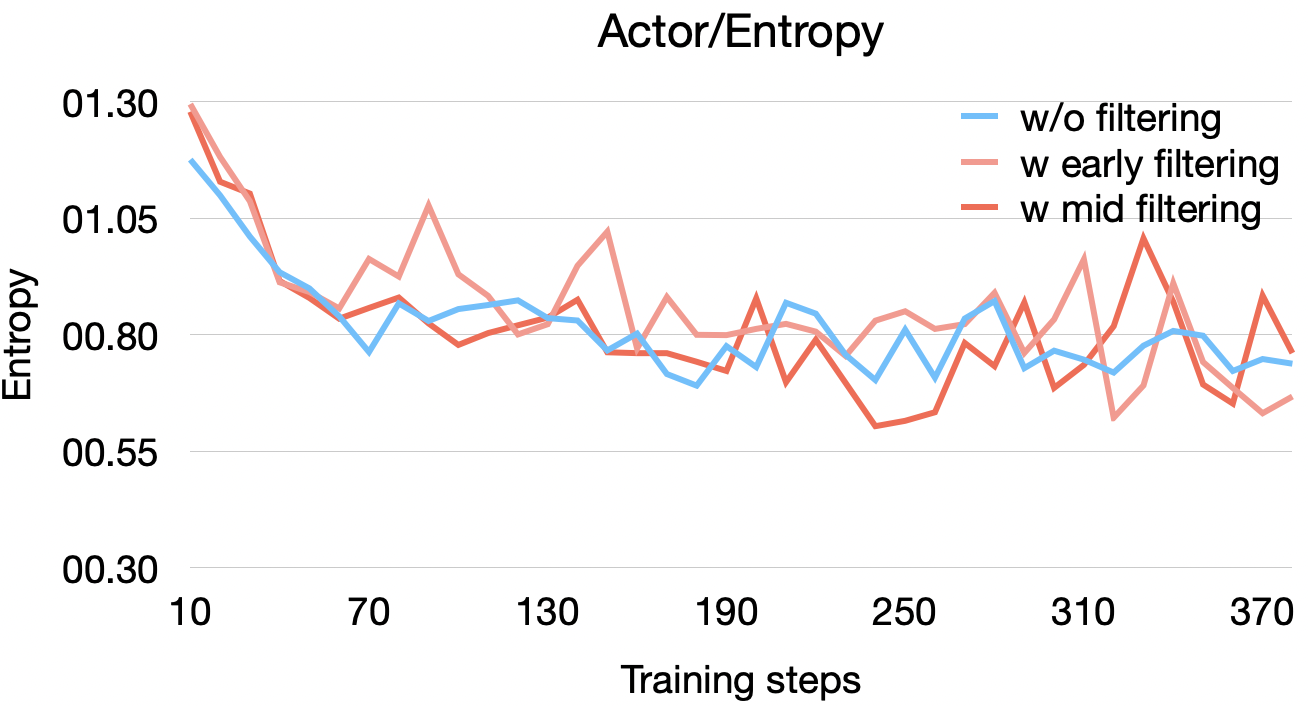}
        \caption{Entropy Reduction}
        \label{fig:entropy_sampling}
    \end{subfigure}
    \hfill
    \begin{subfigure}{0.48\linewidth}
        \centering
        \includegraphics[width=\linewidth]{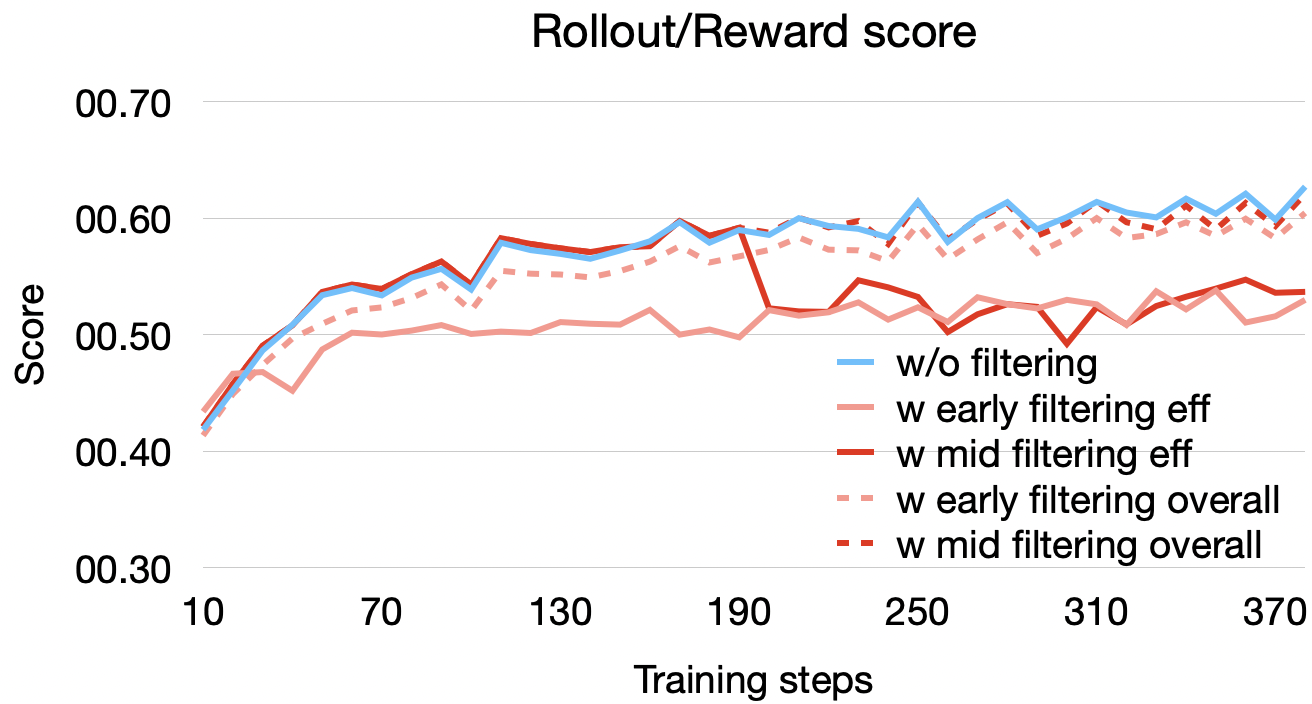}
        \caption{Rollout score~(Pass@1)}
        \label{fig:rollout_score_sampling}
    \end{subfigure}
    \caption{\textbf{Entropy and rollout scores across sampling strategies in MAPO.} 
Left: All strategies (no sampling, early filtering, mid-training filtering) achieve equivalent entropy reduction and rollout scores (solid: effective scores; dotted: including filtered samples). 
Right: Early filtering substantially reduces training time per step while maintaining optimization quality.}

    \label{fig:sampling_comparison}
\end{figure*}

\begin{figure*}
    \centering
    \begin{subfigure}{0.33\linewidth}
        \centering
        \includegraphics[width=\linewidth]{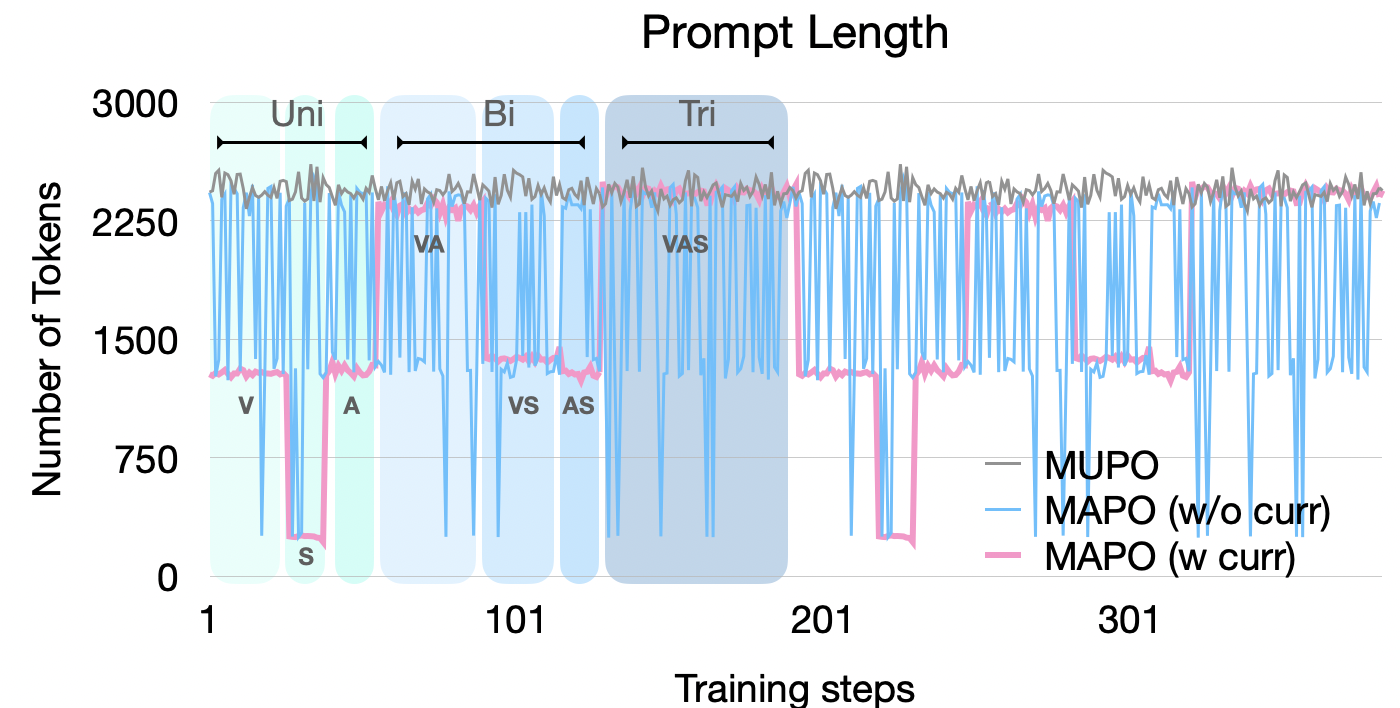}
        \caption{Prompt Length}
        \label{fig:curr_prompt_length}
    \end{subfigure}
    \hfill
    \begin{subfigure}{0.33\linewidth}
        \centering
        \includegraphics[width=\linewidth]{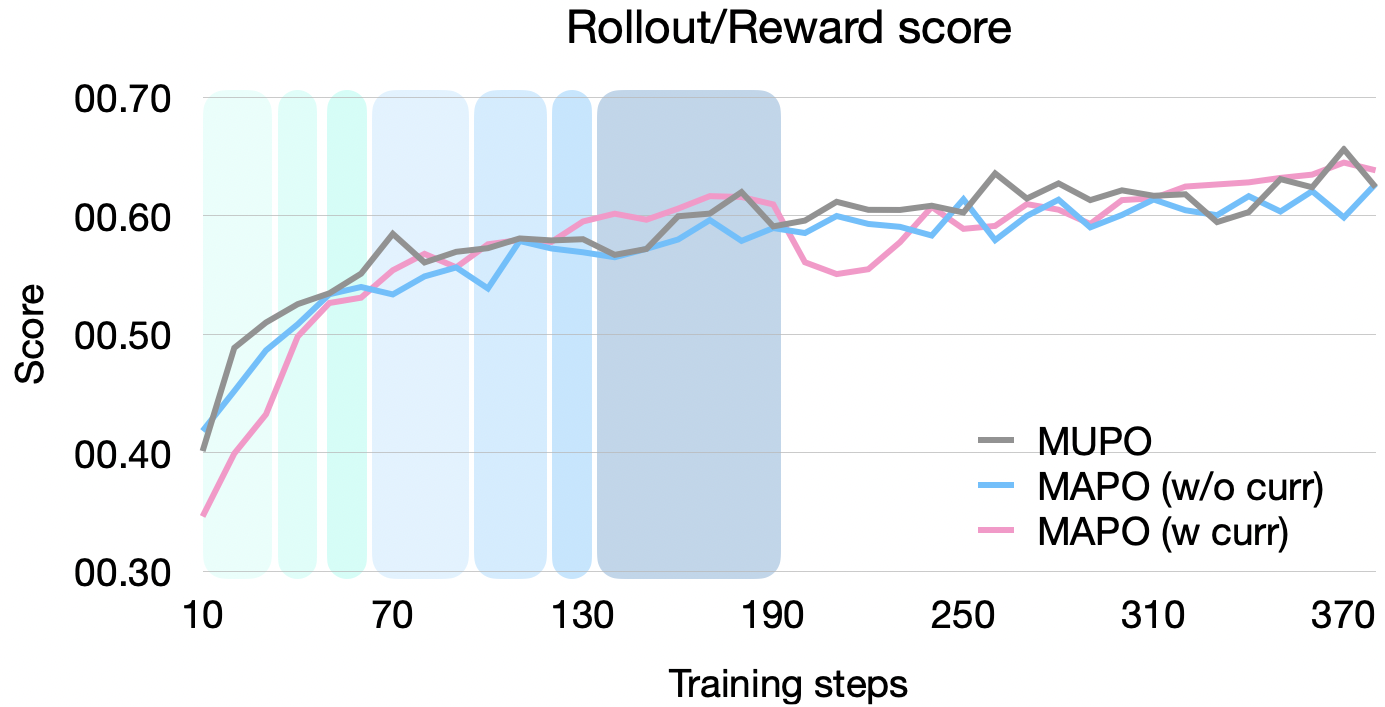}
        \caption{Rollout score~(Pass@1)}
        \label{fig:curr_rollout_score}
    \end{subfigure}
     \hfill
    \begin{subfigure}{0.33\linewidth}
        \centering
        \includegraphics[width=\linewidth]{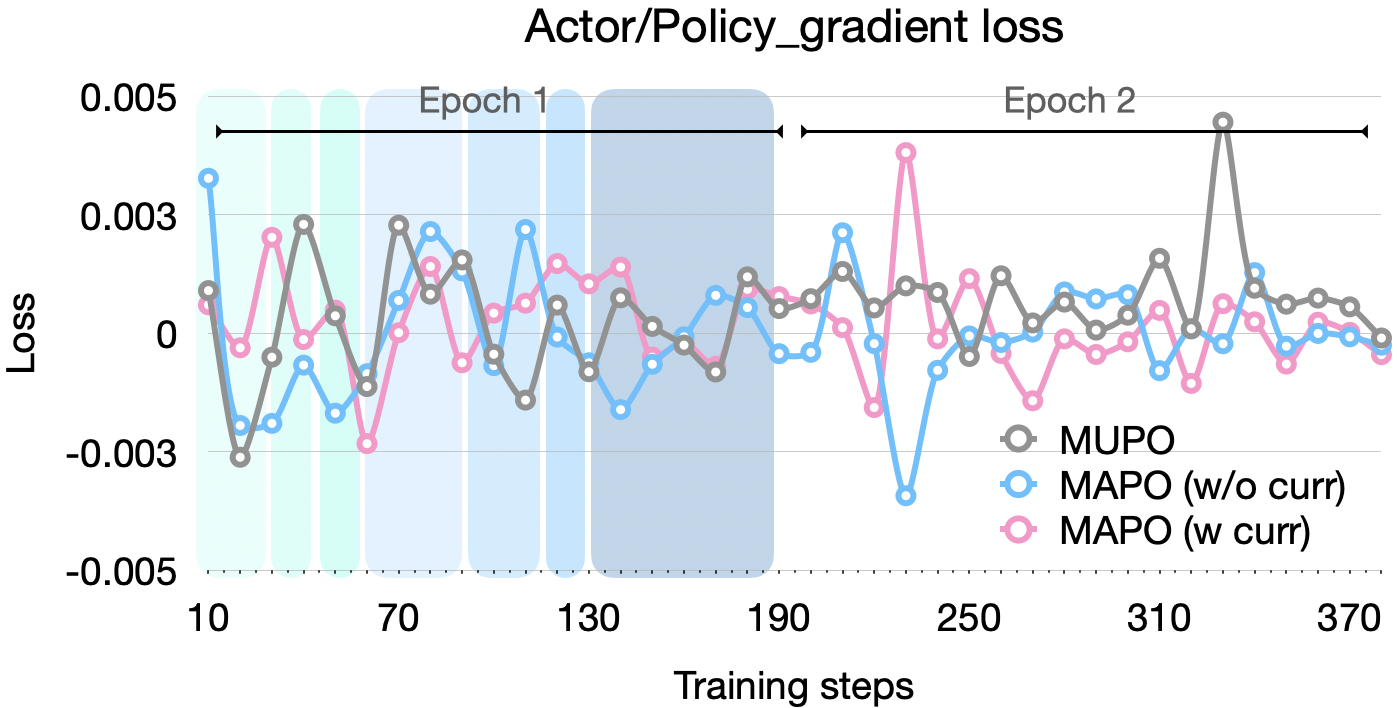}
        \caption{Policy gradient loss}
        \label{fig:curr_pg_loss}
    \end{subfigure}
    \caption{\textbf{Training dynamics across curriculum strategies in MAPO.} 
Left: Prompt length progression (uni$\to$bi$\to$tri-modal) for curriculum vs. mixed training. 
Middle/Right: Curriculum yields higher rollout scores and more stable policy gradient loss throughout training.}
    \label{fig:curriculum_comparison}
\end{figure*}

\begin{figure*}
    \centering
    \begin{subfigure}{0.48\linewidth}
        \centering
        \includegraphics[width=\linewidth]{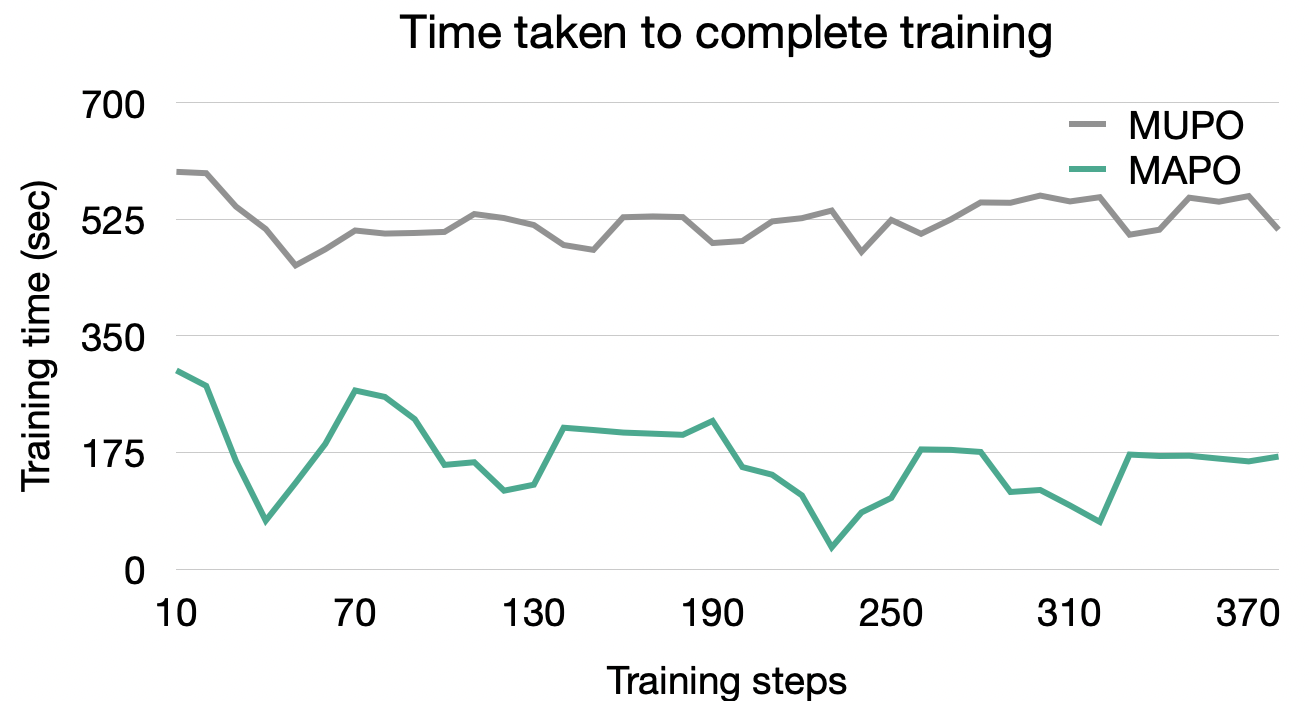}
        \caption{Training time per step}
        \label{fig:curr_prompt_length}
    \end{subfigure}
    \hfill
    \begin{subfigure}{0.48\linewidth}
        \centering
        \includegraphics[width=\linewidth]{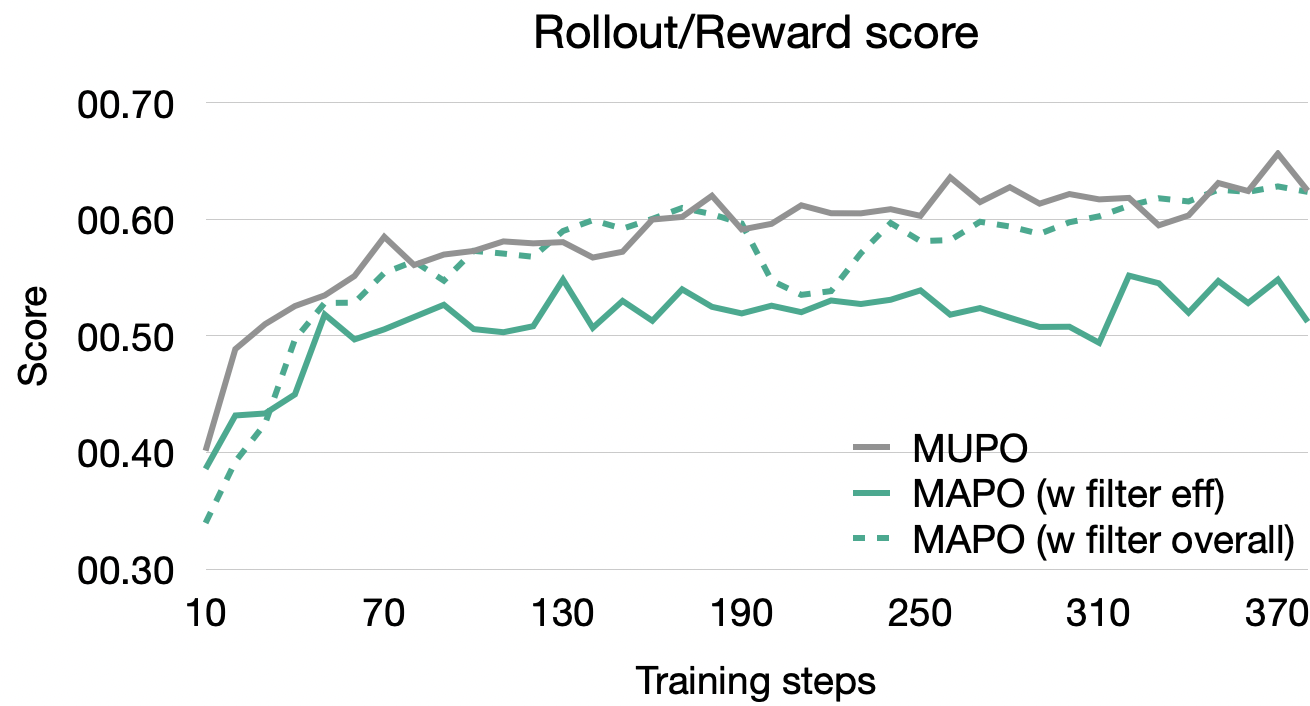}
        \caption{Rollout score~(Pass@1)}
        \label{fig:curr_rollout_score}
    \end{subfigure}
    
    \begin{subfigure}{0.48\linewidth}
        \centering
        \includegraphics[width=\linewidth]{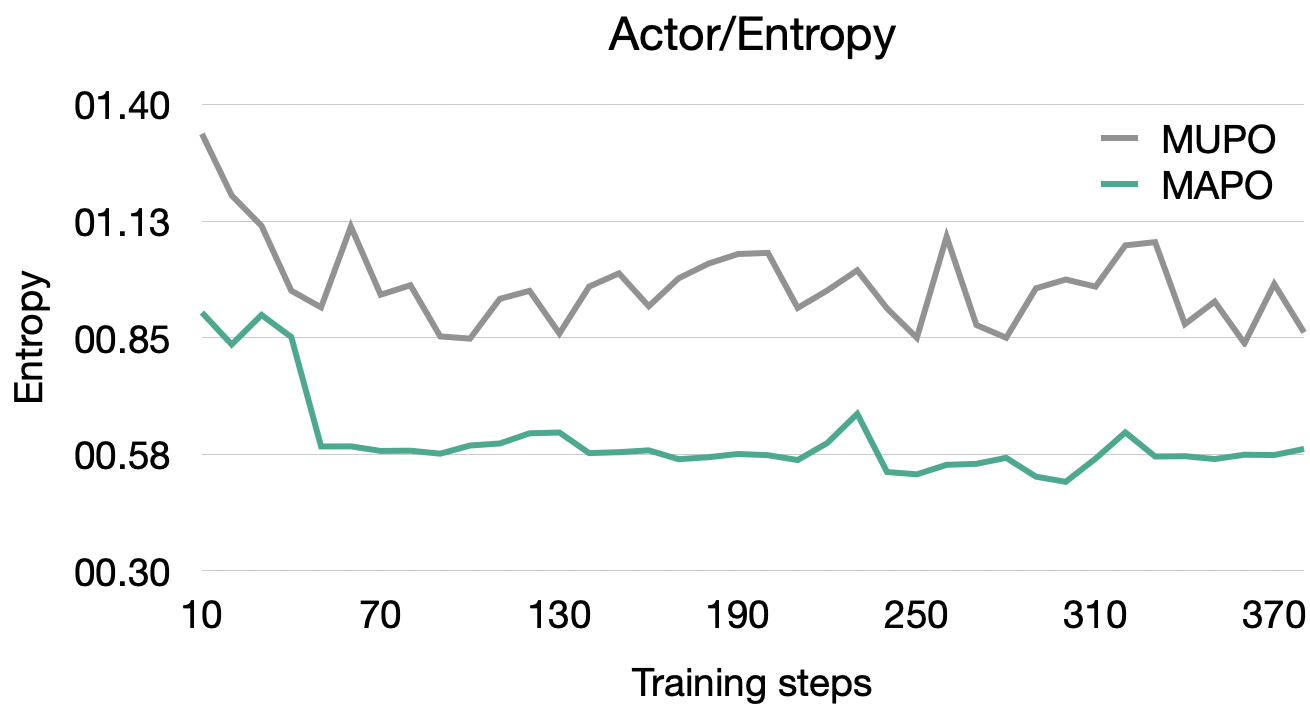}
        \caption{Policy gradient loss}
        \label{fig:curr_pg_loss}
    \end{subfigure}
    \caption{\textbf{MUPO vs. final Full-recipe: comprehensive training dynamics.} 
a): Full-recipe reduces training time per step. 
b): Equivalent rollout scores throughout training. 
c): Lower, more stable entropy. Full-recipe improves output performance, robustness, and cost-efficiency across all metrics.}
    \label{fig:maple_recipe_comparison}
\end{figure*}

This appendix presents training curves for algorithmic variants from Sec.~\ref{sec:opt_trn_strat} on MAPLE-QA (2 epochs, ~380 steps). 
Training stabilizes beyond 2 epochs, justifying this duration.
Fig.~\ref{fig:gradient_norm_comparison} reveals MAPO achieving smaller, lower-variance policy gradient norms vs. MUPO with similar training configurations.
Figure~\ref{fig:mu_vs_mapo_time} shows MAPO achieving lower, more stable entropy and reduced step-time vs. MUPO by processing only required modality signals.
Fig.~\ref{fig:loss_agg_comparisons} shows sample-level aggregation achieving superior entropy reduction despite modestly higher step-time vs. token-level aggregation.
Fig.~\ref{fig:clipping_comparison} shows asymmetric clipping reducing clip fractions 71.65\% vs. symmetric, enabling more stable policy gradient loss through increased exploration.
Fig.~\ref{fig:sampling_comparison} shows early/mid-training filtering maintaining equivalent entropy and rollout scores vs. no filtering, but substantially reducing training time per step.
Fig.~\ref{fig:curriculum_comparison} shows curriculum learning (uni$\to$bi$\to$tri-modal) achieving higher rollout scores and more stable policy gradient loss vs. mixed training.
Fig.~\ref{fig:maple_recipe_comparison} shows Full-recipe outperforming MUPO across training time, eval scores, and entropy---delivering superior performance, robustness, and efficiency.

\begin{figure*}
    \centering
    \begin{subfigure}{0.48\linewidth}
        \centering
        \includegraphics[width=\linewidth]{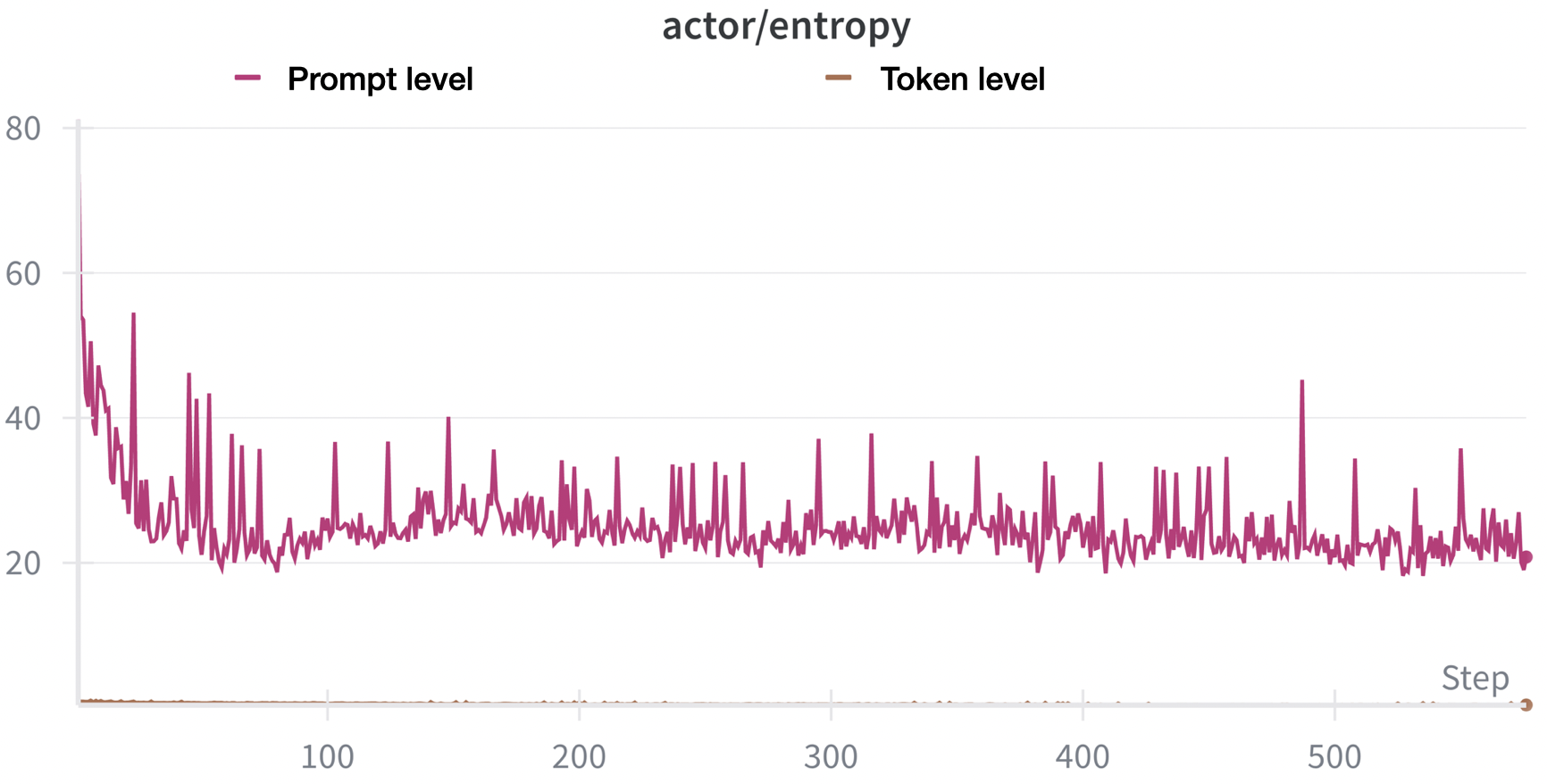}
        \label{fig:mu_vs_mapo_entropy}
    \end{subfigure}
    \caption{\textbf{Entropy curve for loss aggregation at prompt level.} We noticed no convergence of entropy by this mode.}
    \label{fig:loss_agg_prompt_level}
\end{figure*}

Additional observations: Fig.~\ref{fig:loss_agg_prompt_level} shows prompt-level aggregation failing to converge entropy in MAPO even with more epochs of training. 

\newpage

\subsection{Adaptive Training Strategy}
\label{app:adaptive_training_strategy_implementation}


For adaptive training strategies, we train all variants for 5 epochs on both datasets to ensure full convergence and accurately measure performance gains from each component.
In modality-aware post-training, data are partitioned into modality tags
$M \in \mathcal{M}$ (e.g., $\mathrm{V},\mathrm{A},\mathrm{S},\mathrm{VA},\mathrm{VS},\mathrm{AS},\mathrm{VAS}$),
and GRPO-style updates scale with observed advantages.
This creates tag-wise imbalance where harder tags yield smaller gradient signals and can be overshadowed by easier tags.
Accordingly, our implementation consists of:
(i) a KL-based difficulty metric,
(ii) difficulty-adaptive reweighting (controlling \emph{how much} to update per tag), and
(iii) KL-driven curriculum scheduling (controlling \emph{when} to update each tag).

\subsubsection{KL-Based Difficulty Metric}
\label{app:kl_difficulty}

\noindent\textbf{Target distribution.}\quad
We quantify tag difficulty using the KL divergence between an empirical reward distribution $p_{\mathrm{emp}}$
(measured on a tag-batch) and a fixed near-optimal target $p_{\mathrm{tgt}}$.
For captioning, we use $p_{\mathrm{tgt}} \sim \mathrm{Beta}(100,1)$, whose mode is $0.99$.



\noindent\textbf{Captioning (continuous rewards).}\quad
Let $R \in [0,1]$ denote per-sample captioning rewards.
We approximate $p_{\mathrm{emp}}$ on $[0,1]$ via histogram discretization
using $K$ uniform bins (width $\Delta x$), bin centers $\{x_i\}_{i=1}^{K}$,
and a normalized histogram with \texttt{density=True} to obtain an empirical
\emph{density} $p(x_i)$.
With $q(x_i)=\mathrm{BetaPDF}(x_i; 100, 1)$, we compute
\begin{equation}
\label{eq:app_kl_caption}
D_{\mathrm{KL},M}^{\text{Cap}}
\approx
\sum_{i=1}^{K}
p(x_i)\,
\log \frac{p(x_i)+\varepsilon}{q(x_i)+\varepsilon}
\, \Delta x,
\end{equation}
where we use $K=50$ and a small $\varepsilon$ for numerical stability.

\medskip
\noindent\textbf{QA (discrete rewards).}\quad
For QA, rewards are binary $R \in \{0,1\}$.
To match the captioning target, we set the Bernoulli target success rate to the mean of $\mathrm{Beta}(100,1)$:
\begin{equation}
\label{eq:app_p_tgt}
p_{\text{tgt}} = \frac{100}{101} \approx 0.9901.
\end{equation}
For a tag-batch of size $B_M$, the empirical accuracy is
\begin{equation}
\label{eq:app_p_emp}
p_{\text{emp}}
=
\frac{1}{B_M}
\sum_{j=1}^{B_M}
\mathbbm{1}\{R_j = 1\},
\end{equation}
and we compute the Bernoulli--Bernoulli KL divergence
\begin{equation}
\label{eq:app_kl_qa}
D_{\mathrm{KL},M}^{\text{QA}}
=
p_{\text{emp}}
\log \frac{p_{\text{emp}}}{p_{\text{tgt}}}
+
(1 - p_{\text{emp}})
\log \frac{1 - p_{\text{emp}}}{1 - p_{\text{tgt}}},
\end{equation}
with probabilities clipped to $[\varepsilon, 1-\varepsilon]$ if needed.

\noindent\textbf{Interpretation and usage.}\quad
Hard tags typically exhibit $p_{\mathrm{emp}} \ll p_{\mathrm{tgt}}$ (large KL), while easy tags approach the target (KL near zero).
We use KL as a unified difficulty signal for both reweighting and curriculum scheduling.

\subsubsection{Difficulty-Adaptive Reweighting: Practical Details}
\label{app:adaptive_weighting_design}

\noindent\textbf{What is adaptive about the weights.}\quad
The difficulty-adaptive reweighting (defined in the main paper) is intended to correct \emph{relative} imbalance across tags,
rather than to rescale absolute reward magnitudes. We therefore base the weight update on each tag's KL statistics and
normalize them using a short KL history to make the weights robust to reward drift over training.

\noindent\textbf{History window size $L_{\mathcal{H}}$.}\quad
We set the history length $L_{\mathcal{H}}$ to be approximately the number of steps in one epoch (i.e., steps-per-epoch).
This choice ensures that the history aggregates KL values across (almost) all modality tags within a training cycle,
which better approximates the global reward distribution and stabilizes \emph{relative} difficulty estimation.
In contrast, overly short windows yield noisy, batch-specific weights, while overly long windows make the weights slow to react
to changes in learning dynamics.

\noindent\textbf{Normalization and squashing.}\quad
We compute a z-score from the running mean and standard deviation over the KL history, and squash it with a sigmoid to map to $(0,1)$.
In practice, most z-scores fall within a moderate range (e.g., $[-3,3]$), which avoids extreme scaling and reduces oscillatory updates
caused by noisy per-batch rewards.

\noindent\textbf{Why this is complementary to curriculum.}\quad
Reweighting controls \emph{how much} each tag updates, but when difficulty gaps are extreme, hard-tag gradients can still be
overshadowed by easy-tag batches if trained in an unfavorable order.
We therefore additionally control \emph{when} each tag is optimized via curriculum scheduling.

\subsubsection{KL-Driven Curriculum: Practical Details}
\label{app:adaptive_curriculum_impl}

\noindent\textbf{Window size ($L_W$).}\quad
We compute curriculum priorities from a short recent window of KL values and set $L_W=5$.
This value was chosen as a practical heuristic rather than through explicit statistical tuning:
we use the \emph{last five} tag-batches because their performance signals are typically the most \emph{mature} (i.e., after the initial transient),
providing a stable yet responsive estimate of current tag difficulty.

\noindent\textbf{First-epoch initialization.}\quad
In the first epoch, KL histories are unavailable.
We therefore initialize the curriculum order using zero-shot performance, assigning higher priority to tags with lower
zero-shot accuracy. After the first epoch, once sufficient KL history has accumulated, we switch to KL-based scheduling.

\noindent\textbf{Why ordering matters.}\quad
Curriculum scheduling complements reweighting by controlling \emph{when} each tag is optimized.
By front-loading underperforming (high-KL) tags, the model learns from their weaker gradient signals before the optimization
trajectory is dominated by easier tags with stronger gradients.
\subsubsection{Reward Evaluation Note (pointer to D.2)}
\label{app:ref_combine}

\noindent\textbf{Ref-combine protocol.}\quad
We adopt a ref-combine evaluation setup for captioning rewards to minimize ambiguity in assessing \emph{hallucination} and \emph{modality-fusion} quality in multimodal captions.
The judge is provided with the tag-specific ground-truth caption along with
uni-modal reference captions (denoted $\{\mathrm{GT}_M,\mathrm{Ref}_{\mathrm{V}},\mathrm{Ref}_{\mathrm{A}},\mathrm{Ref}_{\mathrm{S}}\}$),
and scores the generated fusion caption along three axes (missing, hallucination, fusion), producing a continuous reward in $[0,1]$.
See Appendix~\ref{sec:llm_judge} for the full judge prompt and rubric.

\paragraph{Modality-unaware Policy Optimization for Captioning.}
While QA tasks naturally accommodate full-modality inputs (V, A, S) because answers are derived from content rather than fixed reference captions, captioning tasks differ fundamentally: each video–modality pair has a \emph{fixed ground-truth caption} determined by its specific tag combination. 
This creates a structural mismatch for MUPO training, as modality dropout or random subsampling cannot be applied without invalidating the supervision signal. 
Consequently, similar to \textsc{MAPO}, we restrict caption-level MUPO training to the \textit{VAS} condition on the same underlying videos, ensuring consistency between the provided inputs and the available ground-truth captions. 
This necessity leads to a significantly reduced training scale (\emph{all-combi}: 5140 captions; \emph{VAS-only}: 730 captions), but preserves the correctness of supervision while enabling modality-agnostic optimization in the captioning setting.

\subsection{Contrastive Reward Weighting}
\label{app:crw}
We apply a modality-aware reward weighting mechanism that adjusts the reward of each positive rollout according to its semantic similarity to a negative response. The final reward used for optimization is defined as:
\begin{equation}
R_{\mathrm{weighted}}
= \operatorname{clip}_{[0,1]}\!\left(
R_{\mathrm{pos}} \cdot \left[ 1 + \alpha\, w(s;\tau) \right]
\right),
\label{eq:weighted_reward}
\end{equation}
where $R_{\mathrm{pos}}$ is the base positive reward, $\alpha$ is a scaling factor, $s$ is the similarity score between the positive and negative responses, and $\tau$ is a similarity threshold.

\paragraph{Positive and Negative Responses.}
For each sample $x$, we denote by $y^{+}$ a positive rollout generated under the complete modality set $M$, and by $y^{-}$ a negative response generated from the same sample under a deficit condition, where one or more modalities are removed or degraded. Thus, $M$ represents the full available modalities, while ``deficit'' indicates missing-modality or degraded-modality inputs.

\paragraph{Similarity Function.}
The similarity score $s$ is computed as:
\begin{equation}
s = \mathrm{Sim}\bigl( y^{+}\!\mid x,\; y^{-}\!\mid x \bigr),
\label{eq:similarity}
\end{equation}
where $\mathrm{Sim}(\cdot)$ measures semantic closeness between the embeddings of $y^{+}$ and $y^{-}$. In our experiments, we instantiate $\mathrm{Sim}(\cdot)$ using \textsc{BERTScore}, which produces a similarity value in the range $[0,1]$.

\paragraph{Gating Function.} We use discrete gating $w(s;\tau) = \mathbbm{1}\{s < \tau\}$, where $s$ is semantic distance and $\tau$ the threshold.




\paragraph{Interpretation.}
The formulation in Eq.~\ref{eq:weighted_reward} increases the effective reward when the positive rollout remains semantically distant from deficit-induced responses, while the clipping operator $\operatorname{clip}_{[0,1]}(\cdot)$ ensures numerical stability. This encourages the model to produce modality-consistent and semantically distinct outputs during training. In the single-modality setting, CRW has no effect because no modality variation exists to form deficit-induced negative responses.

\begin{tcolorbox}[
    colback=gray!4,
    colframe=gray!40,
    title=\textbf{CRW - Qualitative Example},
    fonttitle=\small,
    boxrule=0.6pt,
    arc=2mm,
    left=1.2mm,
    right=1.2mm,
    top=1.2mm,
    bottom=1.2mm
]
\noindent\textbf{Seed image (from video \texttt{AvM4H8CskWDk.14.mp4}).}\par
\vspace{0.4em}

{\centering
\includegraphics[width=0.55\linewidth]{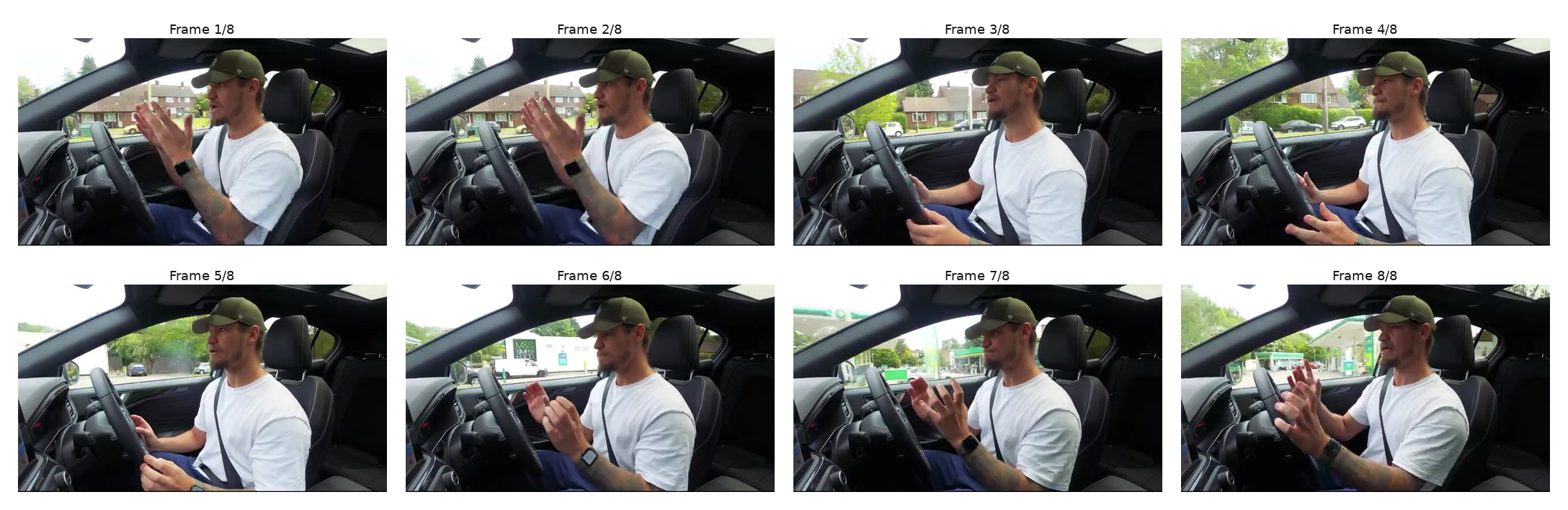}\par
}

\vspace{0.6em}

\noindent\textbf{Subtitle: I am going to be sensibly splitting the money between buying a daily and buying a supercar.}

\noindent\hfill\rule{1.0\linewidth}{0.4pt}\hfill\noindent

\vspace{0.4em}
\noindent\textbf{MAPO w/o CRW : }\par
\vspace{0.3em}

\par
\vspace{0.25em}

\textbf{A:} Someone discusses their plans to sensibly split money between buying a daily vehicle and a supercar while  \textcolor{blue}{driving in a car.}\par
\vspace{0.25em}

\textbf{AS:} Someone describes their plan to sensibly allocate money between purchasing a daily vehicle and a supercar.\par
\vspace{0.25em}

\textbf{S:} Someone explains their plan to responsibly divide their budget between buying a daily vehicle and a supercar.\par
\vspace{0.25em}

\textbf{V:} The video shows \textcolor{green}{a man sitting in the driver's seat of a car, wearing a white T-shirt, a green cap, and a seatbelt.} He \textcolor{green}{gestures with his hands} as he speaks, with \textcolor{green}{the car’s interior and outdoor scenery visible through the windows}.\par
\vspace{0.25em}

\textbf{VA:} A person in a moving car discusses their plan to sensibly split spending between a daily driver and a supercar.
\par
\vspace{0.25em}

\textbf{VAS:} A person inside a car explains their plan to responsibly allocate money between purchasing a daily driver and a supercar.
\par
\vspace{0.25em}

\textbf{VS:} The video shows a man \textcolor{green}{driving a car, wearing a white shirt and a cap, gesturing} as he talks about how he plans to sensibly divide his budget between buying a daily car and a supercar.\par
\vspace{0.25em}

\noindent\hfill\rule{1.0\linewidth}{0.4pt}\hfill\noindent

\vspace{0.4em}
\noindent\textbf{MAPO w/ CRW:}\par
\vspace{0.3em}

\textbf{A:} The audio describes someone discussing their plan to sensibly split money between buying a daily car and a supercar while  \textcolor{blue}{in a vehicle.}\par
\vspace{0.25em}

\textbf{AS:} The audio describes someone planning to sensibly split their money between purchasing a daily car and a supercar, while the subtitle \textcolor{orange}{reiterates this decision regarding financial allocation forthese two types of vehicles}.\par
\vspace{0.25em}

\textbf{S:} The context is about someone sensibly planning to split their money between purchasing a daily use item and a supercar.\par
\vspace{0.25em}

\textbf{V:} A man in a car, \textcolor{green}{wearing a cap and white shirt}, is \textcolor{green}{gesturing with his hands while driving}, with the \textcolor{green}{car's interior and some outdoor scenery visible through the windows}.\par
\vspace{0.25em}

\textbf{VA:} The video shows a man inside a car, \textcolor{green}{wearing a cap and white shirt, gesturing with his hands}, while \textcolor{blue}{speaking} about sensibly splitting money between buying a daily driver and a supercar.\par
\vspace{0.25em}

\textbf{VAS:} The video \textcolor{green}{shows a man inside a car}, \textcolor{green}{wearing a cap and white shirt,talking with  hand gestures} about sensibly splitting money between buying a daily car and a supercar.\par
\vspace{0.25em}

\textbf{VS:} The video shows a man in a car \textcolor{green}{gesturing with his hands} while talking about sensibly splitting money between buying a daily car and a supercar.\par
\vspace{0.25em}

\end{tcolorbox}

\begin{figure}[t]
    \centering
    \includegraphics[width=0.95\linewidth]{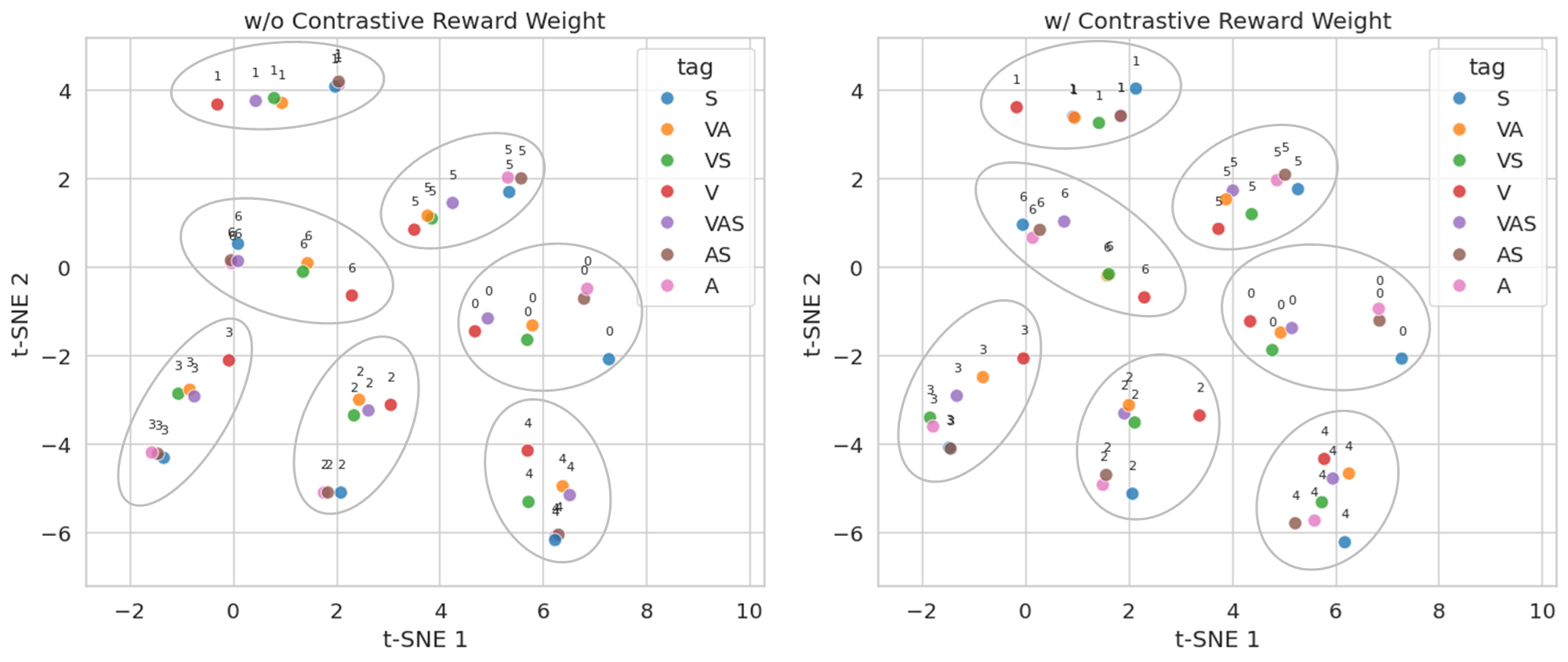}
    \caption{: Joint t‑SNE visualization of response representations before and after applying CRW. Points that were densely overlapped within a sample group are more widely dispersed under CRW, indicating reduced representation collapse. The number annotated above each point denotes the corresponding video ID.}
    \label{fig:crw_tsne}
\end{figure}

\section{Contributions \& Acknowledgments}
We sincerely thank all members for their core contributions across idea origination, benchmark design, method design, implementation, experiments, writing, and project coordination, as detailed in Table~\ref{tab:contributions}.

\begin{table}[h]
\centering
\caption{Author Contributions}
\label{tab:contributions}
\centering
\scriptsize

\resizebox{\linewidth}{!}{
\begin{tabular}{
    l
    >{\raggedright\arraybackslash}p{2.4cm}
    >{\raggedright\arraybackslash}p{2.4cm}
    >{\raggedright\arraybackslash}p{1.4cm}
    >{\raggedright\arraybackslash}p{2.2cm}
    >{\raggedright\arraybackslash}p{1.2cm}
    >{\raggedright\arraybackslash}p{1.2cm}
    >{\raggedright\arraybackslash}p{1.0cm}
    >{\raggedright\arraybackslash}p{1.2cm}
}
\toprule
Author 
& Problem Formulation \& Core Ideation
& Auxiliary Ideation \& Ablation Design
& Benchmark Creation
& Method Design \& Core Implementation
& Experiment Execution
& Writing
& Project Coordination
& Advisory \& Supervision \\
\midrule
Nikhil Verma & \checkmark &  &  & \checkmark & \checkmark & \checkmark & \checkmark & \\
\midrule 
Minjung Kim &  & \checkmark &\checkmark &  &\checkmark & \checkmark & \checkmark &  \\
\midrule 
JooYoung Yoo & \checkmark &  & \checkmark & \checkmark & \checkmark & \checkmark &   \\
\midrule 
Kyung-Min Jin &  & \checkmark &  \checkmark &  & \checkmark & \checkmark &  \\
\midrule 
Manasa Bharadwaj &  &  &  & &  &  & \checkmark & \checkmark\\
\midrule 
Kevin Ferreira &  &   &  & &  &  &  & \checkmark \\
\midrule 
Ko Keun Kim &  &   &  &  &  &  &  &  \checkmark  \\
\midrule 
Youngjoon Kim &  &   &  &  &  &  &  & \checkmark \\
\bottomrule
\end{tabular}
}
\end{table}


\end{document}